\documentclass[10pt,journal,compsoc]{IEEEtran}

\usepackage{booktabs}
\usepackage{graphicx}
\usepackage{epstopdf}
\usepackage{amsmath}
\usepackage{subfigure}
\usepackage{algorithmic}
\usepackage{algorithm}
\usepackage{bm}
\usepackage{multirow}

\usepackage{amssymb}
\usepackage[colorlinks,linkcolor=red]{hyperref}
\usepackage{colortbl}
\usepackage{url}
\definecolor{mygray1}{gray}{.9}
\definecolor{mygray2}{gray}{.7}
\begin{document}

\title{Latent Semantic Consensus For Deterministic Geometric Model Fitting}

\author{Guobao Xiao,~\IEEEmembership{Senior Member,~IEEE,}
            Jun Yu,~\IEEEmembership{Senior Member,~IEEE,}
           Jiayi Ma,~\IEEEmembership{Senior Member,~IEEE,}
            Deng-Ping Fan,~\IEEEmembership{Senior Member,~IEEE,}
            and Ling Shao,~\IEEEmembership{Fellow,~IEEE}
\IEEEcompsocitemizethanks{\IEEEcompsocthanksitem Guobao Xiao is with School of Electronics and Information Engineering, Tongji University, Shanghai, 201804, China.
\IEEEcompsocthanksitem Jun Yu is with School of Computer Science and Technology, Hangzhou Dianzi University, Hangzhou, 310018, China, and he is also with the Department of Computer Science and Technology, Harbin Institute of Technology (Shenzhen), Shenzhen, China.
\IEEEcompsocthanksitem Jiayi Ma is with Electronic Information School, Wuhan University, Wuhan, 430072, China.
\IEEEcompsocthanksitem Deng-Ping Fan is with Nankai International Advanced Research Institute (SHENZHEN FUTIAN), Nankai University, Shenzhen, 518045, China, and he is also with College of Computer Science, Nankai University, Tianjin, 300071, China.
\IEEEcompsocthanksitem Ling Shao is with UCAS-Terminus AI Lab, University of Chinese Academy of Sciences, Beijing, 100049, China.
\IEEEcompsocthanksitem Jun Yu is the corresponding author.
}
}

\markboth{IEEE TRANSACTIONS ON PATTERN ANALYSIS AND MACHINE INTELLIGENCE}%
{Shell \MakeLowercase{\textit{et al.}}: Bare Demo of IEEEtran.cls for Computer Society Journals}
\IEEEtitleabstractindextext{%
\begin{abstract}
Estimating reliable geometric model parameters from the data with severe outliers is a fundamental and important task in computer vision. This paper attempts to sample high-quality subsets and select model instances to estimate parameters in the multi-structural data. To address this, we propose an effective method called Latent Semantic Consensus (LSC). The principle of LSC is to preserve the latent semantic consensus in both data points and model hypotheses. Specifically, LSC formulates the model fitting problem into two latent semantic spaces based on data points and model hypotheses, respectively. Then, LSC explores the distributions of points in the two latent semantic spaces, to remove outliers, generate high-quality model hypotheses, and effectively estimate model instances. Finally, LSC is able to provide consistent and reliable solutions within only a few milliseconds for general multi-structural model fitting, due to its deterministic fitting nature and efficiency. Compared with several state-of-the-art model fitting methods, our LSC achieves significant superiority for the performance of both accuracy and speed on synthetic data and real images. The code will be available at https://github.com/guobaoxiao/LSC.
\end{abstract}

}

\maketitle
\IEEEdisplaynontitleabstractindextext
\IEEEpeerreviewmaketitle
\IEEEraisesectionheading{\section{Introduction}\label{sec:introduction}}
\label{intro}
\IEEEPARstart{T}{his} study focuses on the problem of estimating reliable geometric model parameters from the data with severe outliers. Many computer vision tasks, such as homography/fundamental matrix estimation~\cite{wang2019}, motion segmentation~\cite{xiao2021segmentation}, vanishing point detection~\cite{lee2019joint}, 3D reconstruction~\cite{ma2021image} and pose estimation~\cite{9444575}, involve the geometric models. 
\begin{figure}
\centering
\begin{minipage}{.22\textwidth}
\centerline{\includegraphics[width=1.0\textwidth]{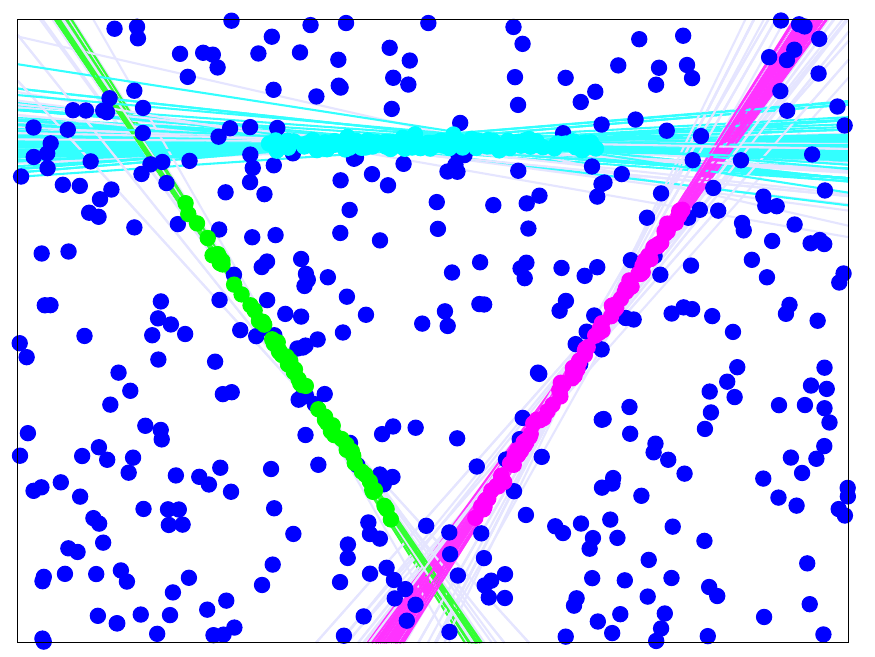}}
  \centerline{\footnotesize(a) Input data}
  \centerline{\includegraphics[width=1.0\textwidth]{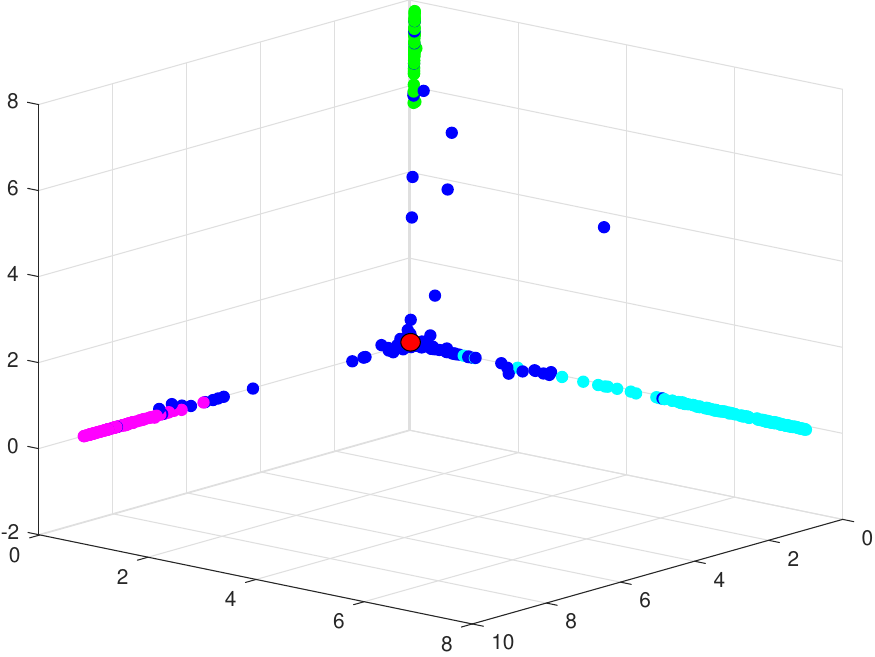}}
  \centerline{\footnotesize(c) LSS of data points}
\end{minipage}
\begin{minipage}{.22\textwidth}
\centerline{\includegraphics[width=1.0\textwidth]{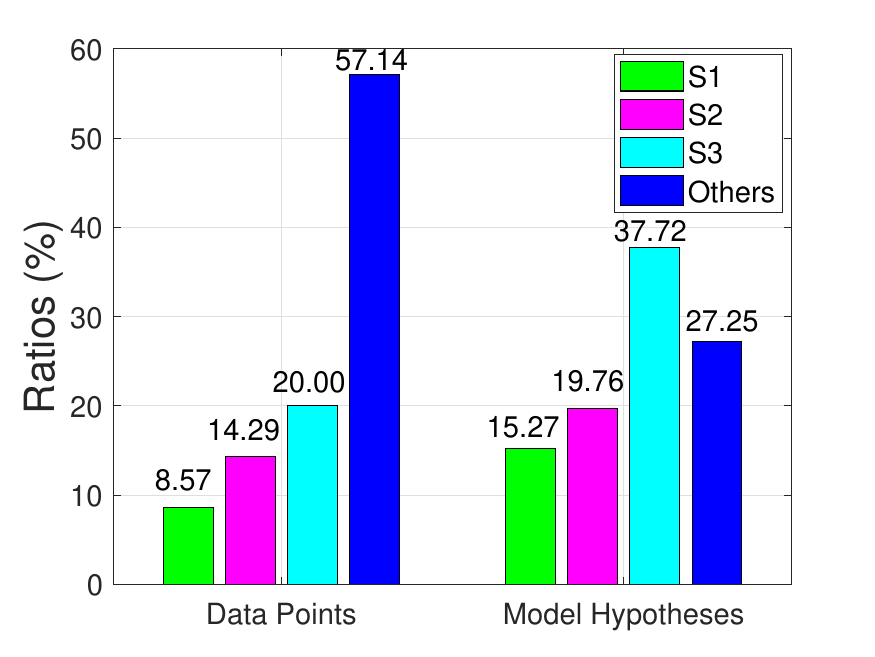}}
  \centerline{\footnotesize(b) Data distribution}
  \centerline{\includegraphics[width=1.0\textwidth]{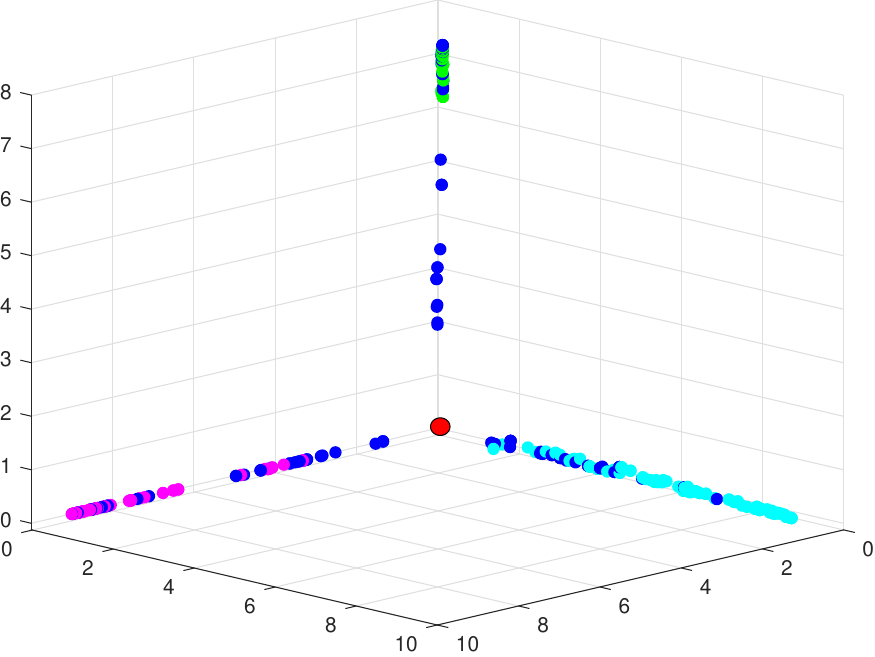}}
  \centerline{\footnotesize(d) LSS of model hypotheses}
\end{minipage}
\caption{An example of latent semantic space for line fitting. (a) The input data with data points and the model hypotheses generated by the proposed sampling method. (b) The distribution of inliers and model hypotheses corresponding to different model instances (i.e., S1, S2 and S3). Others are gross outliers and bad model hypotheses. (c) and (d) The latent semantic space of data points and model hypotheses. The gross outliers and bad model hypotheses are marked in blue, and the inliers and good model hypotheses are marked in different color. The red point is the origin of latent semantic space.}
\label{fig:firstexample}
\end{figure}

{A model fitting problem typically involves two key steps: first, sampling a set of minimal subsets to generate model hypotheses. Here, a minimal subset is the smallest set of data points needed for formulating a model hypothesis, such as $2$ points for a line or a minimum of $4$ points for a homography matrix. The second step is selecting the best model hypotheses as the estimated model instances.

Among the sampling algorithms, RANSAC~\cite{fischler1981random} stands out as one of the most popular, valued for its simplicity and effectiveness.
However, most existing sampling algorithms, including various guided ones~\cite{chin2012accelerated,tennakoon2018effective}, face challenges in providing intractable and consistent solutions due to their inherently random nature. Then, some deterministic sampling methods, e.g., \cite{IJCVXiao2019,li2009consensus,chin2015efficient,le2019de,xiao2020deterministic}, play an important roles for a model fitting method in practical applications. They not only offer tractability but also yield stable sampling results with the same input parameters. Yet, current deterministic sampling methods, i.e., \cite{li2009consensus,chin2015efficient,le2019de}, often adopt a global optimization strategy, limiting their applicability to single-structure data model fitting problems. SDF~\cite{IJCVXiao2019} and LGF~\cite{xiao2020deterministic} address multi-structural model fitting problems, but are confined to handling two-view model fitting problems, as they rely on correspondence information like matching scores, lengths, and directions.}

To address these challenges, we propose the Latent Semantic Consensus (LSC) method. LSC includes a novel latent semantic consensus based sampling algorithm to handle the general multi-structural deterministic sampling problem. {Specifically}, we first sample a small number of minimal subset in a deterministic way. Then, we map the data points and the generated model hypotheses into two {Latent Semantic Spaces (LSSs)}, as shown in Fig.~\ref{fig:firstexample}. For the space of data points, inliers belonging to different model instances are mapped into several independent subspaces, while gross outliers\footnote{Outliers encompass gross outliers and pseudo-outliers: the former are outliers for all model instances in the data, while the latter are inliers for some model instances but outliers for others.} are {concentrated} close to the origin. {Simultaneously, in the LSS of model hypotheses, those} corresponding to the same model instance are mapped into an independent subspace, {while others cluster near} the origin.

{Note that}, the effectiveness of {LSS} depends on the quality of data points and the generated model hypotheses. This is because the outliers and bad model hypotheses will affect the distributions of corresponding points in {LSS}. Thus, we preserve the latent semantic consensus of two {LSSs}, to sample high-quality minimal subsets. {Initially,} we remove some outliers, {specifically focusing on those} points close to the origin of LSS. {Subsequently,} we select the data points that correspond to neighbors of each point in {LSS} as a minimal subset, {leveraging the fact that} the data points belonging to the same model instance share similar distributions in {LSS. Following this, we remap data points and newly} generated model hypotheses to two {LSSs} again. Finally, we {further} remove some outliers and bad model hypotheses to preserve the latent semantic consensus.

For the second step of model fitting, we design a novel model selection in view of the advantage of our high-quality model hypotheses. Specifically, we explore the latent semantic information in {LSS} of model hypotheses. {In this analysis, we refrain from scrutinizing LSS of data points to mitigate} the sensitivity to unbalanced data distribution. Recall that two points share a similar value in a subspace of {LSS}, if they correspond to the same model instance. Thus, we propose to label the points into their respective subspaces in {LSS} via an integer linear program. After that, we select the best model hypothesis among the ones that share the same labels as an estimated model instance. Notably, in {LSS}, all high-order model fitting problems (e.g., homography/fundamental matrix estimation) are regressed to a subspace recovery problem, which is much easier to be dealt with.

Finally, we summarize the key contributions of this work as follows:
\begin{itemize}
\item We propose to preserve latent semantic consensus in {LSS} of data points, for deterministically sampling high-quality minimal subsets. To the best of our knowledge, we are the first one to address general deterministic sampling problem for multi-structural data.
\item We design a novel model selection algorithm {that} formulates the model fitting problem as a simple subspace recovery problem based on {LSS} of model hypotheses.
\item The qualitative and quantitative experiments on several model fitting tasks show that the proposed fitting method is able to achieve better results over several state-of-the-art competing methods.
\end{itemize}

The rest of the paper is organized as follows: We review some related work in Sec.~\ref{sec:relatedwork}, and describe the components of the proposed fitting method in Sec.~\ref{sec:methodology}. We present the experimental results in Sec.~\ref{sec:experiments}. We {perform some ablation studies on different components of LSC} in Sec.~\ref{sec:limitations} and draw conclusions in Sec.~\ref{sec:conclusion}.
\section{Related Work}
\label{sec:relatedwork}
In this section, we briefly review some related works on random and deterministic model fitting.
\subsection{Random Model Fitting Methods}
Random model fitting methods can be classified into sampling and model selection methods according to their process.
{For the first step, to hit all model instances in data, that is, there is at least an all-inlier minimal subset sampled for each model instance, the sampling problem approximately possesses a combinatorial nature, making the number of minimal subsets huge. For an example of data with the outlier ratio $\alpha$, RANSAC~\cite{fischler1981random}, one of the most popular sampling algorithms, requires to sample $\frac{log(1-\beta)}{log(1-(1-\alpha)^\rho)}$ minimal subsets, to hit a model instance with the probability $\beta$. Here $\rho$ is the minimum number of data points in a minimal subset. Even worse, the sampled minimal subsets contain a large number of bad ones.} Then, there are many variants of RANSAC to improve the performance on more complex cases, e.g., ROSAC~\cite{chum2005matching}, LO-RANSAC~\cite{chum2003locally}, HMSS~\cite{tennakoon2016robust} and CBS~\cite{tennakoon2018effective}. 

For model selection methods, they can be roughly classified into the consensus analysis-based and the preference analysis-based methods, according to the space they perform. The consensus analysis-based fitting methods, e.g., MSHF~\cite{wang2019} and CBG~\cite{lin2022co}, select the best model hypotheses as the estimated model instances; The preference analysis-based methods, e.g., SWS~\cite{hypergraphWithLargeHyperedges}, RPA~\cite{magri2017multiple}, RansaCov~\cite{Magri_2016_CVPR}, T-linkage~\cite{Magri_2014_CVPR}, MultiLink~\cite{Magri_2021_CVPR}, MCT~\cite{Magri_2019_CVPR} and CLSA~\cite{xiao2021segmentation}, label data points and then estimate the parameters of model instances.
These two kinds of model fitting methods have their own advantages and disadvantages. Specifically, the former methods are not sensitive to unbalanced data distribution, but they rely heavily on the quality of model hypotheses; in contrast, the latter methods are very sensitive to unbalanced data distribution, but they estimate model instances based on data points, so they do not rely heavily on the generated model hypotheses. Thus, there is not a model selection algorithm that has clear advantage.

Deep learning-based methods have gained popularity in various computer vision tasks, including model fitting~\cite{liu2023pgfnet,chen2023ssl,miao2024bclnet,guo2023graph}. These methods offer a significant advantage in learning powerful representations directly from raw data, eliminating the need for hand-crafted features. However, a drawback is their dependency on large amounts of training data, and its sensitivity to outliers, particularly in multi-structural datasets.
\subsection{Deterministic Model Fitting Methods}
Compared with random model fitting methods whose results often vary, deterministic model fitting methods are able to provide stable solutions and they also can be tractable. Most of currently existing deterministic model fitting methods cast the fitting task as a global optimization problem. Such as, Li~\cite{li2009consensus} and Chin et al. \cite{chin2015efficient} adopted a tailored branch-and-bound scheme and the A* algorithm to search for a globally optimal solution, respectively; Doan et al.~\cite{doan2022hybrid} and Le et al.~\cite{le2019de} introduced a hybrid quantum-classical algorithm and the maximum consensus criterion for deterministic fitting, respectively;  Fan et al.~\cite{fan2021efficient} discussed two loss functions to seek geometric models.

These above deterministic methods only work for single-structural data and they also often suffer from high computational complexity. Xiao et al.~\cite{IJCVXiao2019,xiao2020deterministic} used the superpixel information and the min-hash technique to propose two heuristic methods for multiple-structural deterministic model fitting. However, they only work for two-view fitting problem due to the information of correspondences they depend on. That is, they cannot be used for general fitting problems, e.g., line fitting and circle fitting.

This paper also proposes a deterministic method, but, it not only can work for multiple-structural data, but also has not the limitation of two-view model fitting. That is, this paper proposes a deterministic methods by preserving latent semantic consensus for general multiple-structural model fitting problems.

{It is important to note that CLSA~\cite{xiao2021segmentation} and the proposed method (LSC) both utilize latent semantic analysis to tackle model fitting problems. However, they differ significantly in the following aspects: 1) CLSA employs LSA to eliminate outliers in LSS, whereas LSC preserves latent semantic consensus in LSS for minimal subset sampling. 2) CLSA constructs LSS once, and its effectiveness depends on the input data; in contrast, LSC dynamically updates LSS based on generated model hypotheses, enhancing its effectiveness. 3) CLSA is a random model selection method, whereas LSC follows a deterministic approach for both sampling and model selection. Consequently, LSC provides more stable and superior fitting results compared to CLSA.}
\section{Proposed Method}
\label{sec:methodology}
In this section, we introduce the details of the Latent Semantic Consensus based model fitting method (called LSC). We first describe the problem formulation of model fitting (Sec.~\ref{subsec:problemformulation}). Then, we introduce latent semantic for model fitting (Sec.~\ref{subsec:lscmf}). After that, we propose the latent semantic consensus based sampling algorithm (Sec.~\ref{subsec:lss}) and model selection algorithm (Sec.~\ref{subsec:lsssa}). Finally, we summarize and analyze LSC in Sec.~\ref{sec:completemethod}.
\subsection{Problem Formulation}
\label{subsec:problemformulation}
A geometric model is usually provided beforehand, such as, line, circle, homography matrix, fundamental matrix, etc, for model fitting. Then, given $n$ data points $S=\{s_1,\ldots,s_i,\ldots,s_n\}$, the output of geometric model fitting is the parameter of model instances and the labels of data points. Here, a model instance (or structure) is an instance of the given geometric model, and the labels of data points include inliers belonging to different model instances and outliers.

As aforementioned, a model fitting problem generally contains minimal subsets sampling for model hypothesis generation, and model selection based on the generated model hypotheses. Here, a model hypothesis is a parameter formulation of the given geometric model. For the model of a line, a model hypothesis is $[a~b~c]$, where $ax_i+by_i+c=0$ for a data point $s_i=\{x_i,y_i\}$; For the model of a circle, a model hypothesis is $[a~b~r]$, where $(a,b)$ is the coordinate values of the center of a circle, and $r$ is the value of radius, and {$(x_i-a)^2+(y_i-b)^2=r^2$} for a data point $s_i=\{x_i,y_i\}$; For the model of a homography matrix, a model hypothesis $\theta$ is $3\times3$ matrix in $\mathbb{R}^{3\times3}$, where $\begin{bmatrix} x'_i~y'_i~1\end{bmatrix}^T=\theta\begin{bmatrix} x_i~y_i ~1\end{bmatrix}^T$ for a data point $s_i=(x_i,y_i,x'_i,y'_i)$, and $(x_i,y_i)$ and $(x'_i,y'_i)$ are the image coordinates of the two feature points in the image pair. Of course, the geometric model is not limited to line, circle and homography matrix.

\subsection{Latent Semantic Analysis for Model Fitting}
\label{subsec:lscmf}
{In this section, we delve into Latent Semantic Analysis (LSA)~\cite{lsa1990} as a key component of our deterministic model fitting approach. Drawing inspiration from topic modeling, a widely used technique in uncovering word-use patterns and document connections in word processing~\cite{hao2021}, LSA proves instrumental. Much like how words within the same topic exhibit similar distributions, our data points and model hypotheses share an analogous relationship to words and documents. In the realm of model fitting, certain data points align with inliers of a model hypothesis, mirroring the association between words and documents in topic modeling. The crux lies in recognizing that some model hypotheses correspond to actual model instances. Just as certain words form parts of a document, specific documents correspond to a broader topic. The parallels between these relationships underscore the efficacy of employing topic modeling principles for our model fitting endeavors.

Among various topic modeling methods~\cite{wang2021survey,wang2020kernelized,qian2023adaptive}, we opt for LSA to explore latent semantic information in our deterministic model fitting. Its simplicity, effectiveness, and non-interference with our deterministic approach make it an ideal choice.

To scrutinize the relationship between model hypotheses and data points via LSA, we initiate by constructing a preference matrix~\cite{Magri_2014_CVPR}. This matrix reflects the relationship between model hypotheses and data points based on the residual value. In essence, for a given data point $s_i$ and a model hypothesis $\theta_j$, the preference value is defined as:}

\begin{align}
\label{equ:preferencefunction}
p (s_i,\theta_j) =\exp\{-\frac{d_r(s_i,\theta_j)}{\psi}\},
\end{align}
where $d_r(s_i,\theta_j)$ is the Sampson distance~\cite{torr1997development} (residual value) between $\theta_j$ and $s_i$, and $\psi$ is a non-zero threshold. We note that, for a large residual value, the corresponding preference value is small; Conversely, for a small residual value, the corresponding preference value is large. 


Then we can construct a preference matrix ${\bf P}_{n\times m}$ by Eq.~(\ref{equ:preferencefunction}) for $n$ data points and $m$ model hypotheses. {Note that, the $m$ initial model hypotheses are generated by $\rho$ neighbors of each data point ($\rho$ is the number of a minimal subset), to preserve the deterministic nature of the proposed method.} After that, we introduce LSA to decompose the preference matrix. That is, we decompose ${\bf P}_{n\times m}$ by singular value decomposition (SVD) as follows:

\begin{align}
\label{equ:svdoriginal}
\begin{split}
{\bf P}_{n\times m}&={\bf U}{\bf \Sigma} {\bf V}^T,\\
  \end{split}
\end{align}
i.e.
\begin{footnotesize}
  \begin{align}
\label{equ:svdoriginal21}
&\begin{bmatrix} p(s_1,\theta_1) & \cdots &p(s_1,\theta_m)\\ \vdots&\ddots & \vdots\\ p(s_n,\theta_1)&\cdots &p(s_n,\theta_m)\end{bmatrix}={\bf U}_{n\times r}\begin{bmatrix} \sigma_1 &\cdots& 0\\ \vdots&\ddots & \vdots\\ 0&\cdots& \sigma_r\end{bmatrix}_{r\times r}{\bf V}^T_{m\times r},
\end{align}
\end{footnotesize}
where ${\bf U}$ and {${\bf V}$} are the orthogonal matrices of left and right singular vectors ${\bf U}{\bf U}^T={\bf V}{\bf V}^T=I$ (here $I$ is an identity matrix), respectively. ${\bf \Sigma}=diag (\sigma_1,\ldots,\sigma_r)$ is a diagonal matrix with singular values on the diagonal and the singular values are listed in non-increasing order, i.e., $\sigma_1\geq\sigma_2\geq\cdots\geq\sigma_r>0$, $r=rank({\bf P})$.

Then, the preference matrix is approximately decomposed by the top $k~{(\leq r)}$ singular values in ${\bf \Sigma}$, since the top $10\%$ (even $1\%$) of singular values account for over $99\%$ of all singular values, for SVD.  i.e.,{
\begin{align}
\label{equ:svdapproxi}
{\bf{P}}\approx{\bf U}_{n\times k}{\bf \Sigma}_{k\times k} {\bf V}^T_{m \times k}.
\end{align}}
Based on Eq.~(\ref{equ:svdapproxi}), we construct two latent semantic spaces (LSS) for data points and model hypotheses, respectively. For LSS of data points (we call DP-LSS), the points-to-points inner products are written as:
 \begin{align}
 \label{equ:lsaoriginal2}
   \begin{split}
\bf {P} \bf{P}^T&\approx{\bf U}_{n\times k}{\bf \Sigma}_{k\times k}\underbrace{{\bf V}^T_{m\times k}{\bf V}_{m\times k}}_I{\bf \Sigma}_{k\times k}{\bf U}^T_{n\times k}\\
&={\bf U}_{n\times k}{\bf \Sigma}^2_{k\times k}{\bf U}^T_{n\times k}.
  \end{split}
\end{align}
Thus, following LSA, we utilize the rows of ${\bf P}_s={\bf U}_{n\times k}{\bf \Sigma}_{k\times k}\in \mathbb{R}^{n\times k}$ as coordinates for DP-LSS. By Eq.~(\ref{equ:lsaoriginal2}), data points are project onto LSS, and meanwhile, the dimension $m$ of $ {\bf P}\in \mathbb{R}^{n\times m}$ is also reduced to {$k$}.

For LSS of model hypotheses (we call MH-LSS), the hypotheses-to-hypotheses inner products are written as:
 \begin{align}
 \label{equ:lsaoriginal3}
   \begin{split}
 {\bf P}^T {\bf P}&\approx{\bf V}_{m\times k}{\bf \Sigma}_{k\times k}\underbrace{{\bf U}^T_{n\times k}{\bf U}_{n\times k}}_I{\bf \Sigma}_{k\times k}{\bf V}^T_{m\times k}\\
 &={\bf V}_{m\times k}{\bf \Sigma}_{k\times k}^2{\bf V}^T_{m\times k}.
  \end{split}
\end{align}
Thus, we adopt the rows of {${\bf P}_\theta={\bf V}_{m\times k}{\bf \Sigma}_{k\times k}$} as the coordinates for model hypotheses in MH-LSS. Also, the dimension $n$ of {${\bf P}\in \mathbb{R}^{n\times m}$} is reduced to {$k$}.
\subsection{Latent Semantic Consensus based Sampling Algorithm}
\label{subsec:lss}
In this subsection, we propose a novel latent semantic consensus based sampling algorithm (called LSC-SA). To alleviate the bad influence of outliers, we first remove outliers by analyzing the distribution of the points in DP-LSS.

\begin{figure}[t]
\centering
\begin{minipage}{.35\textwidth}
\centerline{\includegraphics[width=1.0\textwidth]{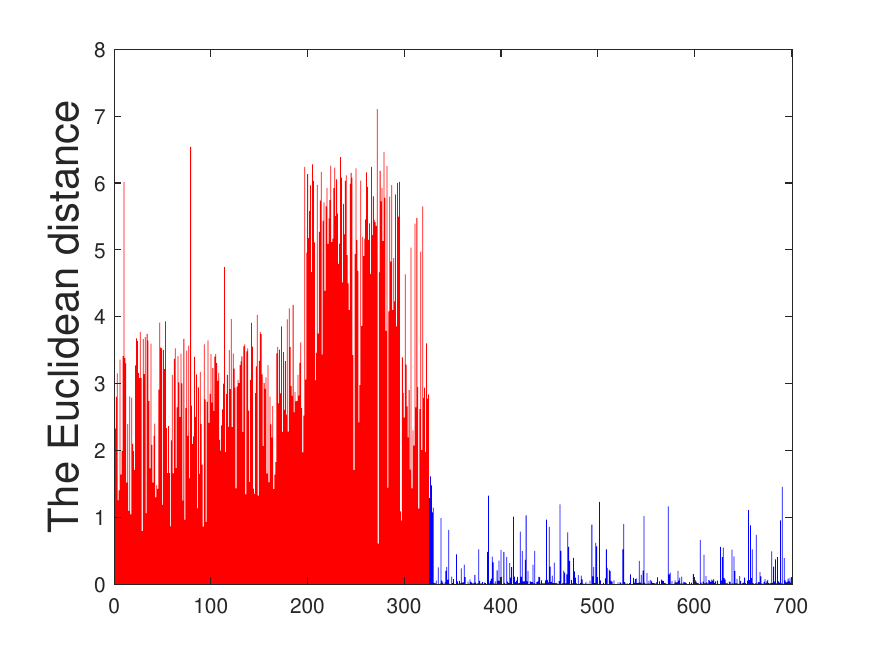}}
\end{minipage}
\caption{An example of the distance distribution from inliers (red) and outliers (blue) in DP-LSS. }
\label{fig:distributionofdistance}
\end{figure}
The points in DP-LSS corresponding to inliers are often situated farther from the origin of coordinates, while other points tend to be closer to the origin, as depicted in Fig.~\ref{fig:distributionofdistance}. This is attributed to the fact that the value of ${\bf P}_s$ is influenced by the preference value, and given that inliers typically receive a larger preference value than outliers. 

Given a data point $s_i$ and the corresponding projected point $\hat{s}_i$ in DP-LSS, the distance between $\hat{s}_i$ and the origin in DP-LSS can be formulated as:
 \begin{align}
 \label{equ:distancexi}
 d_{\hat{s}_i}^o=\sqrt{(u_{i1}\sigma_i)^2+(u_{i2}\sigma_i)^2+\cdots+(u_{i{k}}\sigma_i)^2},
\end{align}
where ${\bf U'}_i=[u_{i1}~u_{i2}~\cdots~u_{i{k}}]$ is the $i$-th row of ${\bf U}$. Then given the $i$-th row of ${\bf P}$, i.e., ${\bf P}_i=[p(s_i,\theta_1)~p(s_i,\theta_2)~\cdots~p(s_i,\theta_m)]$, we can decompose $p_i$ as:
 \begin{align}
  \label{equ:decomposepi}
 {\bf P}_i\approx{\bf U'}_i\sigma_i{\bf V}^T_{m\times {k}}.
\end{align}
For ${\bf P}_i$, the inner products are written as:
 \begin{align}
  \label{equ:innerpi}
    \begin{split}
 {\bf P}_i {\bf P}_i^T&=[p(s_i,\theta_1)~p(s_i,\theta_2)~\cdots~p(s_i,\theta_m)]\\
 &~~~~~[p(s_i,\theta_1)~p(s_i,\theta_2)~\cdots~p(s_i,\theta_m)]^T\\
 &=[p(s_i,\theta_1)^2+p(s_i,\theta_2)^2+\cdots+p(s_i,\theta_m)^2].
    \end{split}
\end{align}
For ${\bf U'}_i\sigma_i{\bf V}^T_{m\times {k}}$, the inner products are written as:
 \begin{align}
  \label{equ:innerxi}
    \begin{split}
 {\bf U'}_i\sigma_i{\bf V}^T{\bf V}\sigma_i{\bf U'}_i^T&={\bf U'}_i\sigma_i^2{\bf U'}_i^T\\
 &=[u_{i1}~u_{i2}~\cdots~u_{i{k}}]\sigma_i^2[u_{i1}~u_{i2}~\cdots~u_{i{k}}]^T\\
 &=[(u_{i1}\sigma_i)^2+(u_{i2}\sigma_i)^2+\cdots+(u_{i{k}}\sigma_i)^2].
    \end{split}
 \end{align}

From Eqs.~(\ref{equ:distancexi}),~(\ref{equ:decomposepi}),~(\ref{equ:innerpi}) and (\ref{equ:innerxi}), we rewrite Eq.~(\ref{equ:distancexi}) as:
 \begin{align}
 \label{equ:distancexi2}
 d_{\hat{s}_i}^o\approx \sqrt{p(s_i,\theta_1)^2+p(s_i,\theta_2)^2+\cdots+p(s_i,\theta_m)^2}.
\end{align}

Therefore, the distance ${\bf D}_{\hat{s}}^o=\{\hat{d}_{\hat{s}_i}^o\}_{i=1,\ldots,n}$ between points and the origin in DP-LSS depends on the preference value. Leveraging the above characteristic allows us to effectively identify and remove outliers. To this end, we introduce information theory~\cite{ferraz2007density} to analyze the distance ${\bf D}_{\hat{s}}^o$. Specifically, we first measure the gap $g_i$ between $\hat{d}_{\hat{s}_i}^o$ and the maximum value in ${\bf D}_{\hat{s}}^o$  as:
 \begin{align}
 g_i= \max\{{\bf D}_{\hat{s}}^o\}-\hat{d}_{\hat{s}_i}^o.
 \end{align}
 Then, we compute the quantity of information provided by each mapped point $\hat{s}_i$:
 \begin{align}
        Q(\hat{s}_i)=-\log \frac{g_i}{\sum_{i=1}^n g_i},
 \end{align}
 where $\frac{g_i}{\sum_{i=1}^n g_i}$ represents the probability of $\hat{s}_i$ to be an outlier.

After that, we remove the points with quantity of information lower than an entropy value and retain the other points:
  \begin{align}
  \label{equ:outlierremove}
{\bf \hat{S}}_I=\{\hat{s}_i|Q(\hat{s}_i)>-\sum_{i=1}^n\big(\frac{g_i}{\sum_{j=1}^n g_j}\log \frac{g_i}{\sum_{j=1}^n g_j}\big)\},
 \end{align}
 where $-\sum_{i=1}^n\big(\frac{g_i}{\sum_{j=1}^n g_j}\log \frac{g_i}{\sum_{j=1}^n g_j}\big)$ is the estimated entropy value.

{After that, we integrate latent semantic consensus into DP-LSS to guide the sampling of subsets for model hypothesis generation. Specifically}, the value of a point in DP-LSS is derived from the corresponding preference value{, fostering similarity among} inliers belonging to the same model instance. {With} the remaining points ${\bf \hat{S}}_I$, we search for $\rho$ (the minimum number of data points to generate a model hypothesis) neighbors for each {point. Subsequently,} we sample subsets from the input data points, {with each subset consisting of neighbors from DP-LSS}. In addition, to further refine the sampled subsets, we introduce the model hypothesis updating strategy~\cite{IJCVXiao2019}. {This involves leveraging global residual information to mitigate sub-optimal results. The overall process is summarized in the proposed Latent Semantic Consensus-based Sampling Algorithm (LSC-SA), outlined in Algorithm~\ref{alg:LSC-SA}.

 In Algorithm~\ref{alg:LSC-SA}, to} select the best model hypotheses in the model hypothesis updating strategy, we assign a model hypothesis $\theta_i$ a positive weighting value~\cite{wang2012simultaneously}:
\begin{equation}
\label{equ:weighting}
w(\theta_i)=\frac{1}{n}\sum_{j=1}^n\frac{\mathbf{EK}(d_r(s_j,\theta_i)/{{b}({\theta_i})})}{\delta({\theta_i}){b}({\theta_i})},
\end{equation}
where $\delta({\theta_i})$ is the estimated inlier noise scale of the $i$-th model hypothesis, and ${b}({\theta_i})$ is the bandwidth of the $i$-th model hypothesis, and it is defined as \cite{wand1994kernel}:
\begin{align}
\label{equ:bandwith}
{b}(\theta_i)=\left[\frac{243\int_{-1}^1{\mathbf{EK}(\lambda)}^2d\lambda}{35n\int_{-1}^1{\lambda}^2\mathbf{EK}(\lambda)d\lambda}\right]^{0.2}\delta({\theta_i});
\end{align}
For the kernel function $\mathbf{EK}(\cdot)$, we employ the popular Epanechnikov kernel:
\begin{align}
\label{equ:kernel}
\mathbf{EK}(\lambda) &=\left\{ \begin{array}    {r@{\quad \quad} l}
0.75(1-{\|\lambda\|}^2), & \|\lambda\|\leq 1,\\
0~~~~~~~~,&\|\lambda\|> 1.
\end{array}\right.\;
\end{align}
By Eq.~(\ref{equ:weighting}), a good model hypothesis is assigned a large weight since it {includes} more inliers with small residual values.


\begin{algorithm}[t] 
\small
\renewcommand{\algorithmicrequire}{\textbf{Input:}}
\renewcommand\algorithmicensure {\textbf{Output:} }
\caption{The proposed {LSC-SA}} 
\label{alg:LSC-SA} 
\begin{algorithmic}[1] 
\REQUIRE 
Data points $S=\{s_i\}_{i=1}^n$, the rank value ${k}$.
\STATE Search for $\rho$ neighbors of each data point to generate initial model hypotheses;
\STATE Construct latent semantic space (DP-LSS) for data point by Eq.~(\ref{equ:lsaoriginal2});
\STATE Remove the points corresponding to outliers in DP-LSS by Eq.~(\ref{equ:outlierremove});
\STATE Search for $\rho$ neighbors of each point in DP-LSS;
\STATE Re-generate model hypotheses $\Theta=\{\theta_1,\theta_2,\ldots,\theta_{qn}\}$ by using the neighbor information in DP-LSS;
\FOR {$i=1$ to $qn$}
\STATE $l\Leftarrow 1$, $\hat{\theta}_i^{l}\Leftarrow\theta_q^i$;
\REPEAT
\STATE Compute the residual between $\hat{\theta}_i^{l}$ and $S$;
\STATE Obtain the ranking list $\{rank_j^l\}_{j=1}^n$ according to the residual;
\STATE $\hat{\theta}_i^{l+1}\Leftarrow LSFit (s_{\{rank_j^l\}_{j=10-\rho+1}^{10}})$ // Update $\hat{\theta}_i^{l}$;
\STATE Assign a weighting value $w(\hat{\theta}_i^{l+1})$ to $\hat{\theta}_i^{l+1}$ by Eq.~(\ref{equ:weighting});
\UNTIL ($l++>10$)
\STATE $\theta_i^*\leftarrow argmax \{w(\hat{\theta}_i^j)\}_{j=1,2,\cdots}$//Obtain the best model hypothesis;
\ENDFOR
\STATE $\bm{\theta}\Leftarrow\{\theta_1^*,\theta_2^*,\ldots,\theta_m^*\}$ //Fuse all model hypotheses.
\ENSURE The generated model hypotheses $\bm{\theta}$.
\end{algorithmic}
\end{algorithm}

\subsection{Latent Semantic Consensus based Model Selection Algorithm}
\label{subsec:lsssa}
In this subsection, we propose a novel latent semantic consensus based model selection algorithm (LSC-MSA). To {mitigate} the sensitivity on the distribution of input data points, we analyze MH-LSS for model selection.
{In the space of generated model hypotheses, the disparities in proportions between different categories are not as pronounced as in the data point space. As illustrated in the example in Fig.~\ref{fig:firstexample} (b), the smallest and largest ratios for different categories of data points are $8.57\%$ and $57.14\%$, respectively, while the corresponding ratios for different categories of model hypotheses are $15.27\%$ and $37.72\%$, respectively.}

Firstly, we remove bad model hypotheses by the same manner as Eq.~(\ref{equ:outlierremove}). This is because, MH-LSS includes the similar character as DP-LSS, that the points corresponding to good model hypotheses are often far from the origin of coordinates while the other points are close to the origin. Then, we formulate the model selection problem as a simple subspace recovery problem in MH-LSS based on the remaining model hypotheses ${\bm\theta}_I$.

Note that, in MH-LSS, the points corresponding to the model hypotheses of the same model instances generally belong to the same one-dimensional subspace. We make some discussions for this claim. Given the preference value of a model hypothesis $\theta_i$, i.e. ${\bf P}_{\theta_i}=[p(s_1,\theta_i)~p(s_2,\theta_i)~\cdots~p(s_n,\theta_i)]^T$, and the corresponding $i$-th row of ${\bf V}$, i.e. ${\bf V'}_i=[v_{i1}~v_{i2}~\cdots~v_{i{k}}]$, we rewrite Eq.~(\ref{equ:lsaoriginal3}) as:
 \begin{align}
 \label{equ:lsaoriginal4}
   \begin{split}
 &[{\bf P}_{\theta_1}~\cdots~{\bf P}_{\theta_i}~\cdots~{\bf P}_{\theta_m}] {\bf P}\\
 &~~~~~~~~~~\approx[{\bf V'}_1~\cdots~{\bf V'}_i~\cdots~{\bf V'}_m]{\bf \Sigma}_{{k}\times {k}}^2{\bf V}^T.
  \end{split}
\end{align}
Then, we can obtain the following equation:
 \begin{align}
 \label{equ:lsaoriginal4_1}
   \begin{split}
 {\bf P}_{\theta_i}{\bf P}&\approx{\bf V'}_i \sigma_i{\bf \Sigma}{\bf V}^T.
  \end{split}
\end{align}
Finally, we obtain the coordinates ${\bf V'}_i \sigma_i$ of $\theta_i$ in MH-LSS as:
 \begin{align}
 \label{equ:lsaoriginal4_2}
   \begin{split}
   {\bf V'}_i \sigma_i&\approx {\bf P}_{\theta_i} {\bf P}{\bf V }{\bf \Sigma}^{-1}.
  \end{split}
\end{align}
We can see that, the coordinates of $\theta_i$ in MH-LSS depend on the preference value of $\theta_i$ to the input data points. {Notably,} the model hypotheses corresponding to the same model instance share similar preference to data points. {Consequently, the projected points in MH-LSS corresponding to these hypotheses will exhibit} the same subspace.

Then, we propose to cluster the remaining mapped points (${\hat{\bm\theta}}_I=\{\hat{\theta}_1,\hat{\theta}_2,\cdots,\hat{\theta}_{n_I}\}$) to several independent subspaces for subspace recovery, where each subspace corresponds to a model instance and each cluster of points belongs to the model instance. Theoretically, we can use a normal clustering method to cluster the remaining points in MH-LSS. However, we propose a simple but effective subspace recovery strategy to label the remaining points in MH-LSS. Specifically, we directly estimate several {origin-lines in MH-LSS (lines passing through $\bf O$), using them for data point clustering.

Our goal is to estimate $k$ origin-lines that best capture the remaining points in MH-LSS. Notably, the proposed clustering strategy simplifies the intricate task of segmenting high-dimensional data, reducing it to the straightforward process of clustering points by estimating origin-lines in LSS.}

For estimating an origin-line, it only requires to sample one point in MH-LSS to generate a model hypothesis of origin-line as the origin-line always passes through the origin of MH-LSS. To ensure with probability $\hat{\rho}$ that at least one all-inlier minimal subset is sampled, the number of minimal subsets $m_I$ required to sample is computed as:
 \begin{align}
m_I=\frac{\log (1-\hat{\rho})}{\log(1-(1-\hat{\alpha})^\eta)},
\end{align}
where $\hat{\alpha}$ denotes the ratio of gross outliers and $\eta$ denotes the number of data points in a minimal subset (here $\eta=1$ as only one point is needed to estimate an origin-line). In our case, the ratio of gross outliers (i.e. bad model hypotheses) is small since we have removed most of gross outliers. Thus, we only need sample a very small number of the minimal subsets for the origin-line estimation. For example, for $\hat{p}=0.99$ and $\hat{\alpha}=0.5$, $m_I$ is approximately equal to $7$.

\begin{figure}[ht]
\centering
\begin{minipage}{.35\textwidth}
\centerline{\includegraphics[width=1.0\textwidth]{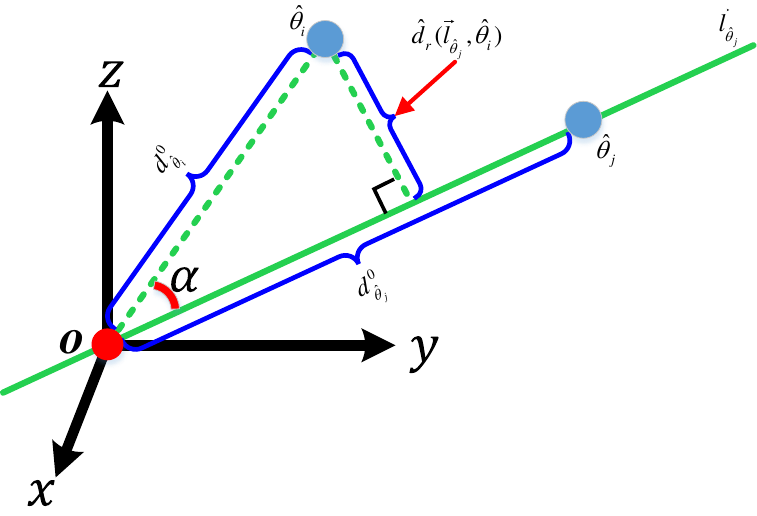}}
\end{minipage}
\caption{An example of computing the residual value between an origin-line $\vec{l}_{\hat{\theta}_j}$ and a point $\hat{\theta}_i$ in MH-LSS. The residual value is the Euclidean distance pointed by a red arrow. }
\label{fig:distancecompute}
\end{figure}

For the residual value between an origin-line $\vec{l}_{\hat{\theta}_j}$ (derived from a sampled point $\hat{\theta}_j$) and a point $\hat{\theta}_i$, as shown in Fig.~\ref{fig:distancecompute}, we can compute it as follows:
 \begin{align}
  \label{equ:residual}
  \begin{split}
\hat{d}_r(\vec{l}_{\hat{\theta}_j},\hat{\theta}_i)&=d_{\hat{\theta}_i}^o\sin\alpha,\\
~s.t.~
\sin \alpha&=\sqrt{1-(\frac{\langle \hat{\theta}_i, \hat{\theta}_j \rangle}{d_{\hat{\theta}_i}^o d_{\hat{\theta}_j}^o})^2 },
\end{split}
\end{align}
where $\langle\cdot,\cdot\rangle$ denotes the standard inner product.

After sampling a small number of model hypotheses, we adopt an integer linear program to estimate the $k$ origin-lines due to its effectiveness. The integer linear program is used to search for $k$ significant model hypotheses {that} cover the largest number of inliers of multiple model instances in data. For $n_I$ points in ${\bf \hat{\theta}}_I$ and $m_I$ model hypotheses $\vec{\bf L}_I$, we estimate the $k$ origin-lines via an integer linear program as~\cite{Magri_2016_CVPR}:
 \begin{align}
 \label{equ:ilp}
 \begin{split}
\max & \sum_{i=1}^{n_I} I_{\hat{\theta}_i},\\
s.t. & \sum_{j=1}^{m_I} I_{\vec{l}_{\hat{\theta}_j}}\leq k,\\
&\sum_{j: \vec{l}_{\hat{\theta}_j}\ni \hat{\theta}_i} I_{\vec{l}_{\hat{\theta}_j}}\geq I_{\hat{\theta}_i},~~ \forall \hat{x}_i \in{\bf \hat{\theta}}_I,
\end{split}
\end{align}
where $I_{\hat{\theta}_i}$ and $I_{\vec{l}_{\hat{\theta}_j}}$ denote the label of a point $\hat{\theta}_i$ and a model hypothesis $\vec{l}_{\hat{\theta}_j}$, respectively. If a point $\hat{\theta}_i$ (or a model hypothesis $\vec{l}_{\hat{\theta}_j}$) is selected to be an inlier of the returned model hypothesis, the label $I_{\hat{\theta}_i}=1$ (or $I_{\vec{l}_{\hat{\theta}_j}}=1$); Otherwise, $I_{\hat{\theta}_i}=0$ (or $I_{\vec{l}_{\hat{\theta}_j}}=0$). $\vec{l}_{\hat{\theta}_j}\ni \hat{\theta}_i$ represents that the point $\hat{\theta}_i$ belongs to the inliers of the model hypothesis $\vec{l}_{\hat{\theta}_j}$:
 \begin{align}
 \label{equ:threshold}
 \vec{l}_{\hat{\theta}_j}\ni \hat{\theta}_i \Longleftrightarrow  \hat{d}_r(\vec{l}_{\hat{\theta}_j},\hat{\theta}_i)\leq\beta,
  \end{align}
where $\beta$ is a threshold. We set $\beta$ to be a fixed value for different model fitting tasks, and the influence of its value on the proposed method will be discussed in Sec.~\ref{sec:parameteranalysis}. 

After that, we can obtain the labels of the remaining model hypotheses, and we select the model hypotheses with the largest weighting scores by Eq.~(\ref{equ:weighting}) for the same label as the final estimated model instances.

\subsection{The Complete Method and Analysis}
\label{sec:completemethod}
Based on the components described in the previous sections, we summarize the complete Latent Semantic Consensus based fitting method (called LSC) in Algorithm~\ref{alg:LSC}.
\begin{algorithm}[t] 
\small
\renewcommand{\algorithmicrequire}{\textbf{Input:}}
\renewcommand\algorithmicensure {\textbf{Output:} }
\caption{The proposed {LSC}} 
\label{alg:LSC} 
\begin{algorithmic}[1] 
\REQUIRE 
Data points $S=\{s_i\}_{i=1}^n$, the rank value ${k}$ and the threshold $\beta$.
\STATE Obtain model hypotheses $\bm{\theta}$ by Algorithm~\ref{alg:LSC-SA};
\STATE Construct a new latent semantic space (MH-LSS) for model hypotheses by Eq.~(\ref{equ:lsaoriginal3});
\STATE Remove the points corresponding to bad model hypotheses in MH-LSS by Eq.~(\ref{equ:outlierremove});
\STATE Sample model hypotheses of origin-line for each remaining points in MH-LSS;
\STATE Obtain labels of each remaining points in MH-LSS by Eq.~(\ref{equ:threshold});
\STATE Select the best model hypotheses from $\bm{\theta}$ corresponding to the points with the same labels by Eq.~(\ref{equ:weighting});
\STATE Estimate inliers for each estimated model instances.
\ENSURE The estimated model instances and the corresponding inliers.
\end{algorithmic}
\end{algorithm}

The proposed LSC method comprises a novel sampling algorithm and a new model selection algorithm. In essence, LSC not only analyzes the distributions of points in DP-LSS to generate high-quality model hypotheses but also exploits MH-LSS to accurately estimate model instances. Importantly, LSC involves deterministic nature, ensuring consistent and reliable solutions for model fitting.

For the computational complexity of LSC, in Algorithm~\ref{alg:LSC-SA}, the neighborhood construction and the latent semantic space constructing take the main time. For the neighborhood construction, the time complexity is close to $O(n \log n)$.
For constructing the latent semantic space (i.e. Step 2), we use truncated SVD where the time complexity depends on the number of singular values retained after truncation. Thus, the computational complexity is approximately $O(n^2*k)$, where ${k}$ is the dimension of MH-LSS.
Step $3$-$7$ is based on the results of remaining model hypotheses (recall that we only deal with a very small number of high-quality model hypotheses), and their time complexity is close to $O(n)$. Therefore, the total complexity of LSC approximately amounts to $O(n^2)$.
\section{Experiments}
\label{sec:experiments}
In this section, we investigate the performance of the proposed LSC fitting scheme on synthetic data and real images for model fitting tasks, including line fitting, circle fitting, homograhy/fundamental matrix estimation and motion segmentation.
All experiments are run on MS Windows $10$ with Intel Core i$7$-$8565$ CPU $1.8GHz$ and $16GB$ RAM in Matlab.

\subsection{Dataset Setting and Evaluation Metrics}
For synthetic data, we use all four datasets from \cite{xiao2021segmentation} for line fitting, and the datasets from \cite{wang2019} for circle fitting.

For real images, we test all competing methods for single- and multiple-structural data. For multiple-structural data, we adopt all $38$ image pairs from the AdelaideRMF dataset~\cite{wong2011dynamic} for homography and fundamental matrix estimation and all $155$ videos from $Hopkins$$155$~\cite{tron2007} for motion segmentation. For single-structural data, we uses $32,950$ image pairs to construct a dataset (called MS-COCO-F) from the MS-COCO images~\cite{lin2014microsoft} for homography estimation and $12,455$ image pairs to construct a dataset (called YFCC100M-F) from the YFCC100M dataset~\cite{thomee2016yfcc100m} for the fundamental matrix estimation task.

To evaluate the performance of a model fitting method,  we follow most of model fitting methods to compute the segmentation error (SE) as follows~\cite{tennakoon2016robust}:
\begin{align}
\label{equ:fittingerror}
SE=\frac{\#~mislabeled~data~points}{\#~data~points}\times 100.
\end{align}
\subsection{Parameter Analysis and Settings}
\label{sec:parameteranalysis}
There are two parameters for LSC, i.e., the threshold $\psi$ in Eq.~(\ref{equ:preferencefunction}) and the threshold $\beta$ in Eq.~(\ref{equ:threshold}). To verify the influence of different values of parameters on the performance of LSC, we randomly select $1,000$ image pairs from the MS-COCO-F dataset for single-structural homography estimation, and $1,000$ image pairs from the YFCC100M-F dataset for single-structural fundamental matrix estimation, and all image pairs from the AdelaideRMF dataset for multiple-structural tasks.

\begin{figure}[ht]
\centering
\begin{minipage}{.23\textwidth}
\centerline{\includegraphics[width=1.0\textwidth]{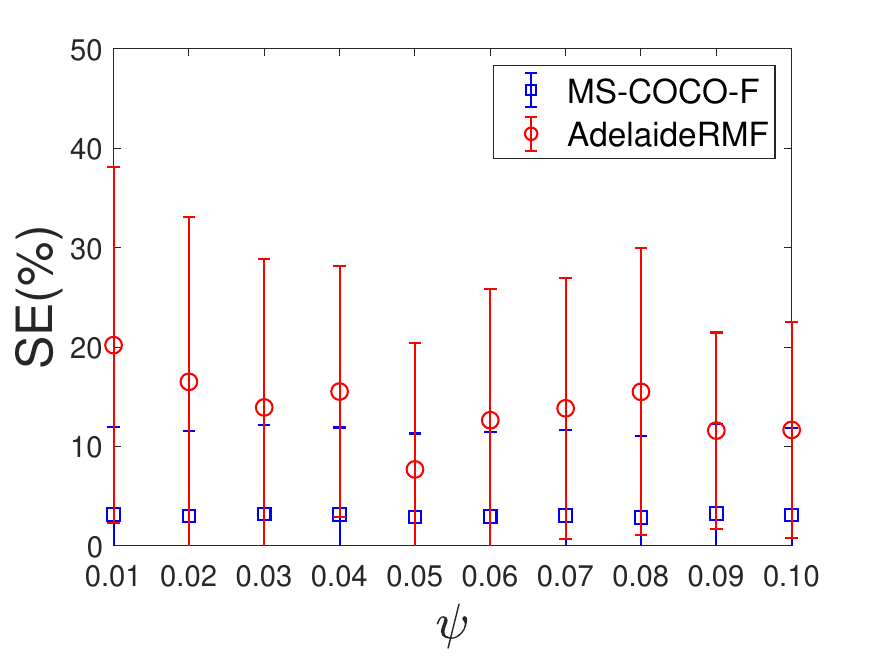}}
\centerline{\includegraphics[width=1.0\textwidth]{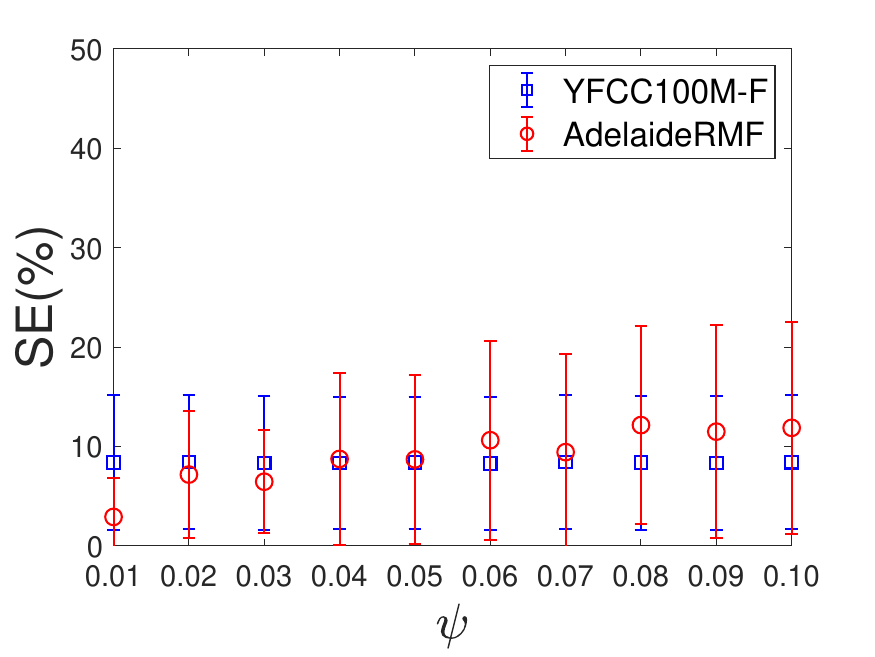}}
\end{minipage}
\begin{minipage}{.23\textwidth}
\centerline{\includegraphics[width=1.0\textwidth]{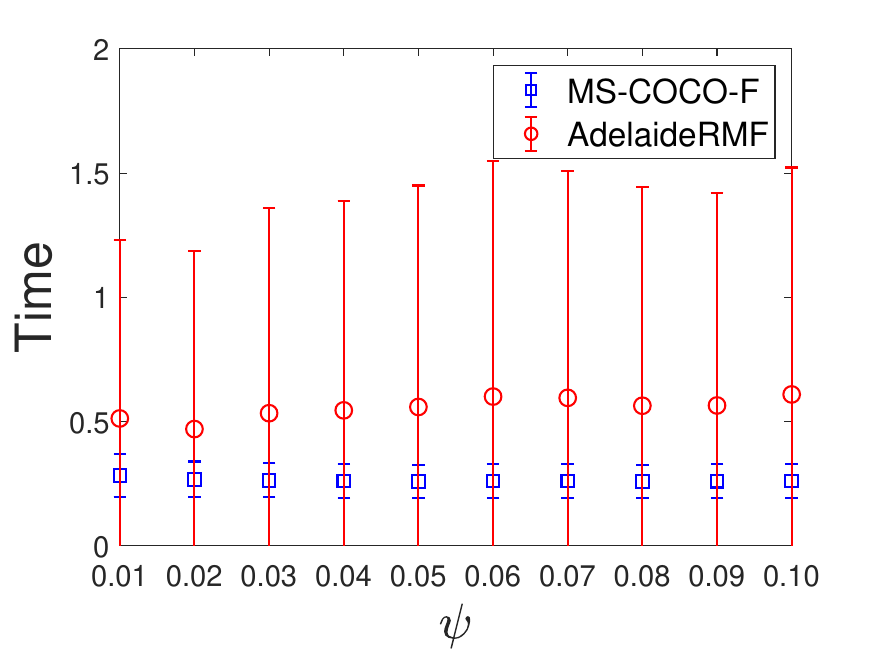}}
\centerline{\includegraphics[width=1.0\textwidth]{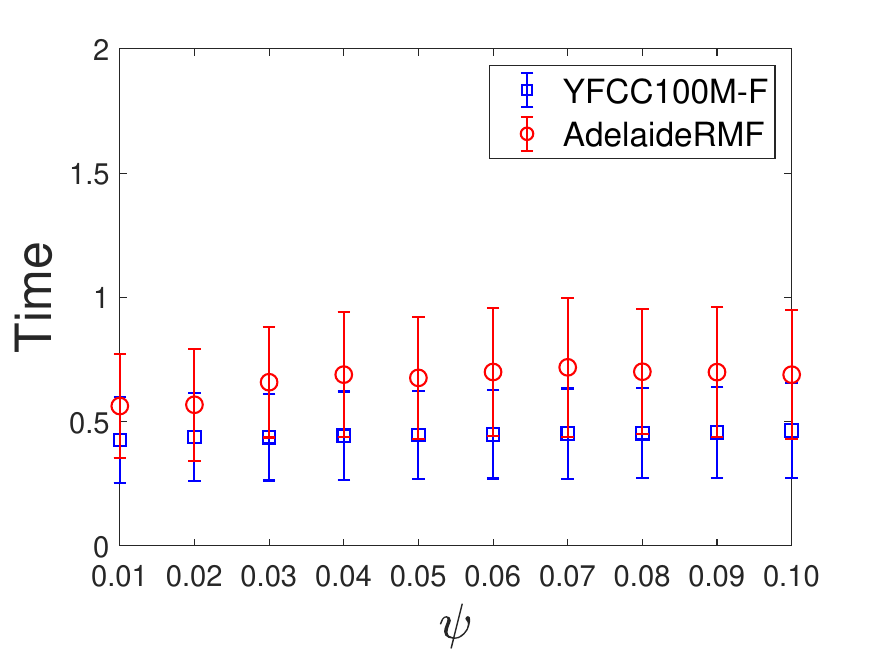}}
\end{minipage}
\caption{Quantitative results obtained by the proposed LSC fitting scheme with different values of $\psi$ for the task of (Top) homography estimation and (Bottom) fundamental matrix estimation on the MS-COCO-F, YFCC100M-F and AdelaideRMF datasets.}
\label{fig:fittingparameter}
\end{figure}

For the threshold $\psi$, we report the quantitative results (the mean and standard deviation values) of SE and the CPU Time obtained by LSC with different values of $\psi$ ($0.01$-$0.10$) for the task of homography and fundamental matrix estimation in Fig.~\ref{fig:fittingparameter}. {Note that, the mean and standard deviation values are obtained by LSC on different image pairs for each $\psi$.}
We can see that, for homography estimation, LSC achieves stable values of SE with different values of $\psi$ on the MS-COCO-F dataset, but it obtains different values of SE on the AdelaideRMF datasets due to the influence of multiple structures. The CPU time used by LSC has not obvious change with different values of $\psi$ on the both datasets. We also find that LSC obtains the lowest values of SE when $\psi=0.05$ on the AdelaideRMF datasets. Thus, we set $\psi=0.05$ for homography estimation in the following experiments.
For fundamental matrix estimation, LSC also achieves the same performance on the datasets with single and multiple structures as homography estimation, and it achieves the lowest values of SE when $\psi=0.01$ on the AdelaideRMF datasets. Thus, we set $\psi=0.01$ for fundamental matrix estimation in the following experiments. For the other tasks, we do not change the values of $\psi$ and set $\psi=0.01$.

\begin{figure}[ht]
\centering
\begin{minipage}{.23\textwidth}
\centerline{\includegraphics[width=1.0\textwidth]{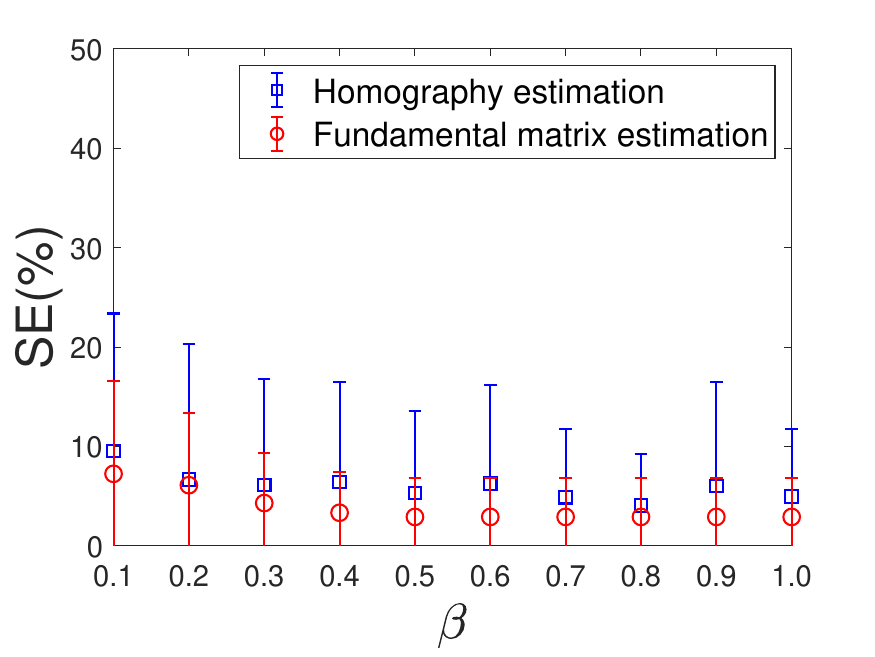}}
\end{minipage}
\begin{minipage}{.23\textwidth}
\centerline{\includegraphics[width=1.0\textwidth]{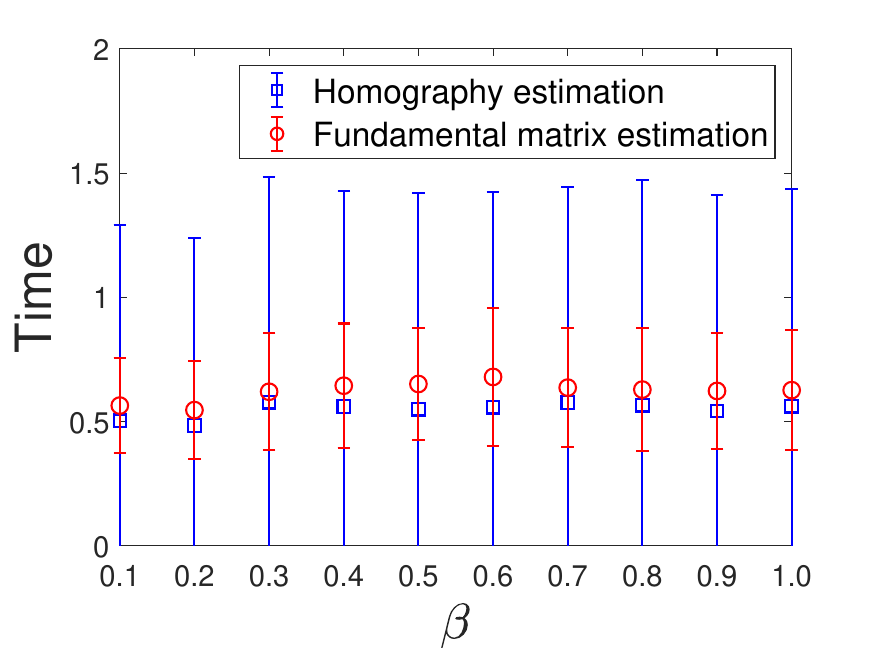}}
\end{minipage}
\caption{Quantitative results obtained by the proposed LSC fitting scheme with different values of $\beta$ for the task of homography estimation and fundamental matrix estimation on the AdelaideRMF dataset.}
\label{fig:fittingparameter2}
\end{figure}
For the threshold $\beta$, we report the quantitative results (the mean and standard deviation values) of SE and the CPU Time obtained by LSC with different values of $\beta$ ($0.1$-$1.0$) for the task of homography and fundamental matrix estimation in Fig.~\ref{fig:fittingparameter2}. Note that, we only test this experiment on the AdelaideRMF datasets since $\beta$ only affects the data with multiple structures. Recall that, we perform the model selection in the latent semantic space and regress all model fitting tasks to the subspace recovery problem, thus, we can set the same values of $\beta$ for all model fitting tasks. From Fig.~\ref{fig:fittingparameter2}, LSC obtains the lowest values of SE for homography and fundamental matrix estimation when $\beta=0.8$. Thus, in the following experiments, we set $\beta=0.8$.
\subsection{Results on Synthetic Data}
\label{sec:syntheticData}
In this section, we compare the proposed LSC with several state-of-the-art model fitting methods, including MSHF~\cite{wang2019}, RansaCov~\cite{Magri_2016_CVPR}, T-linkage~\cite{Magri_2014_CVPR}, RPA~\cite{magri2017multiple}, Multi-link~\cite{Magri_2021_CVPR}, CBG~\cite{lin2022co} and CLSA~\cite{xiao2021segmentation}, for line and circle fitting on synthetic data.
\begin{figure}[t]
\centering
\begin{minipage}[t]{.115\textwidth}
  \centerline{\includegraphics[width=1.00\textwidth]{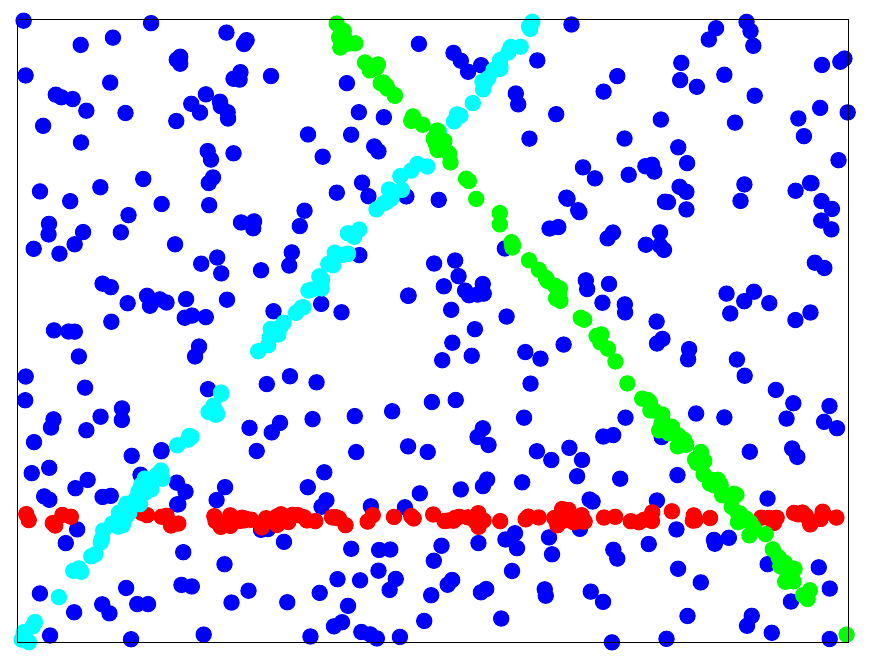}}
  \centerline{\includegraphics[width=1.00\textwidth]{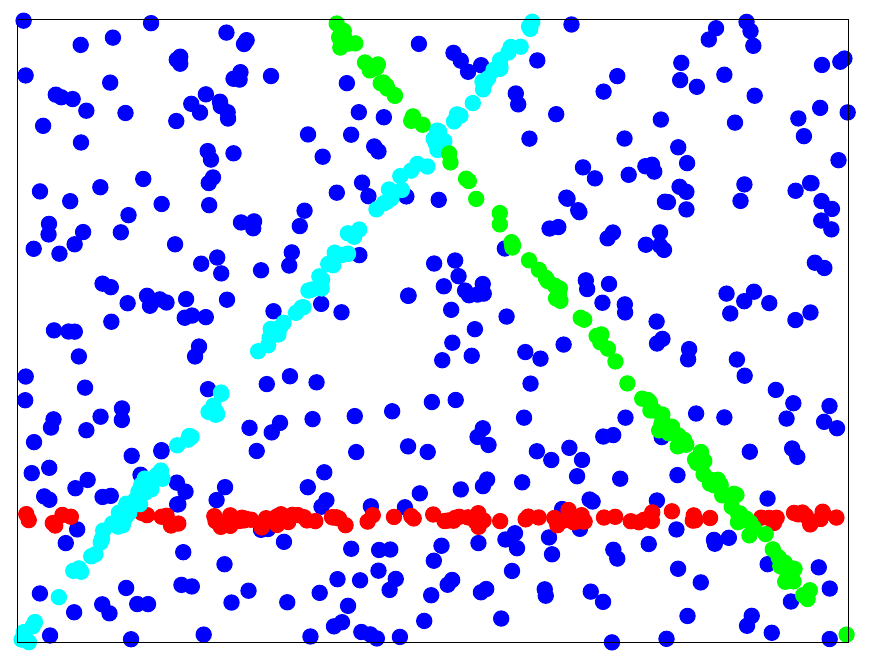}}
  \centerline{\footnotesize(a) 3 lines }
\end{minipage}
\begin{minipage}[t]{.115\textwidth}
  \centerline{\includegraphics[width=1.000\textwidth]{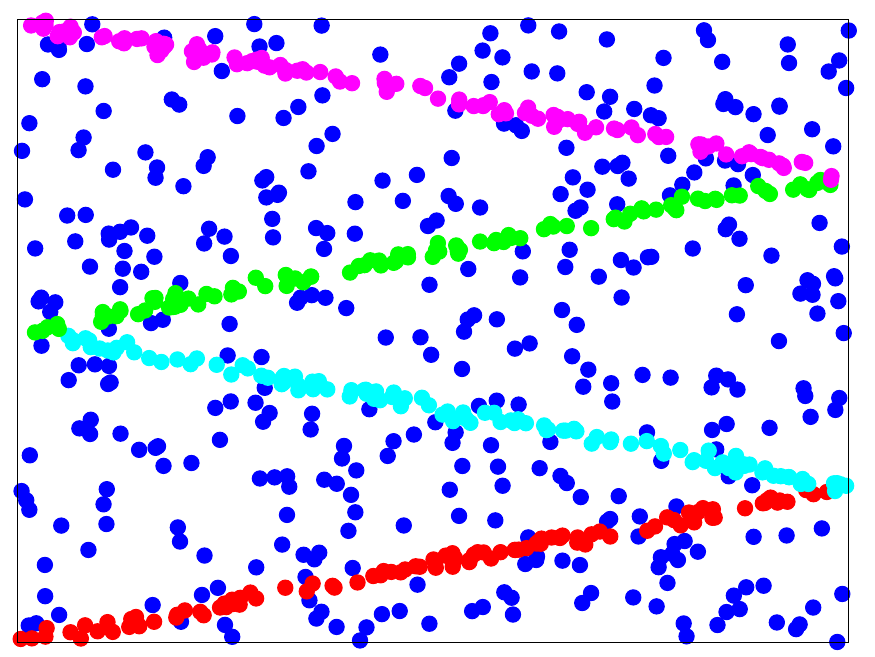}}
  \centerline{\includegraphics[width=1.000\textwidth]{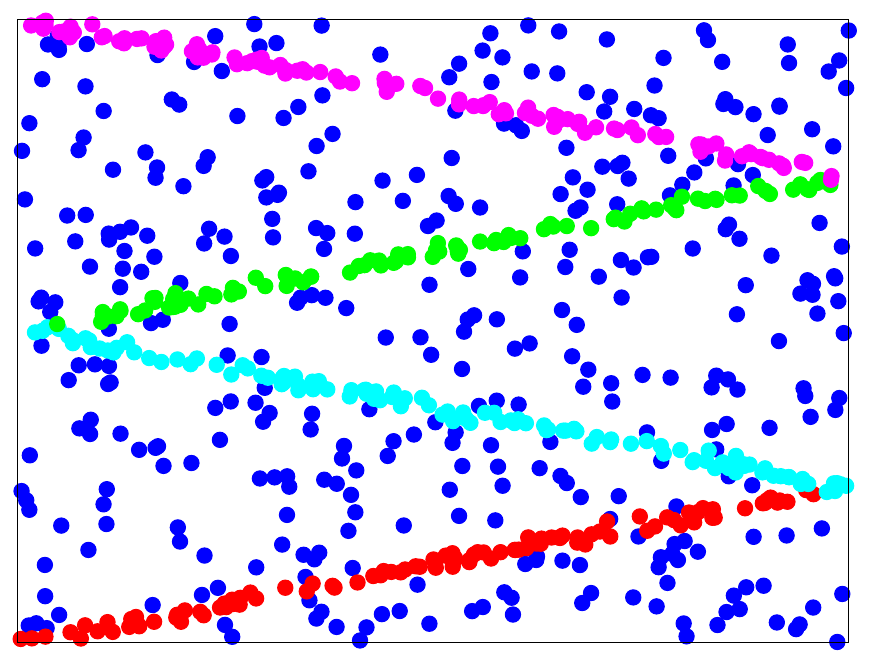}}
  \centerline{\footnotesize(b) 4 lines }
\end{minipage}
\begin{minipage}[t]{.115\textwidth}
  \centerline{\includegraphics[width=1.000\textwidth]{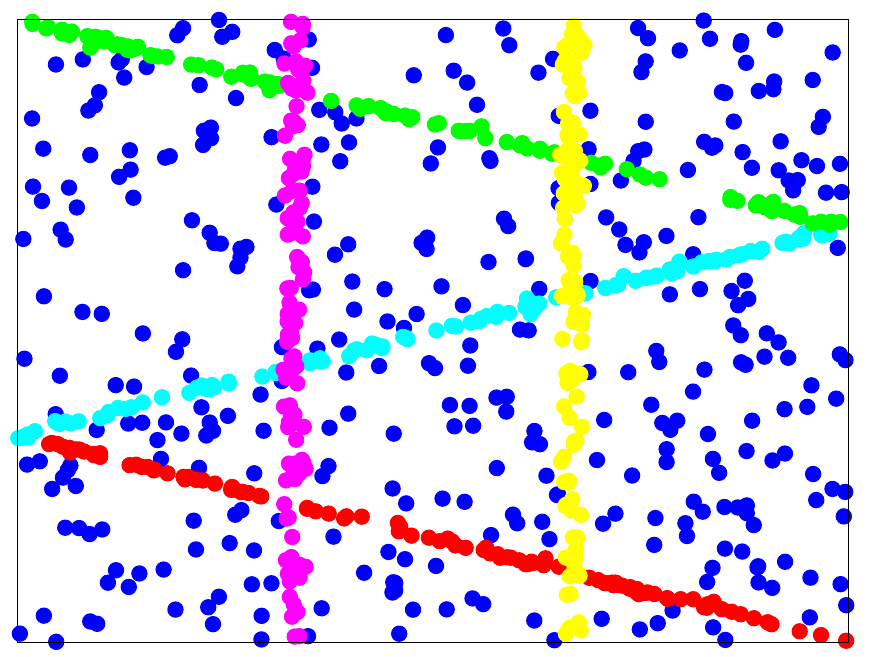}}
  \centerline{\includegraphics[width=1.000\textwidth]{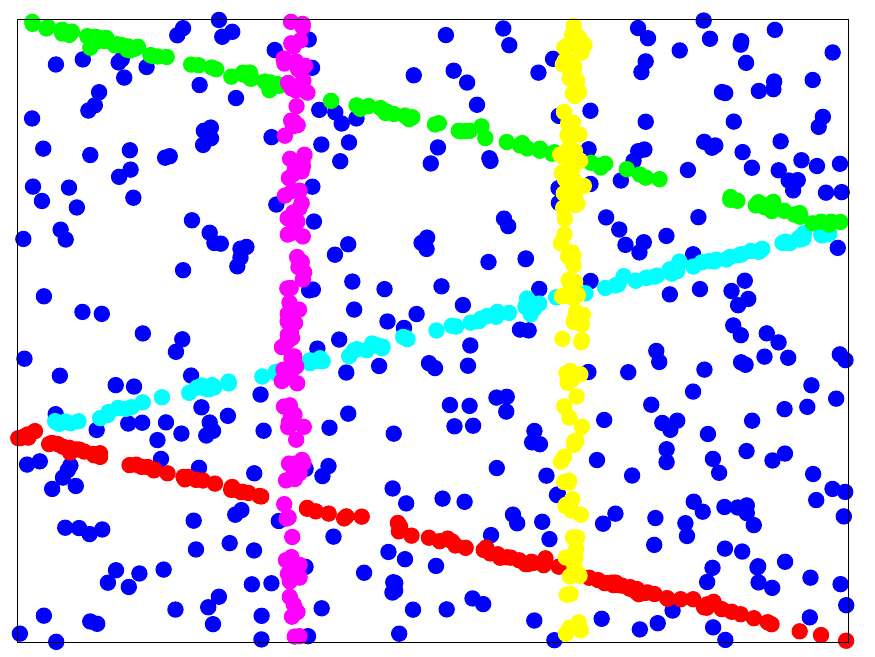}}
 \centerline{\footnotesize(c) 5 lines }
\end{minipage}
\begin{minipage}[t]{.115\textwidth}
  \centerline{\includegraphics[width=1.000\textwidth]{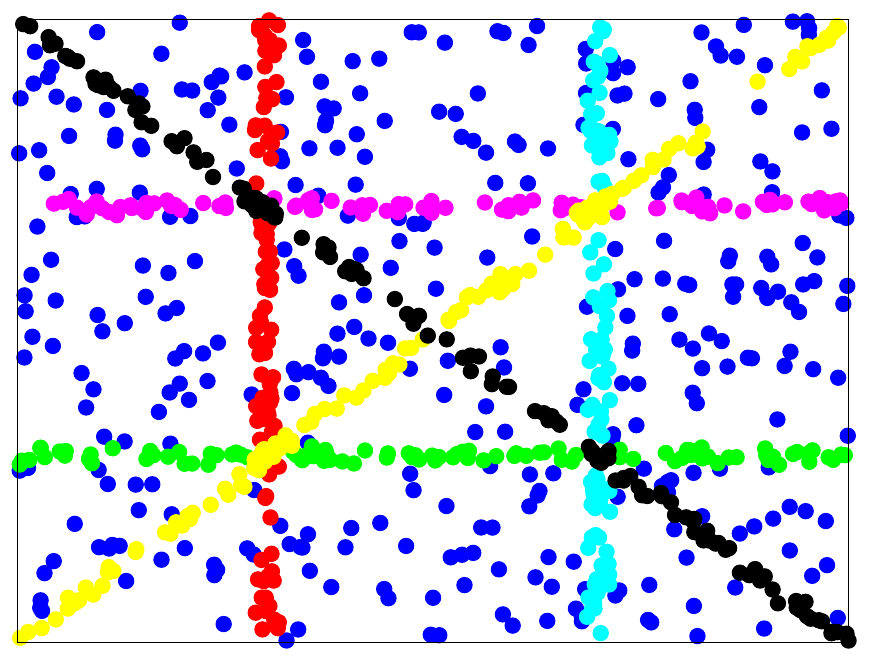}}
  \centerline{\includegraphics[width=1.000\textwidth]{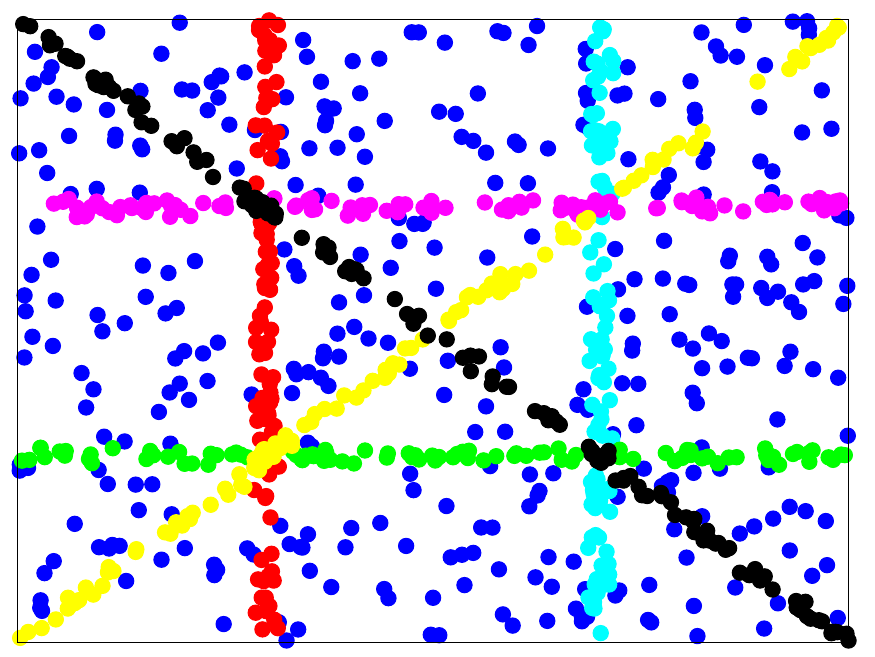}}
  \centerline{\footnotesize(d) 6 lines }
\end{minipage}
\caption{The results obtained by the proposed methods (LSC) for line fitting. The first row of each sub-figure shows the ground truth results and the second row shows the fitting results obtained by LSC. The inlier noise scale is $1.5$ and each line includes $100$ inliers. The outliers are labeled in blue, and the inliers of each estimated model instance are labeled in different colors (the following figures have the same settings). }
\label{fig:lines}
\end{figure}
\begin{table}[t]
\small
\centering
  \caption{Quantitative comparison results obtained by the eight competing methods for line fitting on four synthetic data. The best results of SE ($\%$) and Time (in seconds) are boldfaced. }
\scalebox{0.8}{{\tabcolsep0.02in
\begin{tabular}{cccccccccc}
\toprule
Data          &   & MSHF&RansaCov&T-linkage&RPA&Multi-link&CBG&CLSA&LSC\\
\midrule
         \multirow{3}{*}{ 3  lines }&$Avg.$    & {4.74}& {3.25}& {3.42}&4.62&3.70&2.83&{2.21}& {\bf1.00}\\
         &$Std.$   &{2.89}&{0.09}& {1.17}&4.15& {1.10}& 0.19&{1.08}&-\\
            &$Time$   &{1.96}&{2.88}& {88.94}& {175.31}&2.26&2.85& {1.09}&{\bf0.86} \\
          \rowcolor{mygray1}

     &$Avg.$       & {6.31}& {2.72}& {9.97}& {2.77}&6.42&{3.12}&{\bf2.28}&2.63\\
     \rowcolor{mygray1}
      &$Std.$       & {4.87}& {0.81}& {8.19}& {0.63}&6.75&{0.32}&{0.49}&-\\
     \rowcolor{mygray1}
         \multirow{-3}{*}{4  lines }&$Time$      &{2.12}&{3.46}& {119.12}& {222.11}&2.49&{3.40}&{1.23}&{\bf0.91}\\

   \multirow{3}{*}{ 5  lines }&$Avg.$     & { 2.71}& {2.50}& {10.67}& {3.16}&2.94&{5.33}&{2.41}&{\bf1.33}\\
   &$Std.$     & {1.18}& {0.15}& {11.18}& {1.76}&3.85&0.53&{1.38}&-\\
           &$Time$     &{2.11}&{3.51}& {154.61}& {360.55}&2.84&3.10& {1.42}&{\bf1.23}\\
              \rowcolor{mygray1}
  &$Avg.$    & {10.78}& {8.31}& {35.17}& {40.60}&18.02&{6.76}&{4.62}&{\bf3.70} \\
        \rowcolor{mygray1}
     &$Std.$    & {3.89}&1.68& {9.92}& {8.54}&7.55&{0.42}&{2.14}&-\\

       \rowcolor{mygray1}
        \multirow{-3}{*}{6  lines }&$Time$   &{2.16}&{4.37}& {193.74}& {430.03}& 3.22&{3.84}& {1.64}&{\bf1.35}\\
    \bottomrule\\
\end{tabular}}}
 \label{table:2Dlinetable}
\end{table}

\subsubsection{Line Fitting}
\label{sec:sylinefitting}
We evaluate the performance of the eight fitting methods for line fitting on four challenging synthetic datasets ({see Fig.~\ref{fig:lines}}). We repeat each experiment $50$ times except for our proposed LSC due to its deterministic nature. We report the mean and standard deviation of SE obtained by the eight fitting methods, and the average CPU time (in seconds) used by the competing methods in Table~\ref{table:2Dlinetable}.

{In} Fig.~\ref{fig:lines} and Table~\ref{table:2Dlinetable}, most of competing methods are able to achieve good fitting performance, and the proposed LSC achieves the lowest SE and it also is the fastest one among all eight competing methods. The four synthetic datasets involve different challenges, e.g., the ``$3$ line" data includes cross data points, the ``$4$ line" data includes data points with different distributions, the ``$5$ line" data includes the both challenges and the ``$6$ line" data is most challenging due to its complex distributions. All eight competing methods achieve low SE on the top two data, and they also achieve the similar performance on the ``$5$ line" data expect for T-linkage. This is because that T-linkage automatically estimates the number of model instances in data, and it is hard to do that for complex distributions. For the ``$6$ line" data, the values of SE obtained by only CLSA and LSC are lower than $5\%$.

For the standard deviation performance, the proposed LSC show significant superiority over the other competing methods due to its deterministic nature. For the CPU time, LSC also has advantages since it only handles a small number of high-quality model hypotheses.
\subsubsection{Circle Fitting}
\label{sec:sycirclefitting}
\begin{figure}[t]
\centering
\begin{minipage}[t]{.115\textwidth}
  \centerline{\includegraphics[width=1.00\textwidth]{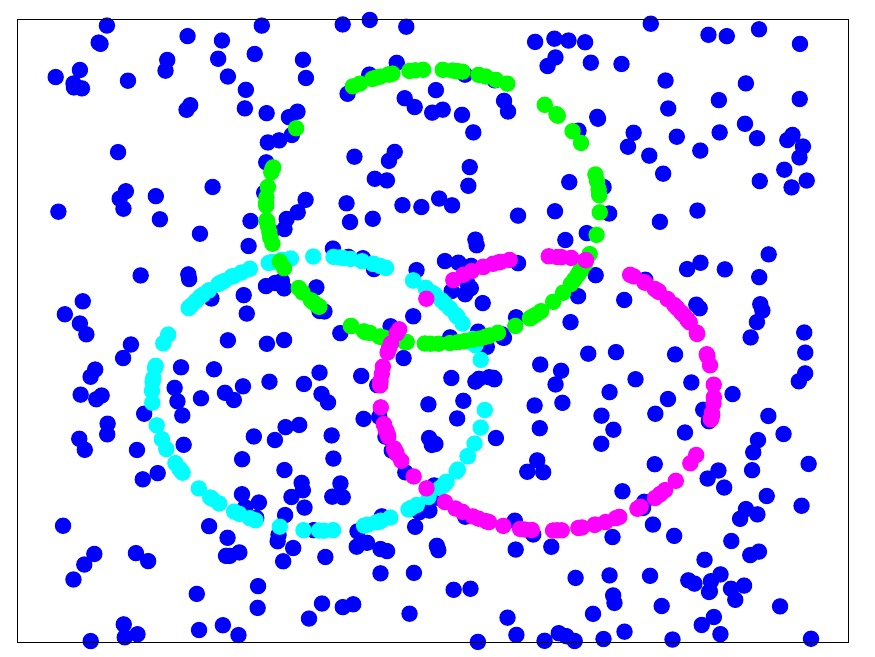}}
  \centerline{\includegraphics[width=1.00\textwidth]{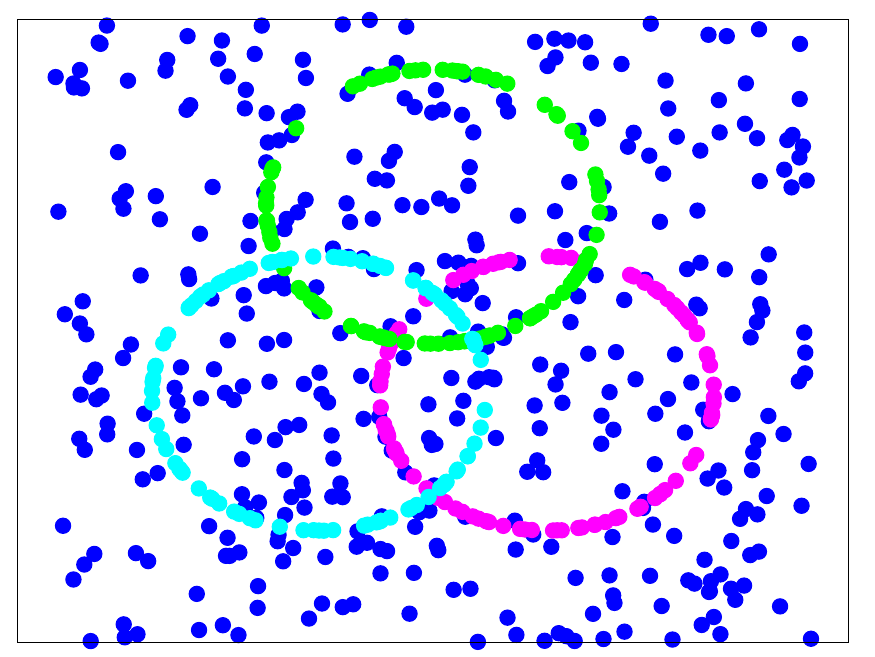}}
  \centerline{\footnotesize(a) 3 circles }
\end{minipage}
\begin{minipage}[t]{.115\textwidth}
  \centerline{\includegraphics[width=1.000\textwidth]{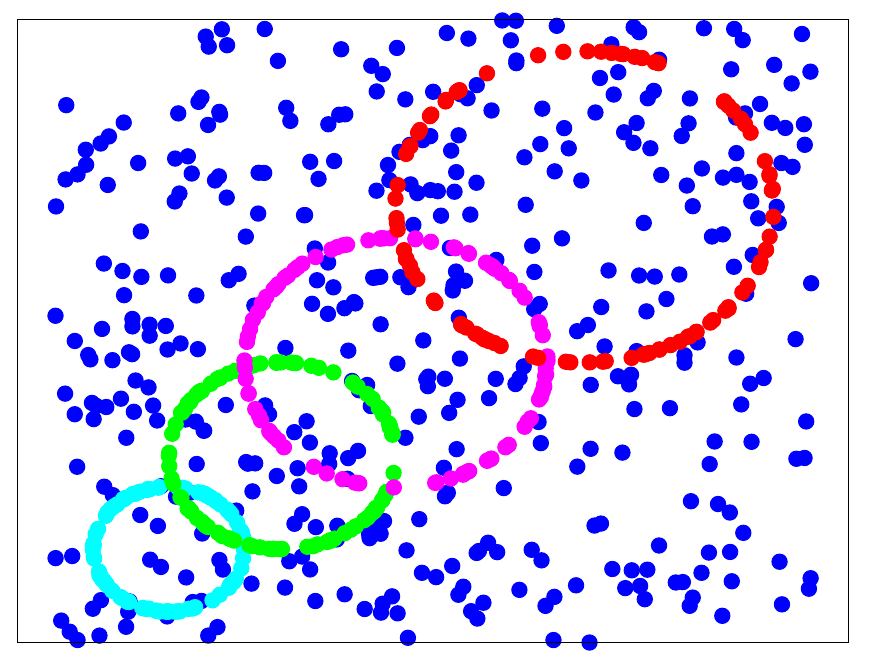}}
  \centerline{\includegraphics[width=1.000\textwidth]{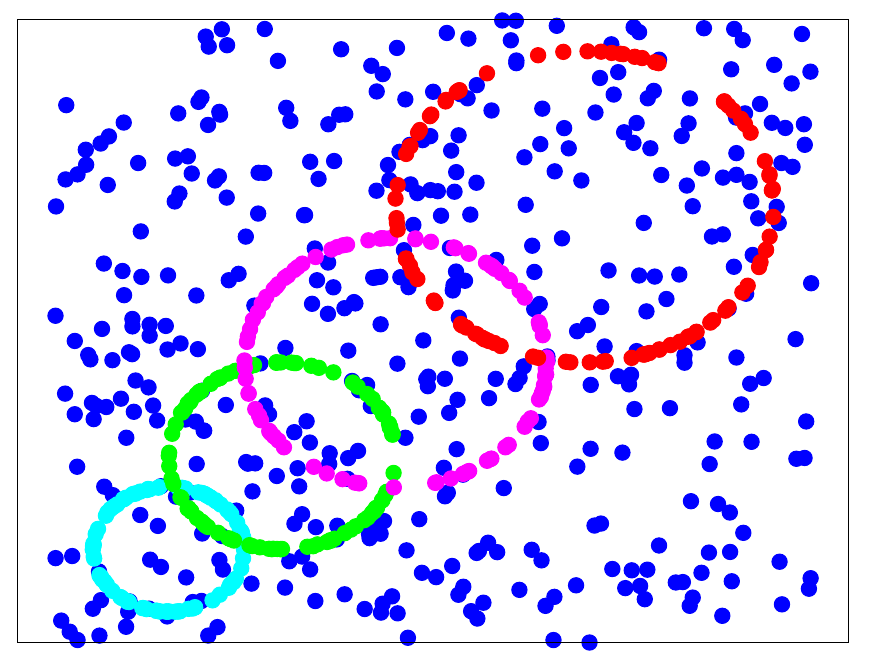}}
  \centerline{\footnotesize(b) 4 circles }
\end{minipage}
\begin{minipage}[t]{.115\textwidth}
  \centerline{\includegraphics[width=1.000\textwidth]{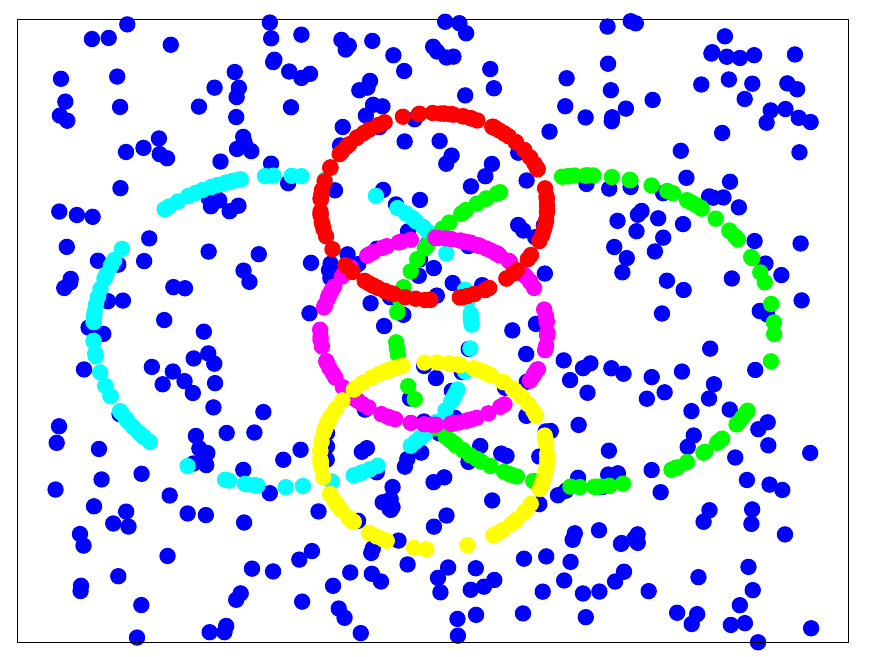}}
  \centerline{\includegraphics[width=1.000\textwidth]{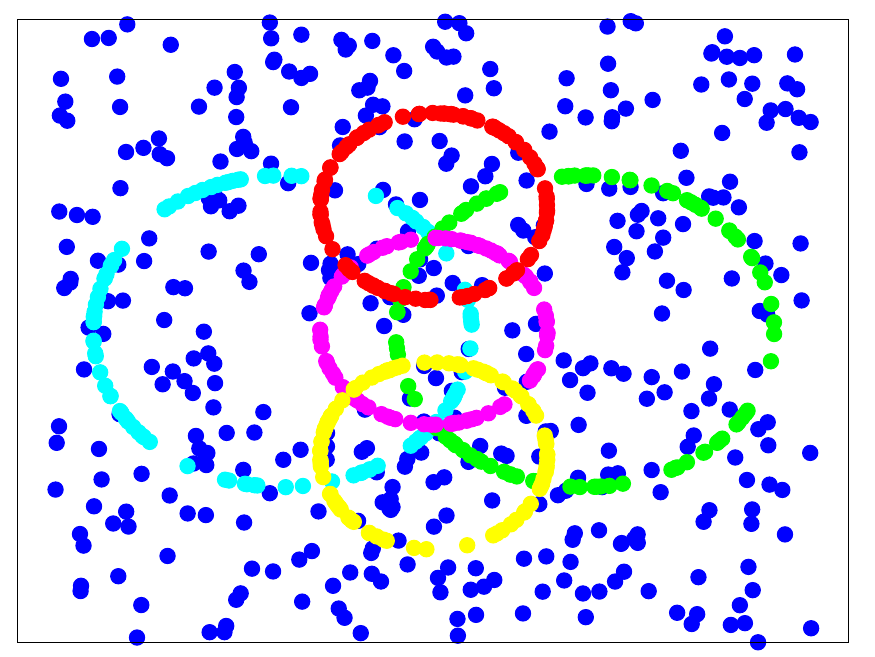}}
 \centerline{\footnotesize (c) 5 circles }
\end{minipage}
\begin{minipage}[t]{.115\textwidth}
  \centerline{\includegraphics[width=1.000\textwidth]{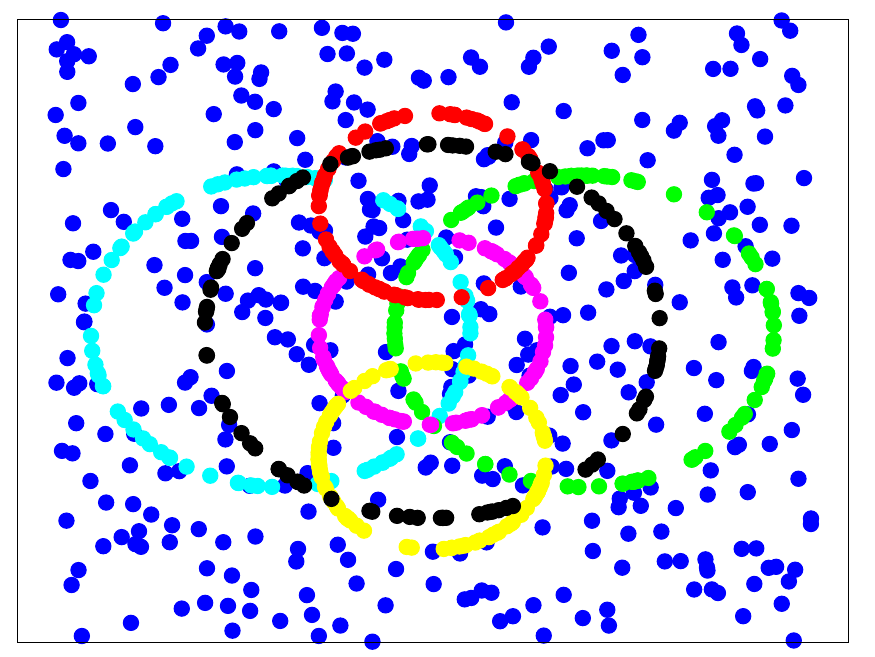}}
  \centerline{\includegraphics[width=1.000\textwidth]{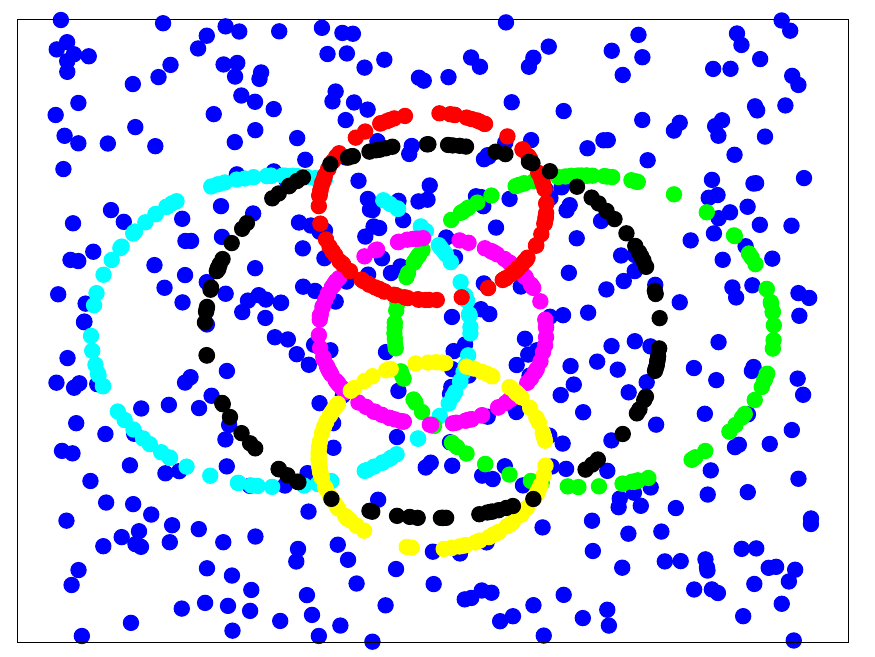}}
  \centerline{\footnotesize (d) 6 circles }
\end{minipage}
\caption{The results obtained by the proposed methods (LSC) for circle fitting. The first row of each sub-figure shows the ground truth results and the second row shows the fitting results obtained by LSC. The inlier noise scale is $0.1$ and each circle has $100$ inliers. }
\label{fig:circles}
\end{figure}
\begin{table}[t]
\small
\centering
  \caption{Quantitative comparison results obtained by the eight competing methods for circle fitting on four synthetic data.}
\scalebox{0.78}{{\tabcolsep0.02in
\begin{tabular}{cccccccccc}
\toprule

Data          &   & MSHF&RansaCov&T-linkage&RPA&Multi-link&CBG&CLSA&LSC\\
\midrule
         \multirow{3}{*}{ 3 circles }&$Avg.$    & {16.84}& {\bf0.52}& {1.53}&0.58&18.714&1.95&{8.35}& {0.71}\\
         &$Std.$   &{8.59}&{0.07}& {3.91}&0.08& {8.179}&0.52&{3.06}&-\\
            &$Time$   &{1.66}&{2.56}& {58.44}& {45.46}&2.26&7.80& {1.37}&{\bf0.63} \\
          \rowcolor{mygray1}
     &$Avg.$       & {20.55}& {2.67}& {21.08}& {5.80}&8.85&{1.61}&{38.55}&{\bf1.25}\\
     \rowcolor{mygray1}
      &$Std.$       & {9.53}& {0.65}& {8.96}& {6.76}&5.92&{0.10}&{2.77}&-\\
     \rowcolor{mygray1}
         \multirow{-3}{*}{4 circles }&$Time$      &{2.10}&{3.35}& {89.42}& {74.67}&2.24&{3.40}&{5.85}&{\bf0.76}\\
   \multirow{3}{*}{ 5  circles }&$Avg.$     & {10.64}& {\bf0.23}& {7.95}& {0.31}&25.84&{1.17}&{10.62}&{0.33}\\
   &$Std.$     & {7.27}& {0.03}& {7.47}& {0.10}&6.29&0.13&{7.31}&-\\
           &$Time$     &{1.91}&{2.96}& {100.86}& {85.12}&2.67&5.60& {1.55}&{\bf0.98}\\
              \rowcolor{mygray1}
  &$Avg.$    & {17.81}& {1.99}& {19.47}& {6.68}&30.32&{1.94}&{31.55}&{\bf1.00} \\
        \rowcolor{mygray1}
     &$Std.$    & {6.12}&2.21& {10.85}& {5.12}&3.94&{0.17}&{7.58}&-\\

       \rowcolor{mygray1}
        \multirow{-3}{*}{6 circles }&$Time$   &{2.07}&{3.30}& {129.12}& {117.40}&3.49&{5.87}& {1.68}&{\bf1.11}\\
    \bottomrule\\
\end{tabular}}}
 \label{table:2Dcircletable}
\end{table}

We perform the same setting as line fitting for circle fitting ({see Fig.~\ref{fig:circles}}), and report quantitative comparison results obtained by the eight fitting methods in Table~\ref{table:2Dcircletable}.

From Fig.~\ref{fig:circles} and Table~\ref{table:2Dcircletable}, we can see that circle fitting is more challenging that line fitting, and only RansaCov, RPA, CBG and LSC are able to achieve values of SE on all the four datasets. Note that, CLSA is able to achieve good performance for line fitting but bad performance for circle fitting because the generated model hypotheses include more bad ones and CLSA is a model selection method and its performance depends on the quality of model hypotheses. In contrast, LSC not only generates high-quality model hypotheses, but also effectively estimates model instances in data.  For the standard deviation performance and the CPU time, LSC shows the same advantages as line fitting for circle fitting.

\subsection{Results on Real Images}
In this subsection, we evaluate the performance of LSC on all $38$ image pairs from the AdelaideRMF dataset with data for two popular fitting tasks, i.e., homography and fundamental matrix estimation. For the competing methods, we add more fitting methods, i.e., RCMSA~\cite{pham2014random}, Prog-X~\cite{barath2019progressive},  SDF~\cite{IJCVXiao2019} and LGF~\cite{xiao2020deterministic}, since they can deal with two-view fitting methods. To show the effectiveness of the proposed LSC, we also test it on two large datasets (i.e., the MS-COCO-F and YFCC100M-F datasets). We compare it with two deterministic fitting methods (i.e., SDF and LGF), and we run RANSAC as a baseline. In addition, we run two deep learning based methods (i.e., OANet~\cite{zhang2019learning} and ConvMatch~\cite{zhang2023convmatch}).
\subsubsection{Homograhy Estimation}
\label{sec:homography}
\begin{figure}[t]
\centering
\begin{minipage}{.156\textwidth}
\centerline{\includegraphics[width=1.0\textwidth]{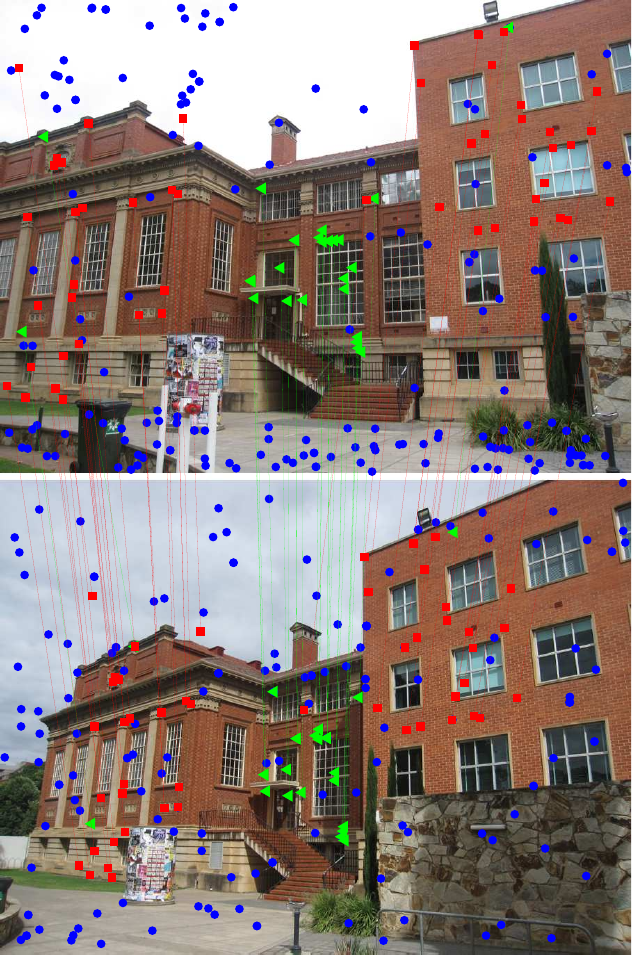}}
 \centerline{\footnotesize (a) Barrsmith}
   \centerline{\includegraphics[width=1.0\textwidth]{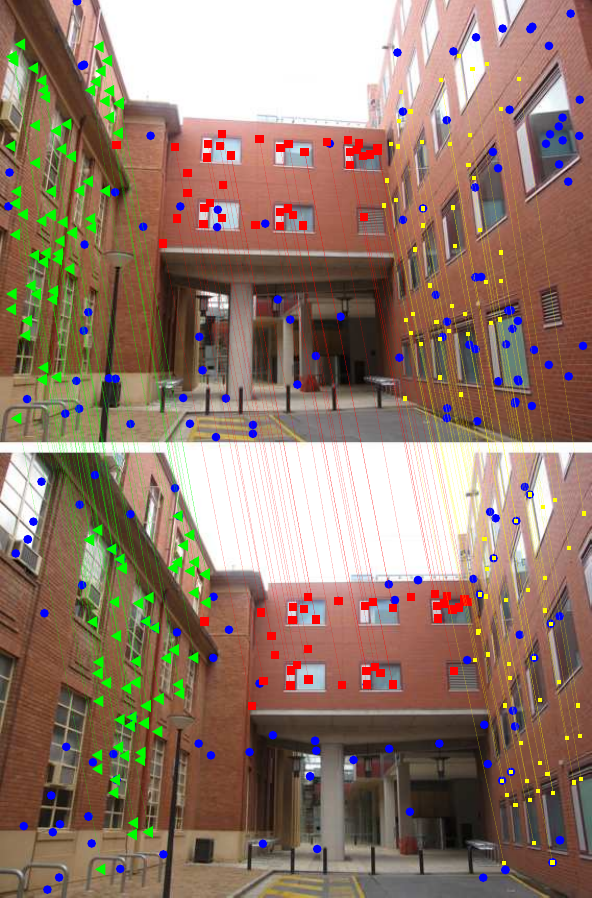}}
 \centerline{\footnotesize(d) Neem}

\end{minipage}
\begin{minipage}{.156\textwidth}
 \centerline{\includegraphics[width=1.0\textwidth]{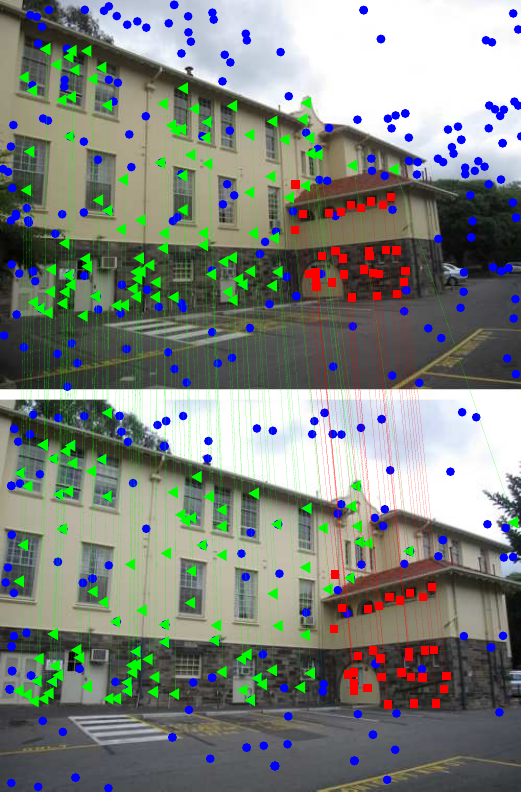}}
 \centerline{\footnotesize(b) Hartley}
   \centerline{\includegraphics[width=1.0\textwidth]{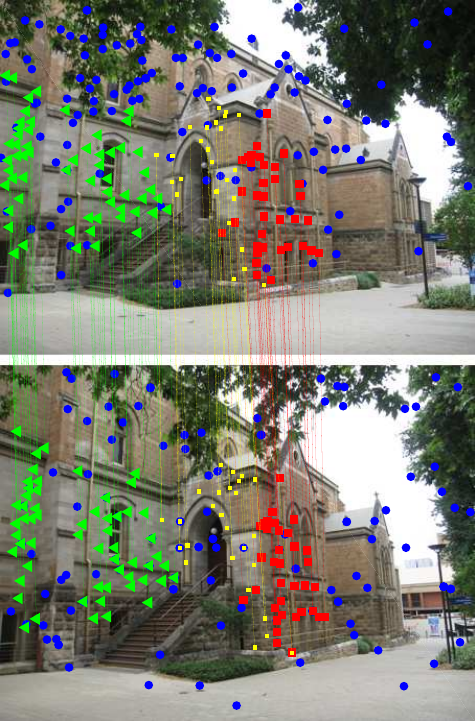}}
 \centerline{\footnotesize(e) Elderhallb}

\end{minipage}
\begin{minipage}{.156\textwidth}
\centerline{\includegraphics[width=1.0\textwidth]{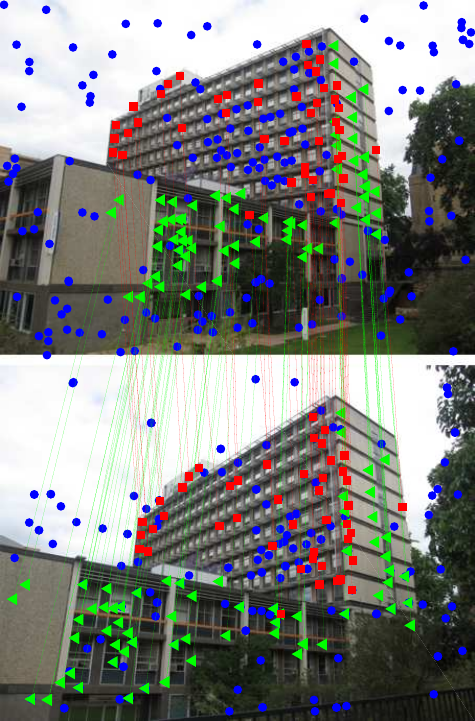}}
 \centerline{\footnotesize (c) Napiera}
   \centerline{\includegraphics[width=1.0\textwidth]{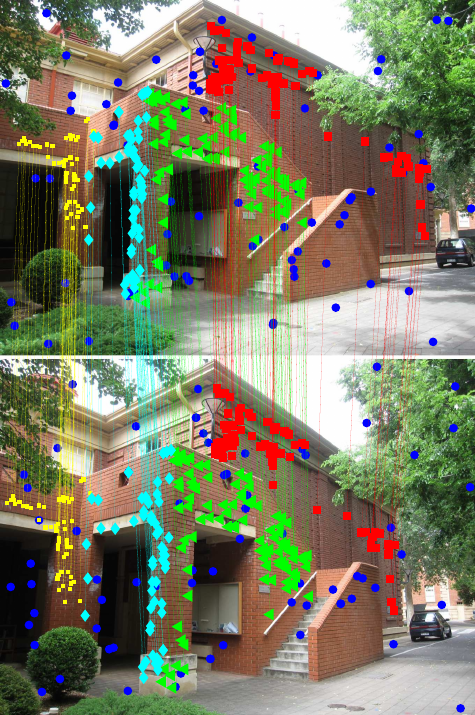}}
 \centerline{\footnotesize(f) Johnsona}
\end{minipage}
\caption{Some challenging multiple-structural cases that LSC successfully estimates model instances for homography estimation.}
\label{fig:examples2H}
\end{figure}
\begin{figure}[t]
\centering
\begin{minipage}{.24\textwidth}
\centerline{\includegraphics[width=1.0\textwidth]{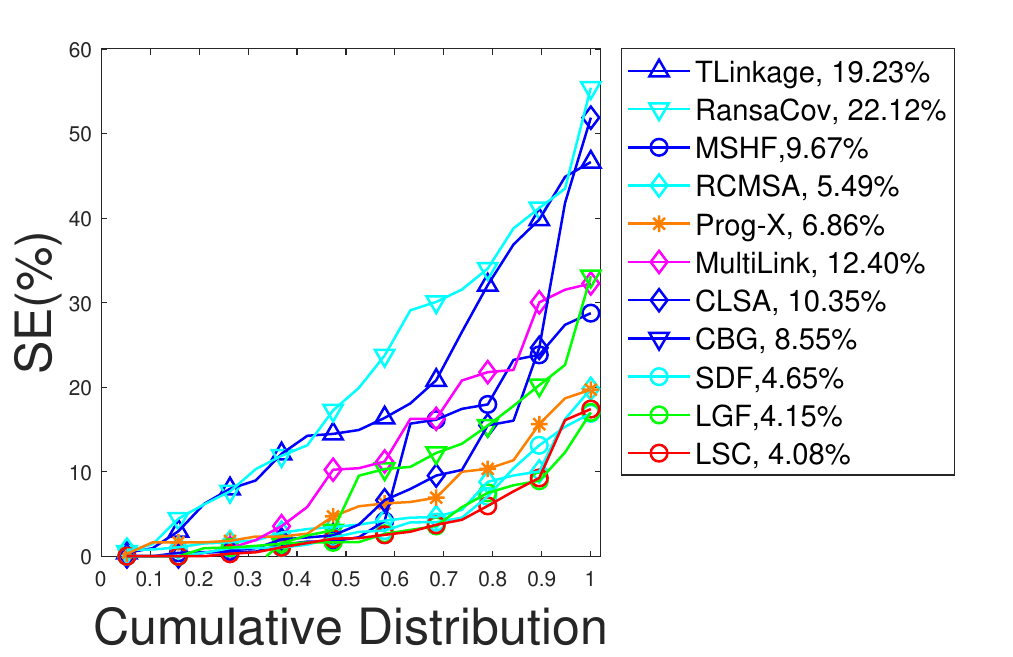}}
\end{minipage}
\begin{minipage}{.24\textwidth}
\centerline{\includegraphics[width=1.0\textwidth]{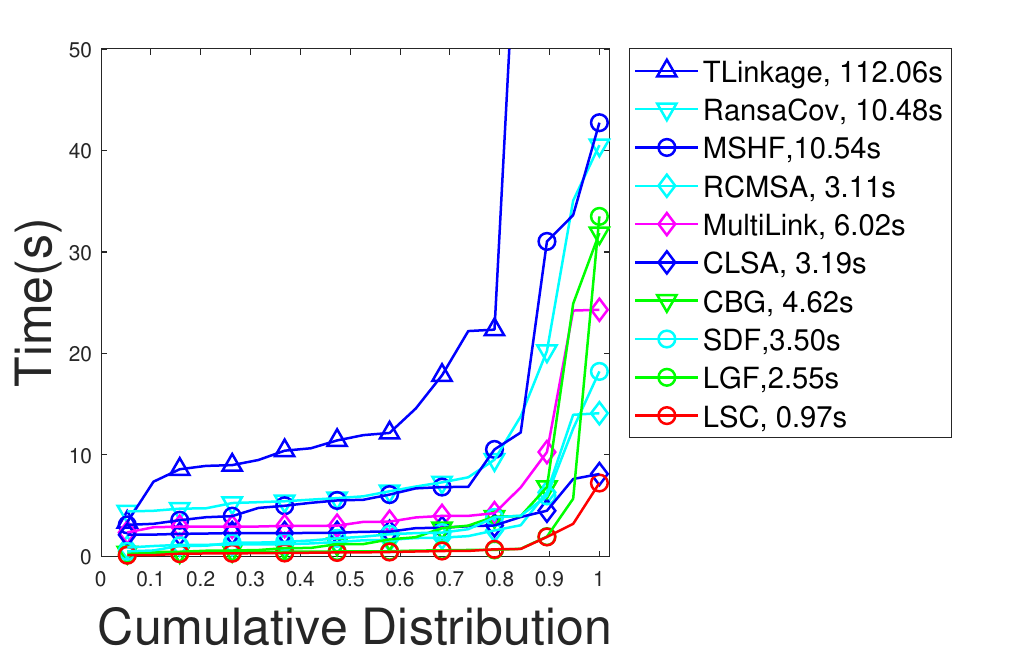}}
\end{minipage}
\caption{Quantitative comparisons of eleven fitting methods on all image pairs with multiple-structural data from the AdelaideRMF dataset for homography estimation. }
\label{fig:fittingH}
\end{figure}

We show some challenging multiple-structural cases obtained by LSC in Fig.~\ref{fig:examples2H}, and report quantitative comparisons of eleven fitting methods for homography estimation on the AdelaideRMF dataset in Fig.~\ref{fig:fittingH}.

{In} Fig.~\ref{fig:fittingH}, three deterministic fitting methods (i.e., SDF, LGF and LSC) achieve better performance on SE over the other competing methods. That is, the values of SE only obtained by SDF, LGF and LSC are below $5\%$. This can show the effectiveness of deterministic fitting methods. For the CPU time, the proposed LSC is the fastest among all competing methods, and it is only one method within one second. From Fig.~\ref{fig:examples2H}, it is hard to handle the multiple-structural data. For example, there is a model instance with a few correspondences and a model instance with correspondences from two different parts on the ``Barrsmith" image pair; Correspondences from two different model instances are close on the ``Elderhallb" image pair; The distributions of correspondences from different model instances are totally different on the ``Napiera" and ``Neem" image pairs; The number of correspondences from different model instances are unbalanced on the ``Hartley" and ``Johnsona" image pairs. Even so, the proposed LSC is able to successfully estimate all model instances in data, and achieves the best performance on both SE and Time among all competing methods.

\begin{figure}[t]
\centering
\begin{minipage}{.23\textwidth}
\centerline{\includegraphics[width=1.0\textwidth]{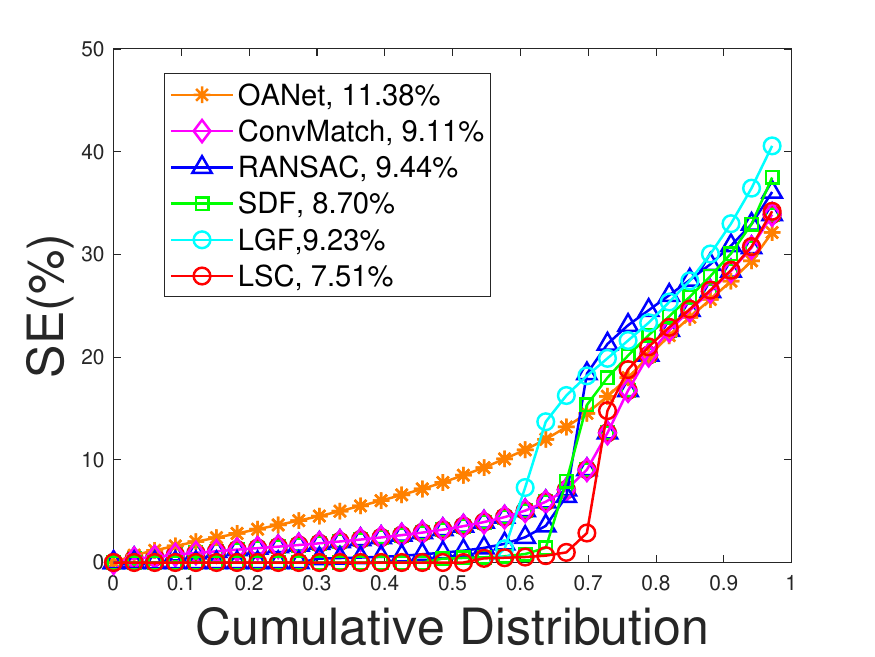}}
\end{minipage}
\begin{minipage}{.23\textwidth}
\centerline{\includegraphics[width=1.0\textwidth]{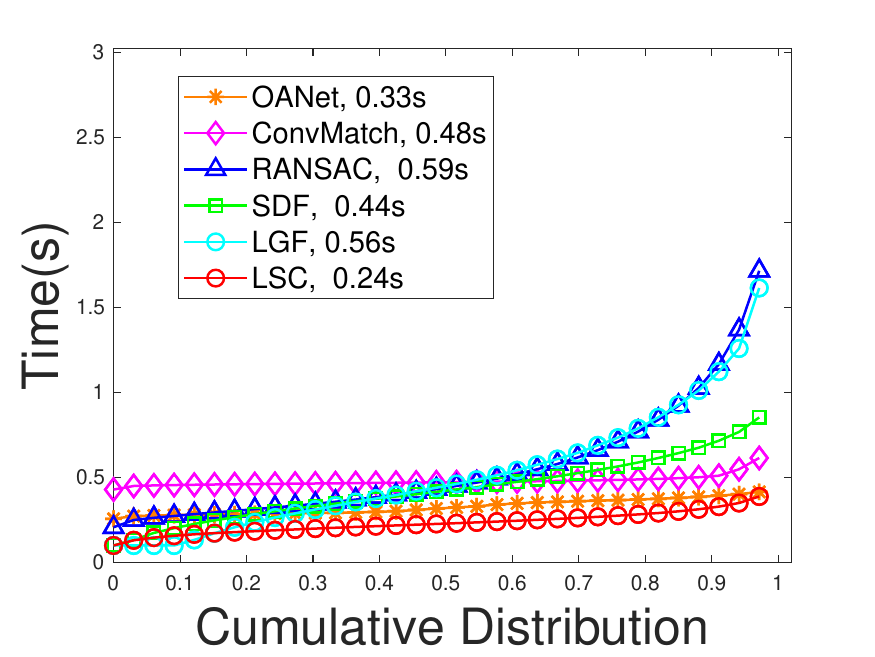}}
\end{minipage}
\caption{Quantitative comparisons of four fitting methods on all image pairs with single-structural data from the MS-COCO-F dataset for homography estimation. {In order to enhance visualization, the values are presented at intervals of $1000$.}}
\label{fig:fittingH2}
\end{figure}

We report quantitative comparisons of six fitting methods for homography estimation on the MS-COCO-F dataset in Fig.~\ref{fig:fittingH2}. {Two} deep learning based methods achieve large values of SE, especially for OANet. This is because we directly use the best model trained by the authors to test, and this is able to evaluate the generation of a deep learning method. RANSAC, SDF, LGF and LSC are unsupervised methods, and LSC achieves the lowest value of SE and it also uses the least time.

\begin{figure}[t]
\centering
\begin{minipage}{.156\textwidth}
\centerline{\includegraphics[width=1.0\textwidth]{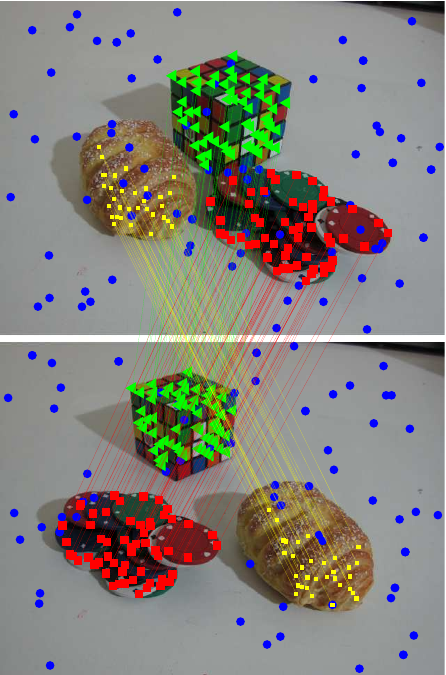}}
 \centerline{\footnotesize (a) Breadcubechips}
   \centerline{\includegraphics[width=1.0\textwidth]{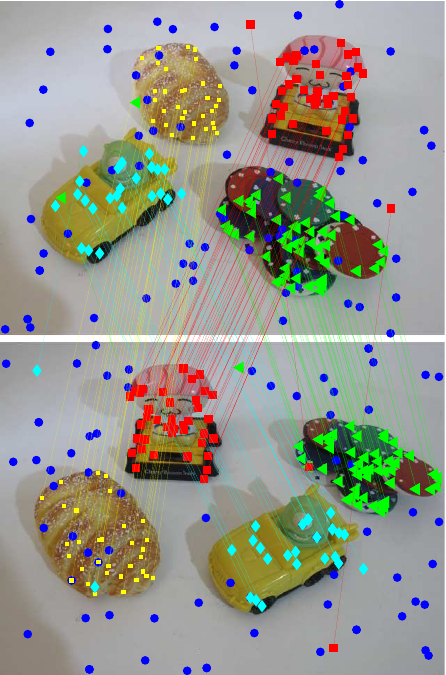}}
 \centerline{\footnotesize(d) Breadcartoychips}

\end{minipage}
\begin{minipage}{.156\textwidth}
 \centerline{\includegraphics[width=1.0\textwidth]{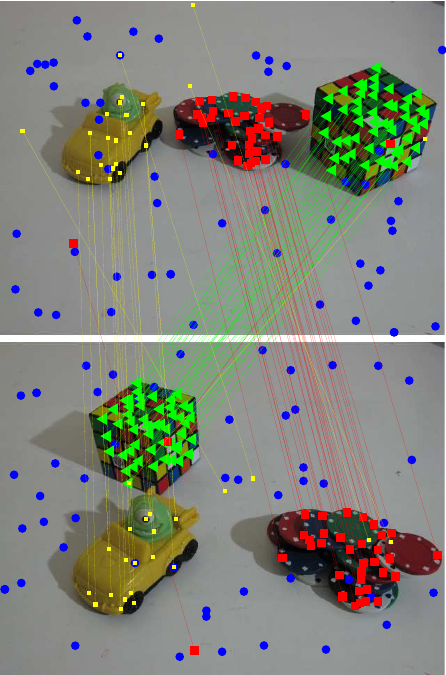}}
 \centerline{\footnotesize(b) Carchipscube}
   \centerline{\includegraphics[width=1.0\textwidth]{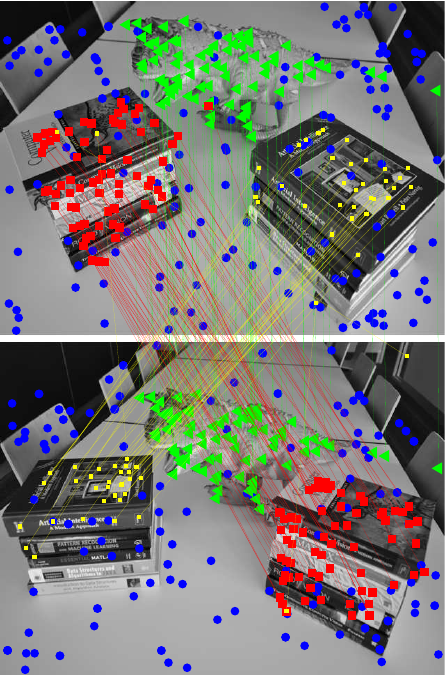}}
 \centerline{\footnotesize(e) Dinobooks}

\end{minipage}
\begin{minipage}{.156\textwidth}
\centerline{\includegraphics[width=1.0\textwidth]{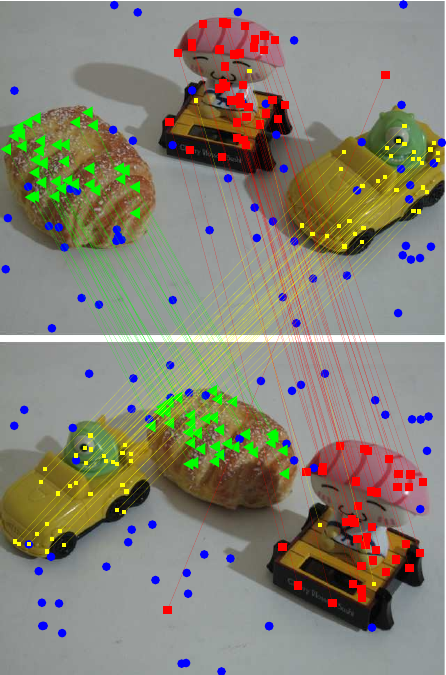}}
 \centerline{\footnotesize (c) Breadtoycar}
   \centerline{\includegraphics[width=1.0\textwidth]{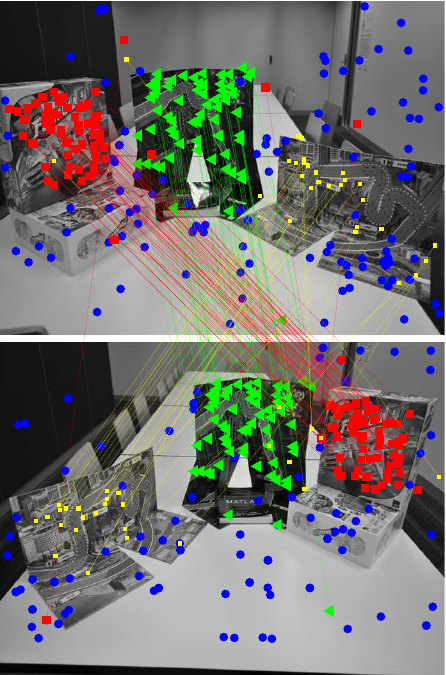}}
 \centerline{\footnotesize(f) Boardgame}
\end{minipage}
\caption{Some challenging multiple-structural cases that LSC successfully estimates model instances for fundamental matrix estimation.}
\label{fig:examples2F}
\end{figure}
\subsubsection{Fundamental Matrix Estimation}
\label{sec:fundamentalmatrix}
\begin{figure}[t]
\centering
\begin{minipage}{.24\textwidth}
\centerline{\includegraphics[width=1.0\textwidth]{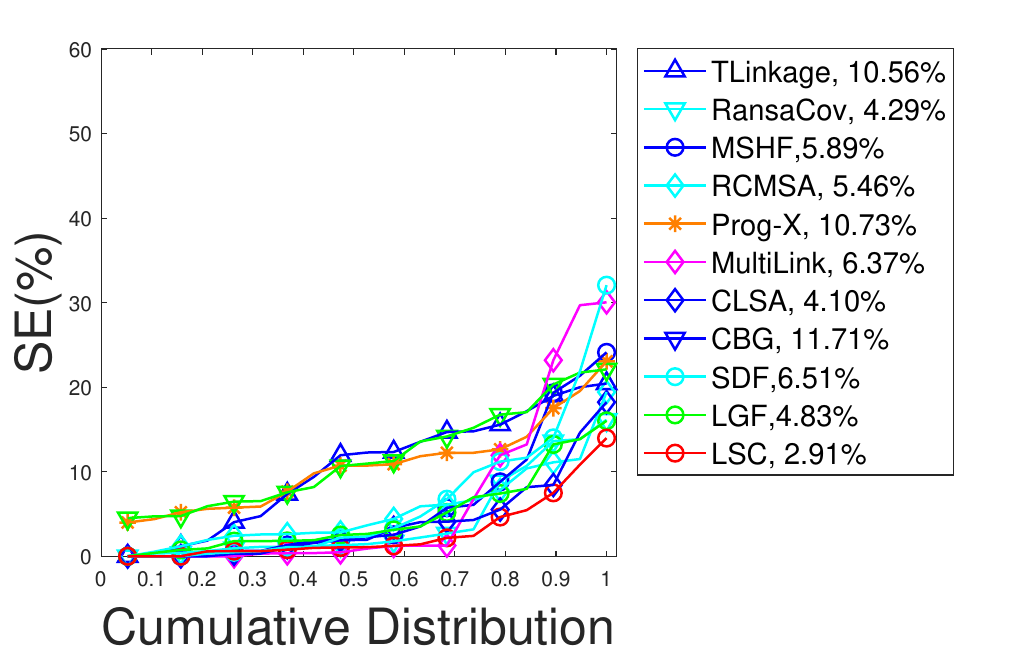}}
\end{minipage}
\begin{minipage}{.24\textwidth}
\centerline{\includegraphics[width=1.0\textwidth]{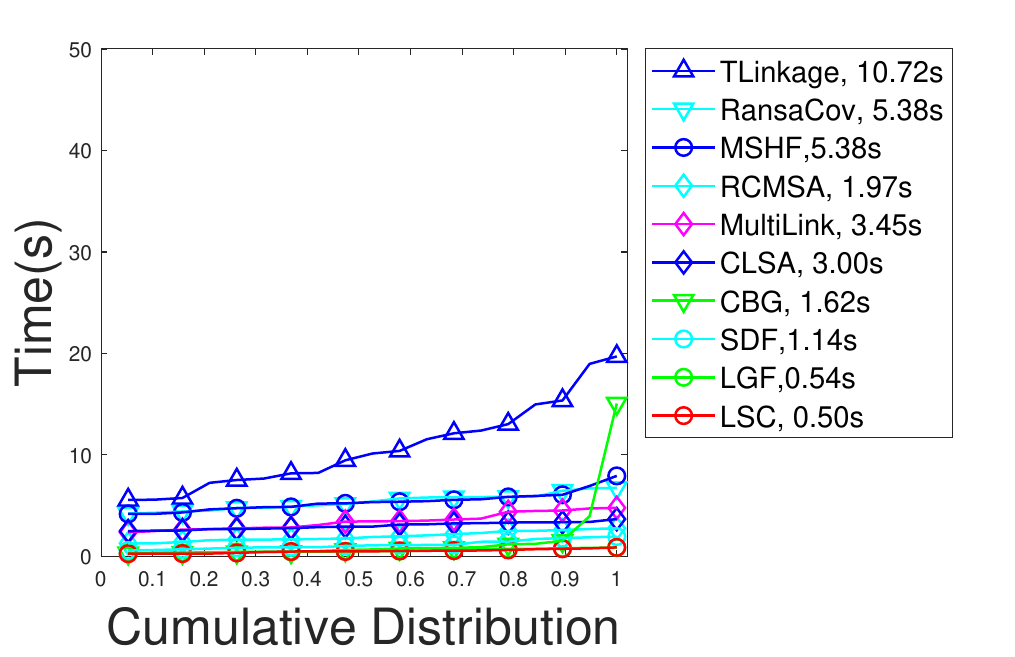}}
\end{minipage}
\caption{Quantitative comparisons of ten fitting methods on all image pairs with multiple-structural data from the AdelaideRMF dataset for fundamental matrix estimation.}
\label{fig:fittingF}
\end{figure}

We show some challenging multiple-structural cases obtained by LSC in Fig.~\ref{fig:examples2F} and report quantitative comparisons of eleven fitting methods for fundamental matrix estimation on the AdelaideRMF dataset in Fig.~\ref{fig:fittingF}.

In Fig.~\ref{fig:fittingF}, some fitting methods, e.g., RansaCov, MSHF, RCMSA, CLSA, SDF and LGF, achieve small values of SE. But, LSC is able to improve the state-of-the-art performance on SE, and achieves the lowest values of SE among all eleven fitting methods due to the effectiveness of LSC-SA and LSC-MSA algorithms. In addition, LSC is the fastest one among all competing methods due to the small number of high-quality model hypotheses. From Fig.~\ref{fig:examples2F}, the proposed LSC successfully estimates all model instances in data. Note that, some correspondences are wrongly labeled on some image pairs because different model instances involves different inlier noise scales and it is hard to know the scales in advance.

\begin{figure}[ht!]
\centering
\begin{minipage}{.22\textwidth}
\centerline{\includegraphics[width=1.0\textwidth]{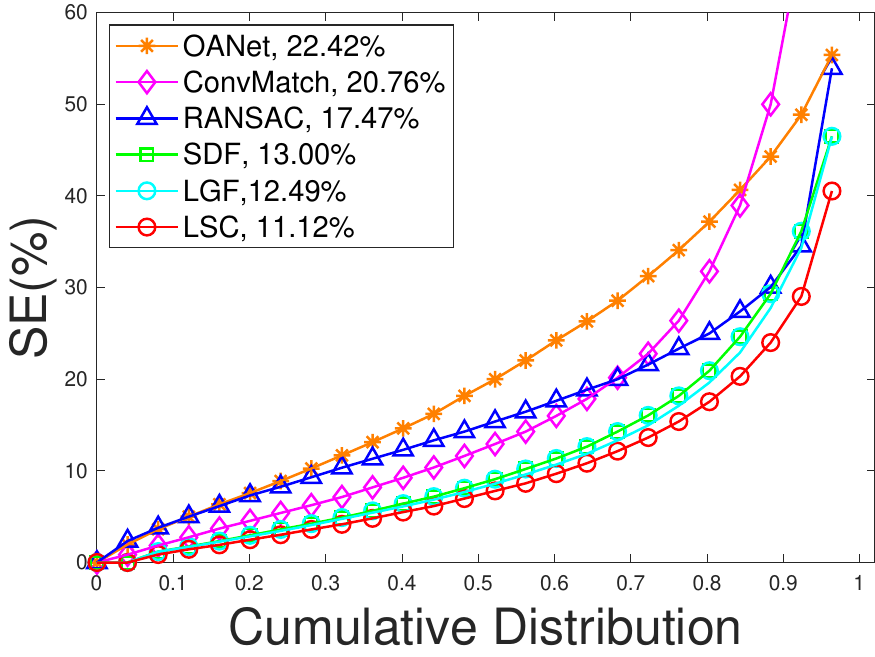}}
\end{minipage}
\begin{minipage}{.22\textwidth}
\centerline{\includegraphics[width=1.0\textwidth]{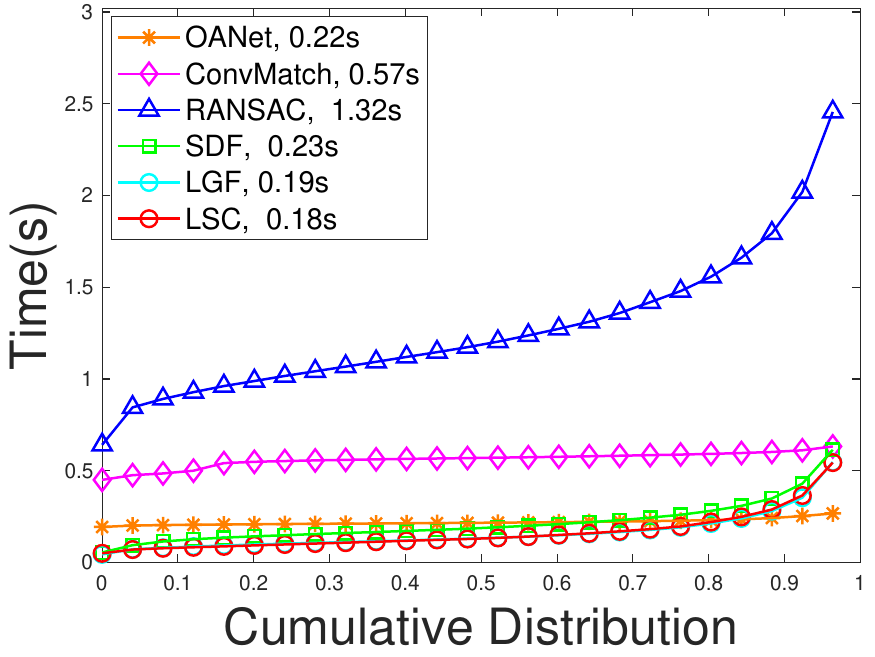}}
\end{minipage}
\caption{Quantitative comparisons of four fitting methods on all image pairs with single-structural data from the YFCC100M-F dataset for fundamental matrix estimation. {In order to enhance visualization, the values are presented at intervals of 500.}}
\label{fig:fittingF2}
\end{figure}
We report quantitative comparisons of six fitting methods for fundamental matrix estimation on the YFCC100M-F dataset in Fig.~\ref{fig:fittingF2}. {All} competing methods do not achieve low values of SE, especially for two deep learning based methods (i.e. OANet and ConvMatch) whose values of SE are over $20\%$. Because the YFCC100M-F dataset involves many challenging cases, e.g., complex background, multiple scales of images and different distributions of correspondences. Even so, LSC achieves the best performance on both SE and Time among all six fitting methods.
\begin{figure}[t]
\centering
\begin{minipage}[t]{.11\textwidth}
  \centering
 \centerline{\includegraphics[width=1.0\textwidth]{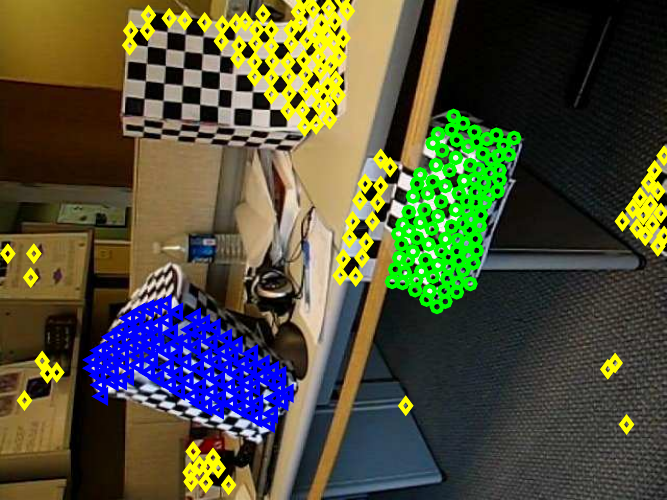}}
  \centerline{\includegraphics[width=1.0\textwidth]{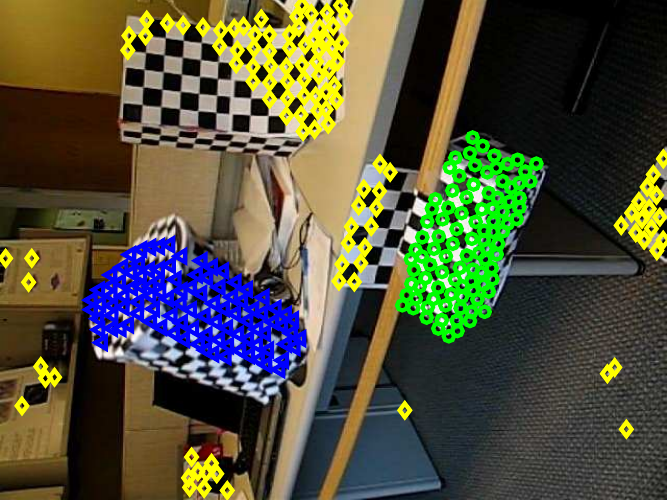}}
   \centerline{\includegraphics[width=1.0\textwidth]{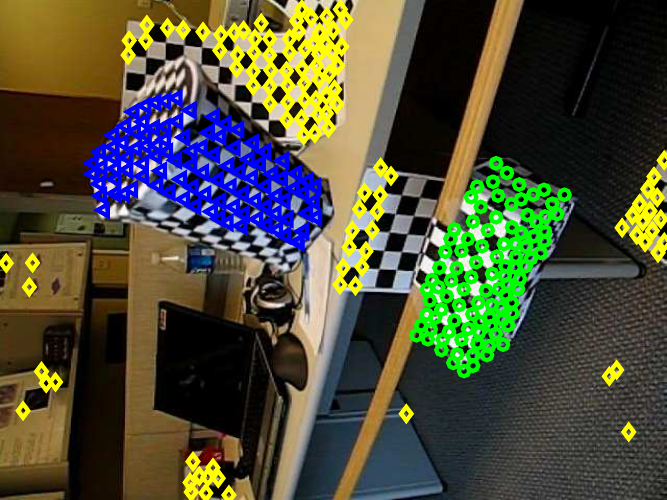}}
    \centerline{\includegraphics[width=1.0\textwidth]{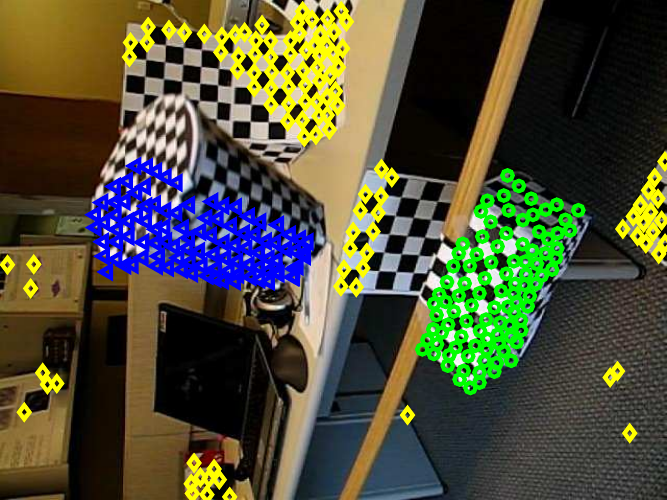}}
  \centerline{\footnotesize(a) 1RT2TC}
\end{minipage}
\begin{minipage}[t]{.11\textwidth}
  \centering
   \centerline{\includegraphics[width=1.0\textwidth]{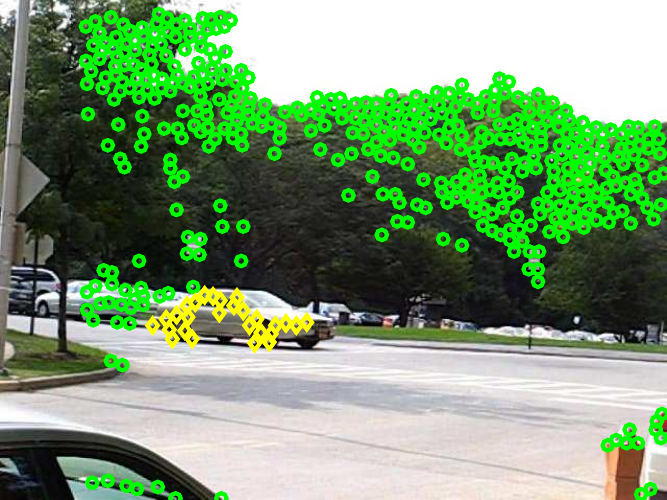}}
     \centerline{\includegraphics[width=1.0\textwidth]{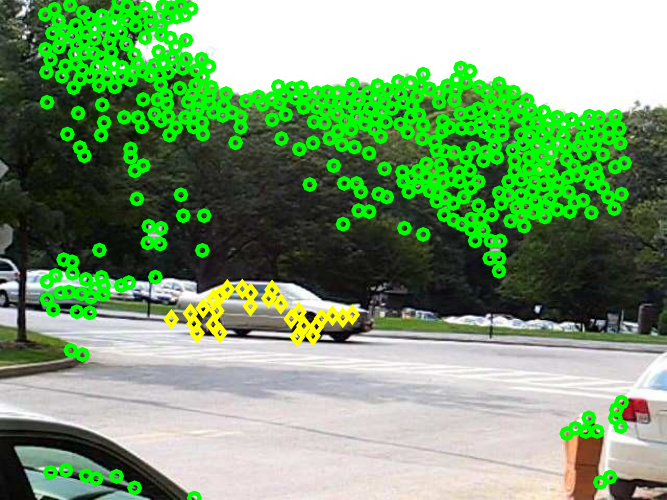}}
       \centerline{\includegraphics[width=1.0\textwidth]{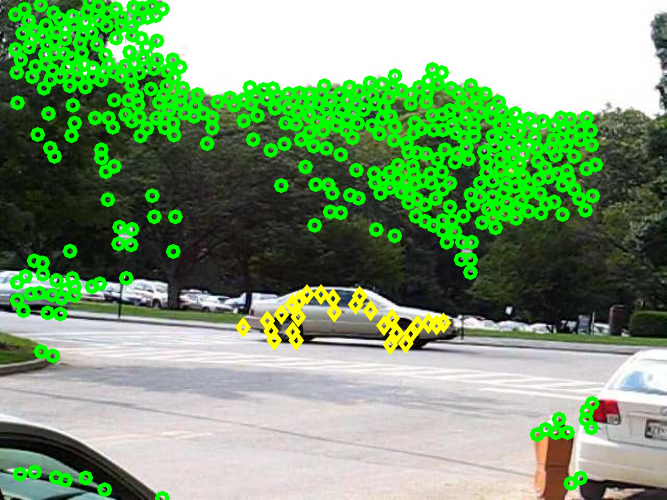}}
         \centerline{\includegraphics[width=1.0\textwidth]{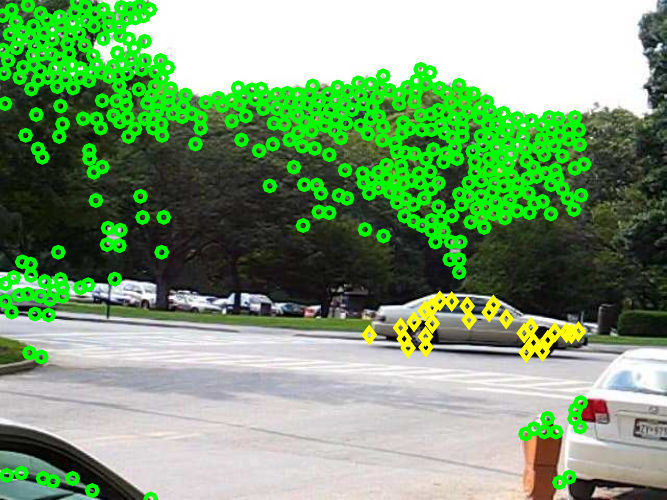}}
          \centerline{\footnotesize(b) Cars6}
\end{minipage}
\begin{minipage}[t]{.11\textwidth}
 \centering
    \centerline{\includegraphics[width=1.0\textwidth]{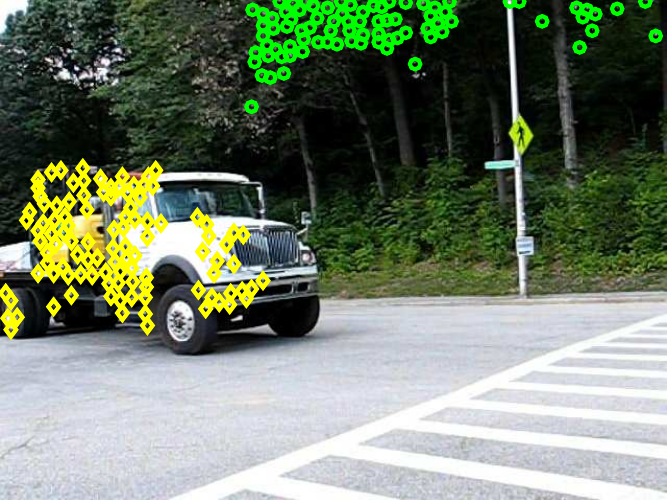}}
        \centerline{\includegraphics[width=1.0\textwidth]{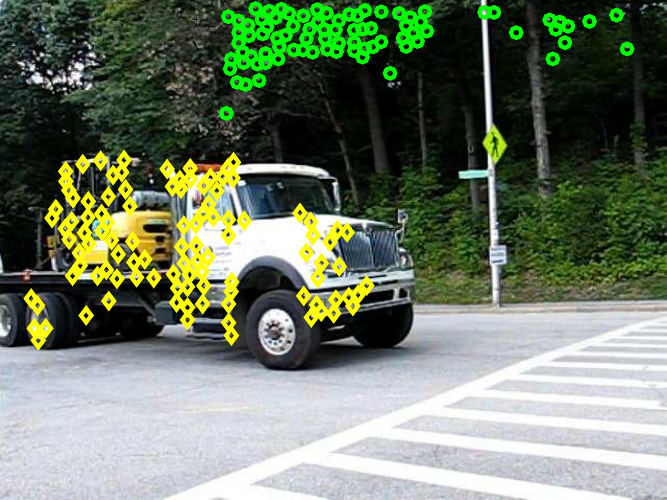}}
            \centerline{\includegraphics[width=1.0\textwidth]{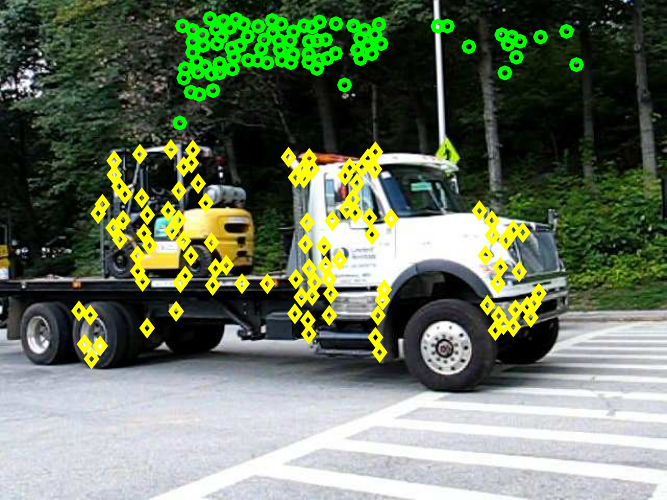}}
                \centerline{\includegraphics[width=1.0\textwidth]{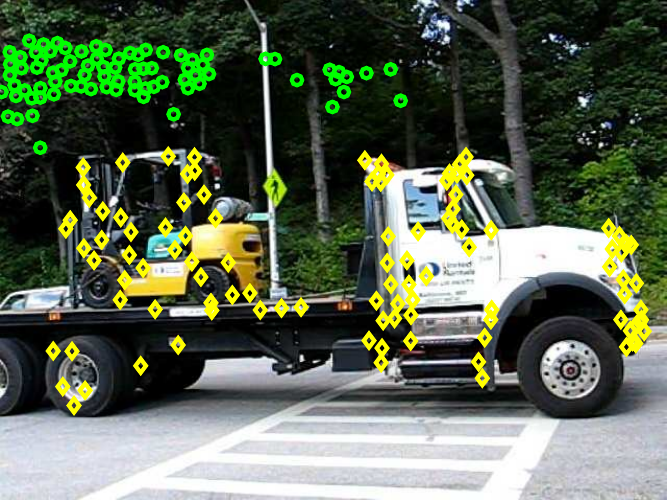}}
             \centerline{\footnotesize(c) Truck}
\end{minipage}
\begin{minipage}[t]{.115\textwidth}
  \centering
  \centerline{\includegraphics[width=1.0\textwidth,height=0.063\textheight]{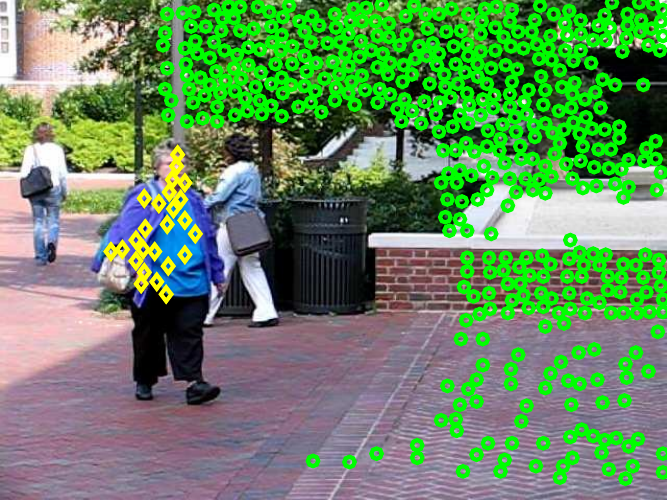}}
  \centerline{\includegraphics[width=1.0\textwidth,height=0.063\textheight]{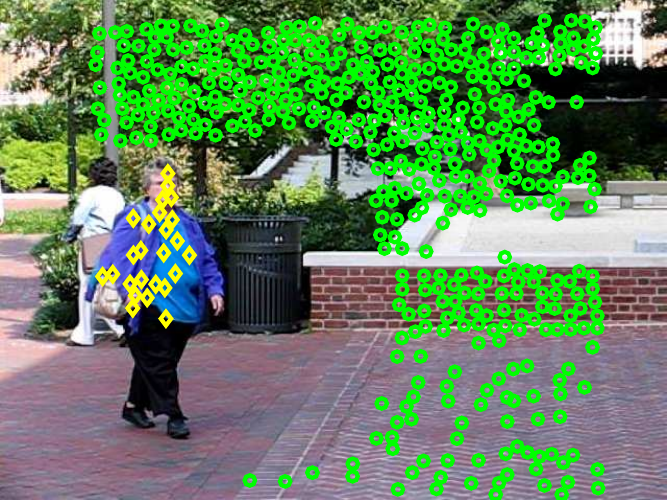}}
  \centerline{\includegraphics[width=1.0\textwidth,height=0.063\textheight]{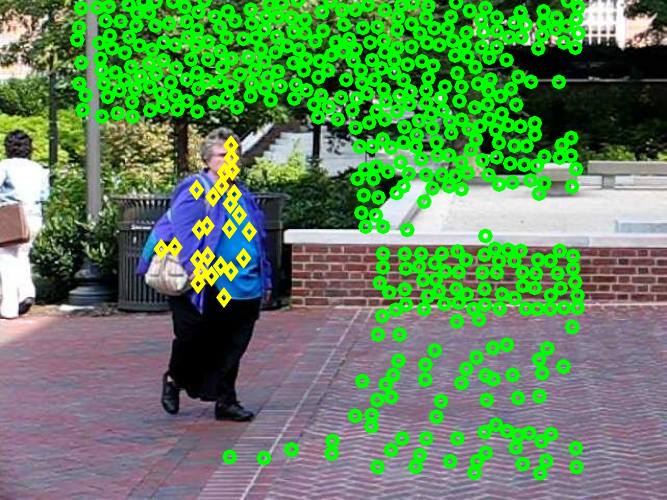}}
    \centerline{\includegraphics[width=1.0\textwidth,height=0.063\textheight]{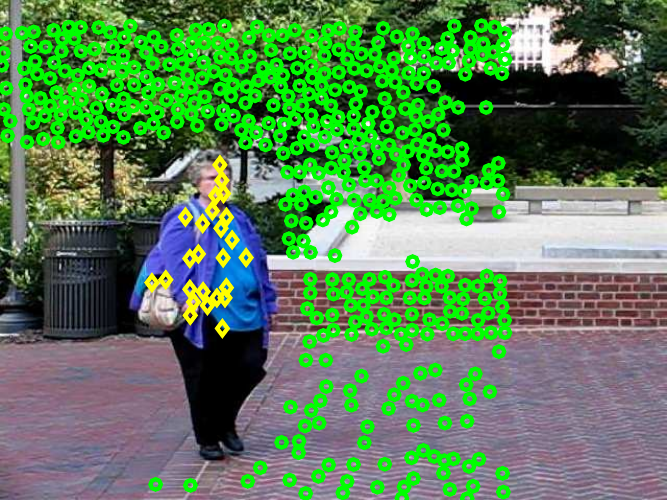}}
    \centerline{\footnotesize(d) People2}
\end{minipage}
\caption{Some motion trajectory segmentation results obtained by the proposed LSC method on $Hopkins$$155$ are shown.  From top to bottom: the results of $1$st, $10$th, $20$th and $30$th frames from the corresponding motion trajectory. }
\label{fig:3Dexample}
\end{figure}
 \begin{table}[t]
\centering
\caption{The segmentation errors (in percentage) obtained by the seven competing methods on the $Hopkins$$155$ dataset. '-' denotes that the corresponding value is not reported or no public code is available.}
\scalebox{0.85}{\tabcolsep0.02in
\begin{tabular}{cccccccc}
  \toprule
Method& SSC~\cite{elhamifar2013} & LRR~\cite{liu2013} &MTPV~\cite{li2013perspective}& RSIM~\cite{Ji_2015_ICCV}& MSSC~\cite{Lai2016ITS}&Subset~\cite{xu20193d}&LSC\\
\midrule
\multicolumn{8}{l}{{2 motions: 120 sequences}}\\
Mean&1.52  &1.33& 1.57 &0.78 &0.54&0.23&{0.59}\\
Median&0.00 &0.00 & -&0.00&0.00&0.00&0.00\\
  \multicolumn{8}{l}{{3 motions: 35 sequences}}\\
Mean&4.40  &2.51   &4.98 &1.77 &1.84&0.58&{1.10}\\
Median&0.56&0.00  &- &0.28&0.30&0.00&0.18\\
    \multicolumn{8}{l}{{All: 155 sequences}}\\
Mean&2.18  &1.59  &2.34 &1.01&0.83&0.31&{0.71}\\
Median&0.00  &0.00 &-&0.00 &0.00&0.00&0.00\\
  \bottomrule
\end{tabular}}
\label{Table:Hopkins 155Error}
\end{table}
\begin{table}[ht!]
\centering
\caption{The CPU time (in seconds) obtained by the two competing methods on the $Hopkins$$155$ dataset.}
{
\begin{tabular}{ccc}
 \toprule
Method&Subset~\cite{xu20193d}&LSC\\
\midrule
$Hopkins$$155$~\cite{tron2007}&2621.80&1508.19\\
    \bottomrule
\end{tabular}}
\label{Table:time}
\end{table}
\subsubsection{Motion Segmentation}
In this subsection, we evaluate the performance of the proposed LSC on the popular dataset,  i.e., $Hopkins$$155$~\cite{tron2007}, for the task of motion segmentation, and we compare it with several state-of-the-art motion trajectory segmentation methods, including: SSC~\cite{elhamifar2013}, LRR~\cite{liu2013}, MTPV~\cite{li2013perspective}, RSIM~\cite{Ji_2015_ICCV}, MSSC~\cite{Lai2016ITS} and Subset~\cite{xu20193d}. We report the segmentation errors (in percentage) obtained by all the seven competing methods in Table~\ref{Table:Hopkins 155Error}, and show some motion trajectory segmentation results obtained by the proposed LSC in Fig.~\ref{fig:3Dexample}. We also report the CPU time used by Subset and LSC on the three datasets in Table~\ref{Table:time}.

{In} Fig.~\ref{fig:3Dexample} and Table~\ref{Table:Hopkins 155Error}, three model fitting based methods (i.e. MSSC, Subset and LSC) achieve lower values of SE than the other subspace based methods. Subset achieves a little better performance on SE than MSSC and LSC, because Subset samples minimal subsets from three kinds of model, i.e., affine, homography and fundamental matrix, while MSSC and LSC only use the information of homography. For MSSC that samples minimal subsets by RANSAC, LSC uses LSC-SA to do that. LSC achieves better performance than MSSC, and this can show the effectiveness of LSC-SA. {In} Table~\ref{Table:time}, LSC is much faster than Subset (about $1.74$ times faster than Subset).

\begin{figure}[t]
\centering
\begin{minipage}[t]{.15\textwidth}
  \centering
 \centerline{\includegraphics[width=1.0\textwidth,height=20mm]{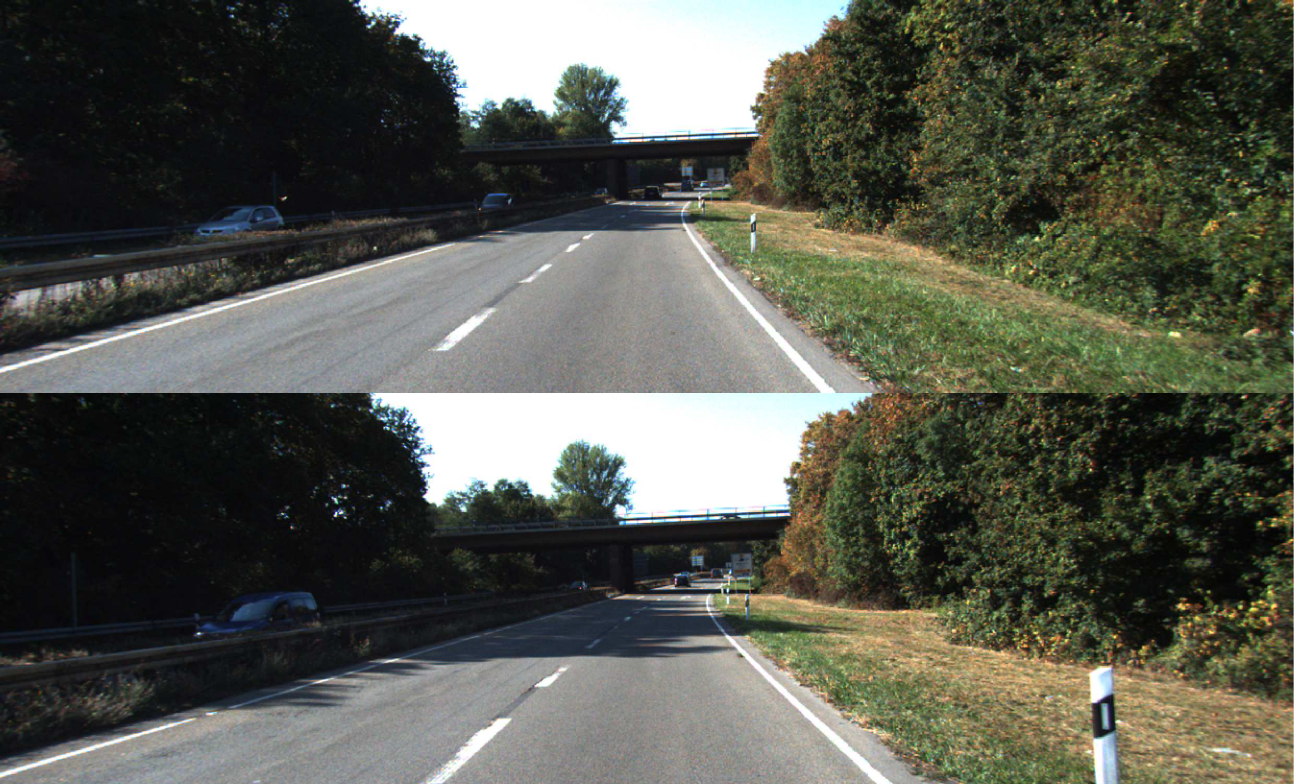}}
  \centerline{\includegraphics[width=1.0\textwidth,height=20mm]{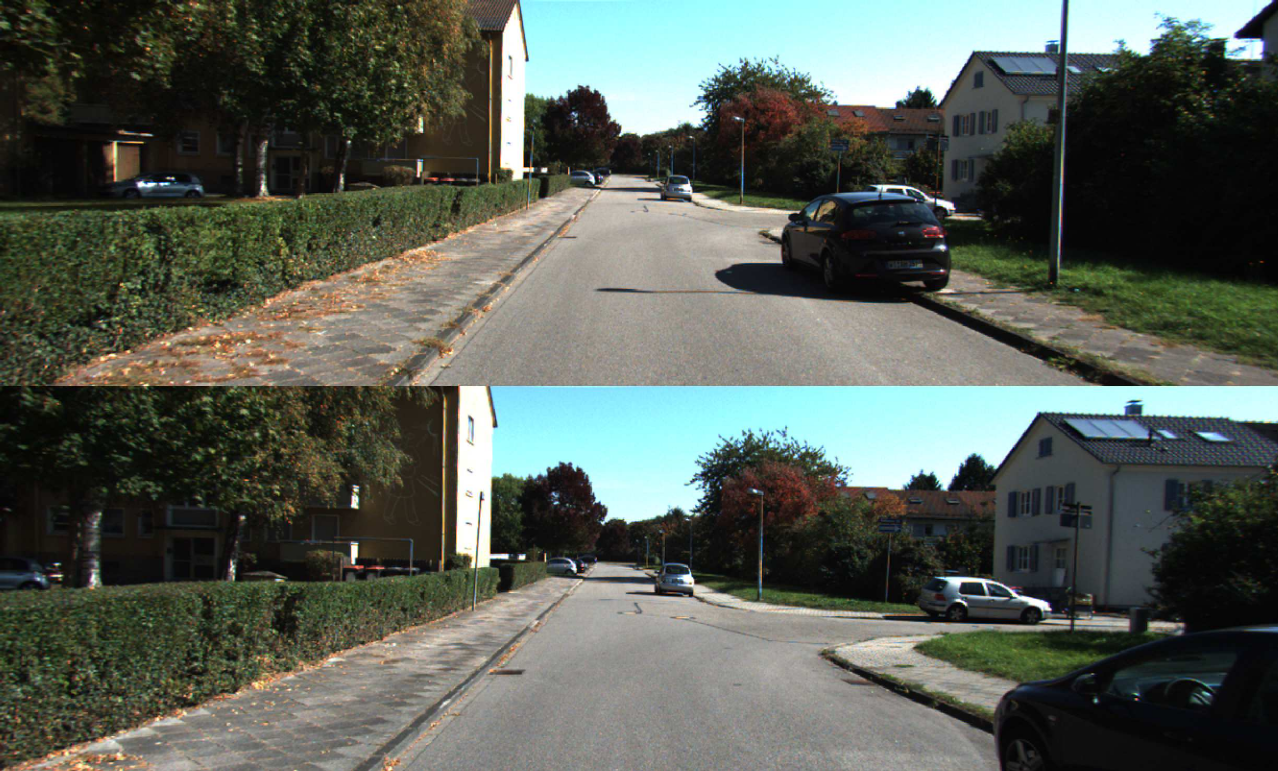}}
   \centerline{\includegraphics[width=1.0\textwidth,height=20mm]{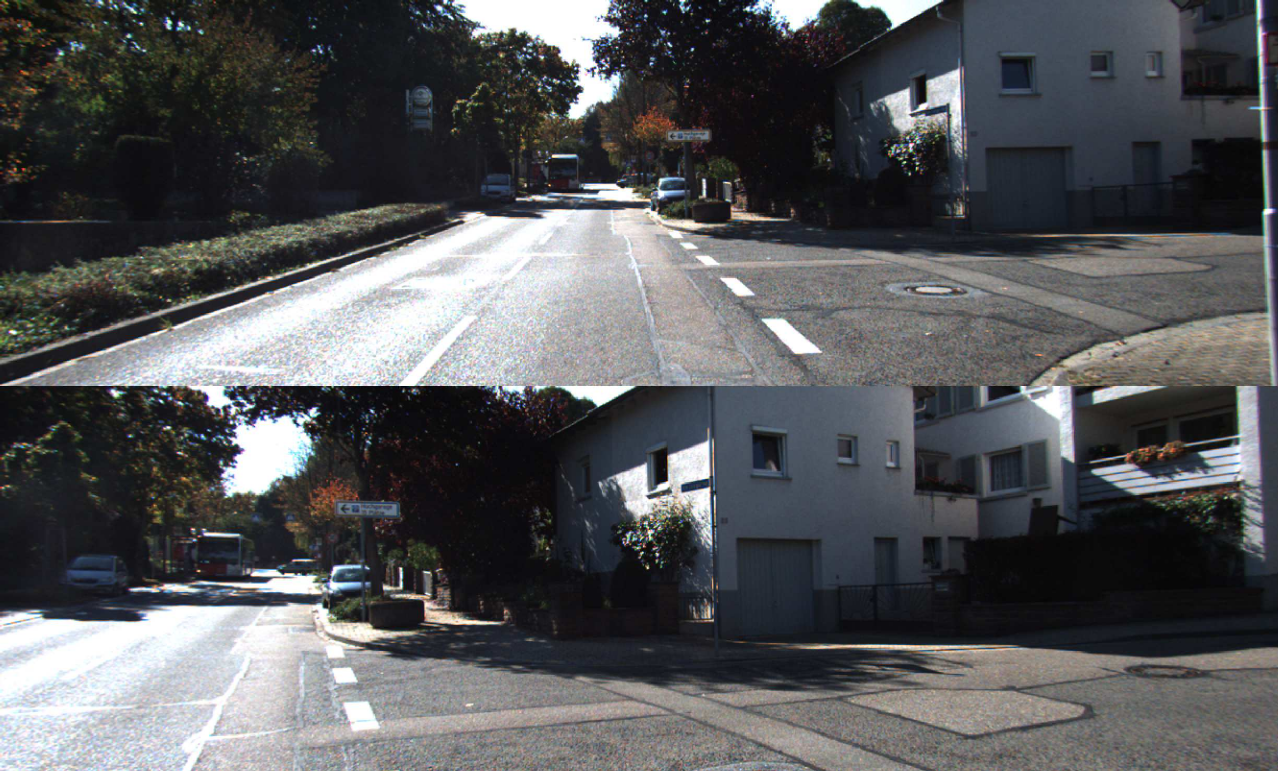}}
 \centerline{\footnotesize(a) Input images}
\end{minipage}
\begin{minipage}[t]{.15\textwidth}
  \centering
  \centerline{\includegraphics[width=1.0\textwidth,height=20mm]{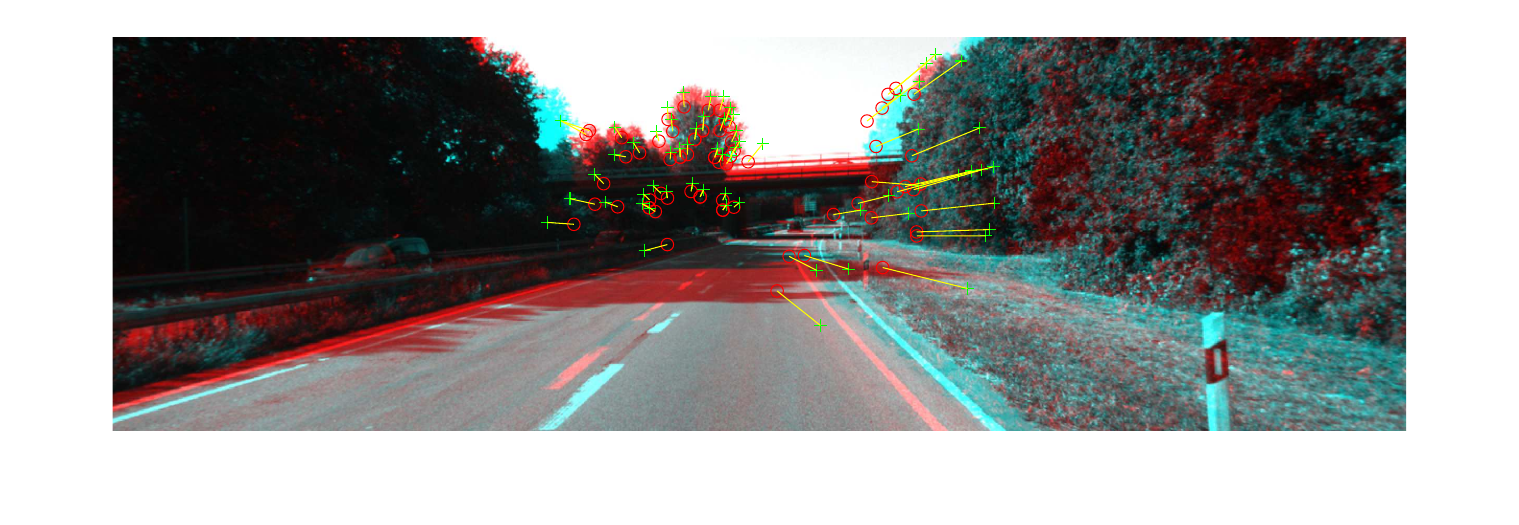}}
    \centerline{\includegraphics[width=1.0\textwidth,height=20mm]{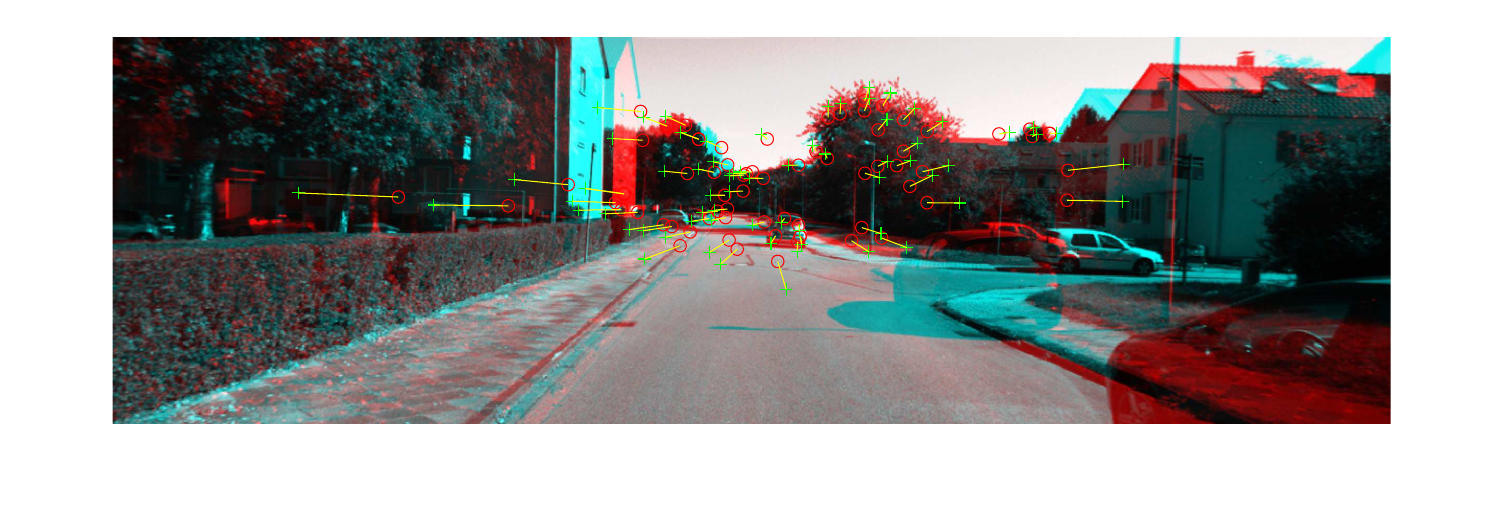}}
   \centerline{\includegraphics[width=1.0\textwidth,height=20mm]{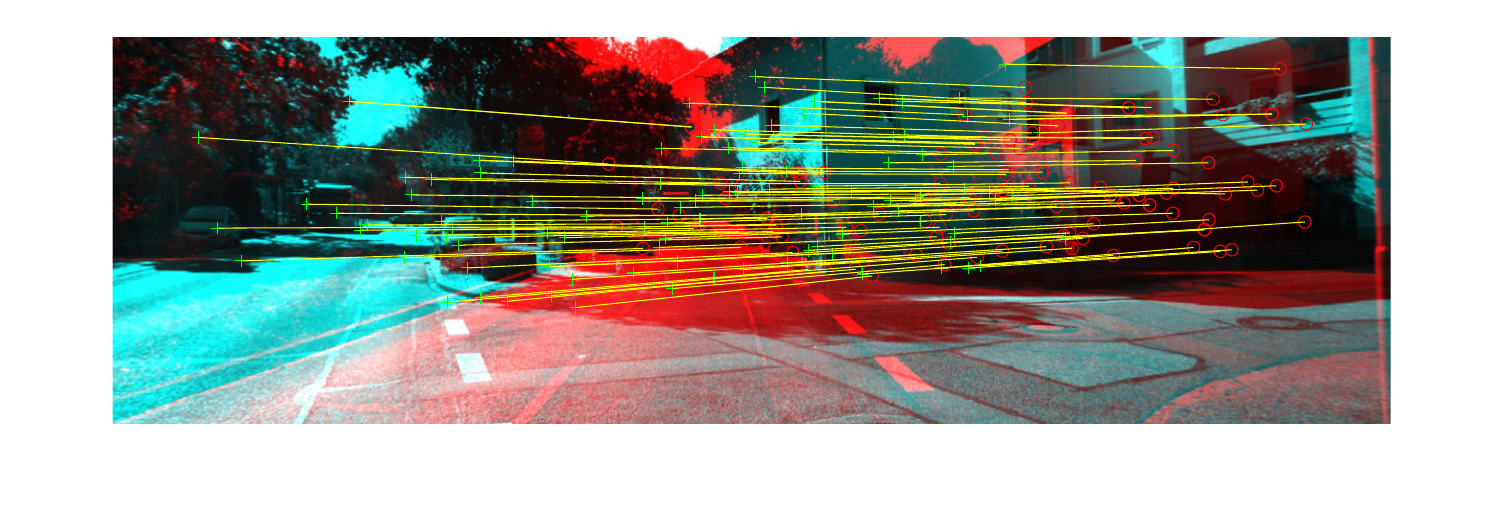}}
 \centerline{\footnotesize(b) Epipolar inliers}
\end{minipage}
\begin{minipage}[t]{.15\textwidth}
 \centering
   \centerline{\includegraphics[width=1.0\textwidth,height=20mm]{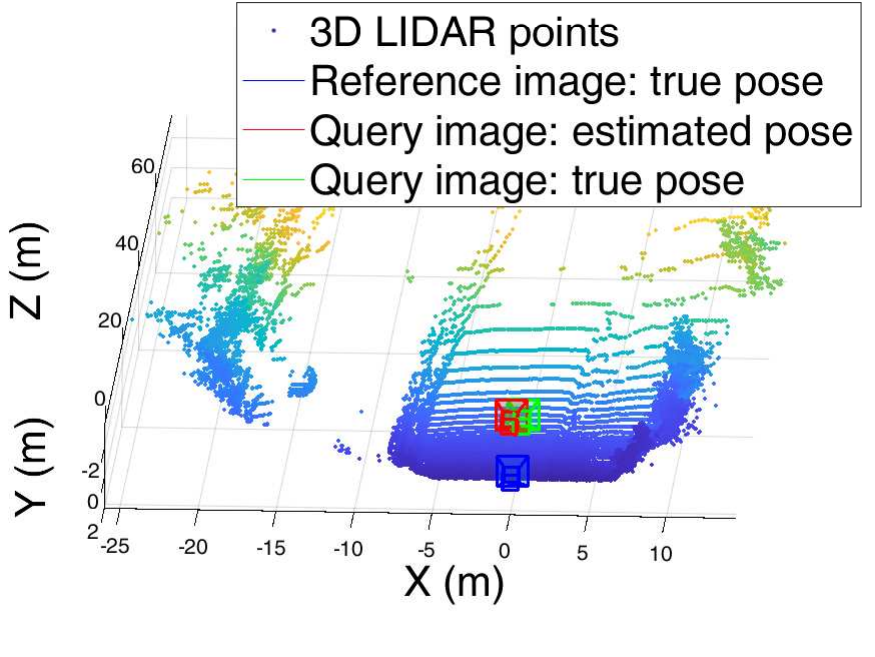}}
     \centerline{\includegraphics[width=1.0\textwidth,height=20mm]{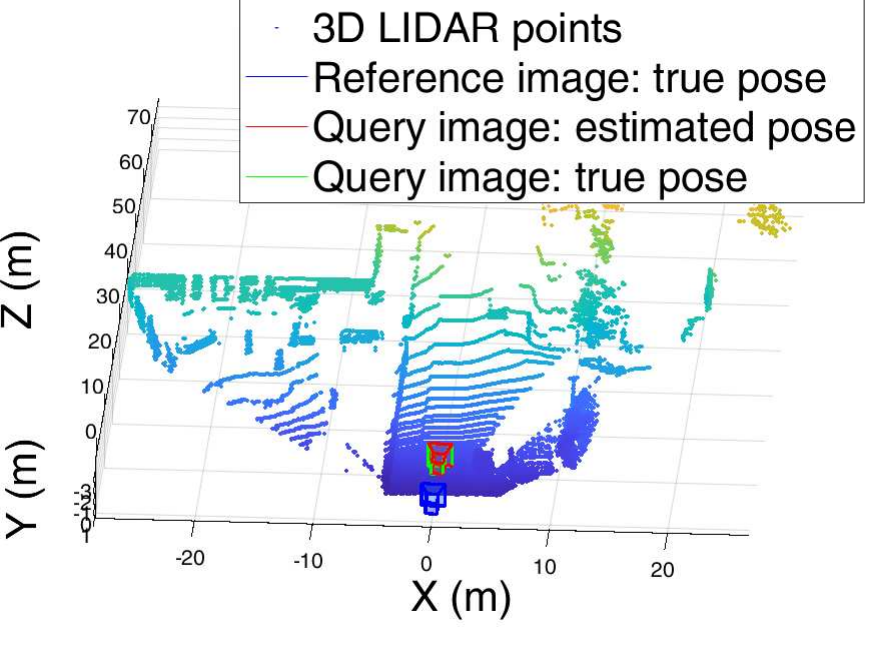}}
   \centerline{\includegraphics[width=1.0\textwidth,height=20mm]{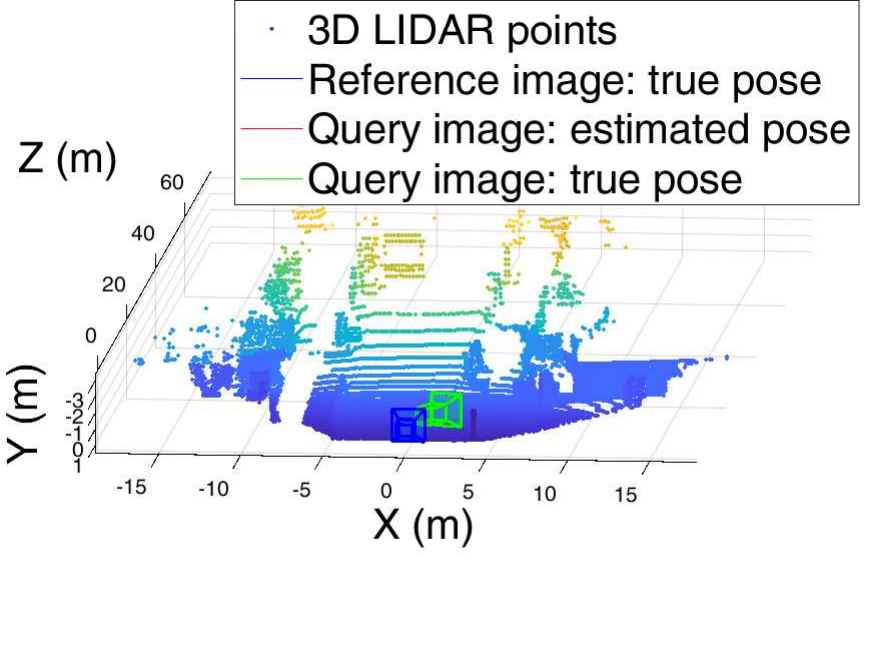}}
             \centerline{\footnotesize(c) Camera pose}
\end{minipage}
\caption{{Some camera pose results obtained by the proposed LSC method on $KITTI$ are shown. Input images consist of the (top) reference and (bottom) query images in each group.}}
\label{fig:camerposeexample}
\end{figure}
\begin{table}[t]
\centering
\caption{{Quantitative comparison results obtained by the three competing methods on the $KITTI$ dataset for the task of 6-DoF camera pose estimation. `\#Image' denotes the number of images in a sequence. TE, MOE and TIME represent the translation error (in meters), maximum orientation error (in degrees) and the CPU time (in seconds), respectively.}}
\scalebox{0.7}{
\begin{tabular}{ccccccccccc}
\toprule
\multicolumn{2}{c}{Method} & \multicolumn{3}{c}{RANSAC} & \multicolumn{3}{c}{LGF} & \multicolumn{3}{c}{LSC} \\
\cmidrule(lr){1-2} \cmidrule(lr){3-5} \cmidrule(lr){6-8} \cmidrule(lr){9-11}
Sequence & \#Image & TE & MOE & TIME & TE & MOE & TIME & TE & MOE & TIME \\
\midrule
00 & 4541 & 1.42& \bf{1.30}& 1.07 &1.27&  1.61& 0.71 &\bf{1.04} & 1.57 &{\bf0.33} \\
01 & 1101 & 3.06& 1.87& 2.04 &3.10 & 3.08& 0.62 &\bf{2.42}&\bf{1.71}& \bf{0.36} \\
02 & 4661 & 4.01 & 3.80 & 3.51 & 2.49 & 3.28 & 0.52 &\bf{ 2.24} & \bf{2.67} & \bf{0.40} \\
03 & 801 & 0.28 & 0.24 & 3.92 & 0.29 & 0.25 & 1.87 & \bf{0.27} & \bf{0.19} & \bf{0.70} \\
04 & 271 & 1.20 & 1.14 & 1.49 & 1.65 & 1.58 & 0.92 & \bf{0.97} & \bf{0.91} &\bf{0.44} \\
05 & 2761 & 1.35 & 1.53 & 2.44 & \bf{0.85} & 1.27 & 1.54 & 0.86 & \bf{0.90} & \bf{0.40} \\
06 & 1101 & 1.09 & 1.08 & 1.26 & 0.34 & 0.42 & 0.94 & \bf{0.14} & \bf{0.15} & \bf{0.45} \\
07 & 1101 & 1.02 & 0.98 & 2.66 & 0.67 & 1.28 & 2.51 & \bf{0.44} & \bf{0.48} & \bf{0.46} \\
08 & 4071 & 1.77 & 2.44 & 2.91 & 3.42 & 2.99 & 0.85 & \bf{1.58} & \bf{1.69} & \bf{0.36} \\
09 & 1591 & 3.43 & 6.20 & 2.48 & 2.82 & 4.55 & 0.37 & \bf{2.19} & \bf{2.69} & \bf{0.26} \\
10 & 1201 & 1.53 & 1.80 & 2.40 & 0.83 & 1.80 & 1.17 & \bf{0.70} & \bf{1.34} & \bf{0.32} \\
\bottomrule
\end{tabular}}
\label{Table:camer}
\end{table}

\subsubsection{Camera Pose Estimation}
In this subsection, we evaluate the performance of the proposed LSC on the popular dataset, i.e., $KITTI$~\cite{geiger2012we}, for the task of 6-DoF camera pose estimation. We compare LSC with the deterministic method LGF, and we also use RANSAC as baseline. We report the mean value of the translation error (in meters), maximum orientation error (in degrees) and the CPU time obtained by the three competing methods on each sequence in Table~\ref{Table:camer}, and we also show some camera pose results obtained by the proposed LSC in Fig.~\ref{fig:camerposeexample}.

From Fig.~\ref{fig:camerposeexample} and Table~\ref{Table:camer}, we can see that LSC is able to effectively estimate the camera pose. LSC obtains the lowest average values of translation error and maximum orientation error for $10$ out of the $11$ sequences among three competing methods, and LSC is the faster than RANSAC and LGF.
\subsubsection{Image Registration}
\begin{figure}[t]
\centering
\begin{minipage}[t]{.23\textwidth}
  \centering
 \centerline{\includegraphics[width=1.0\textwidth,height=20mm]{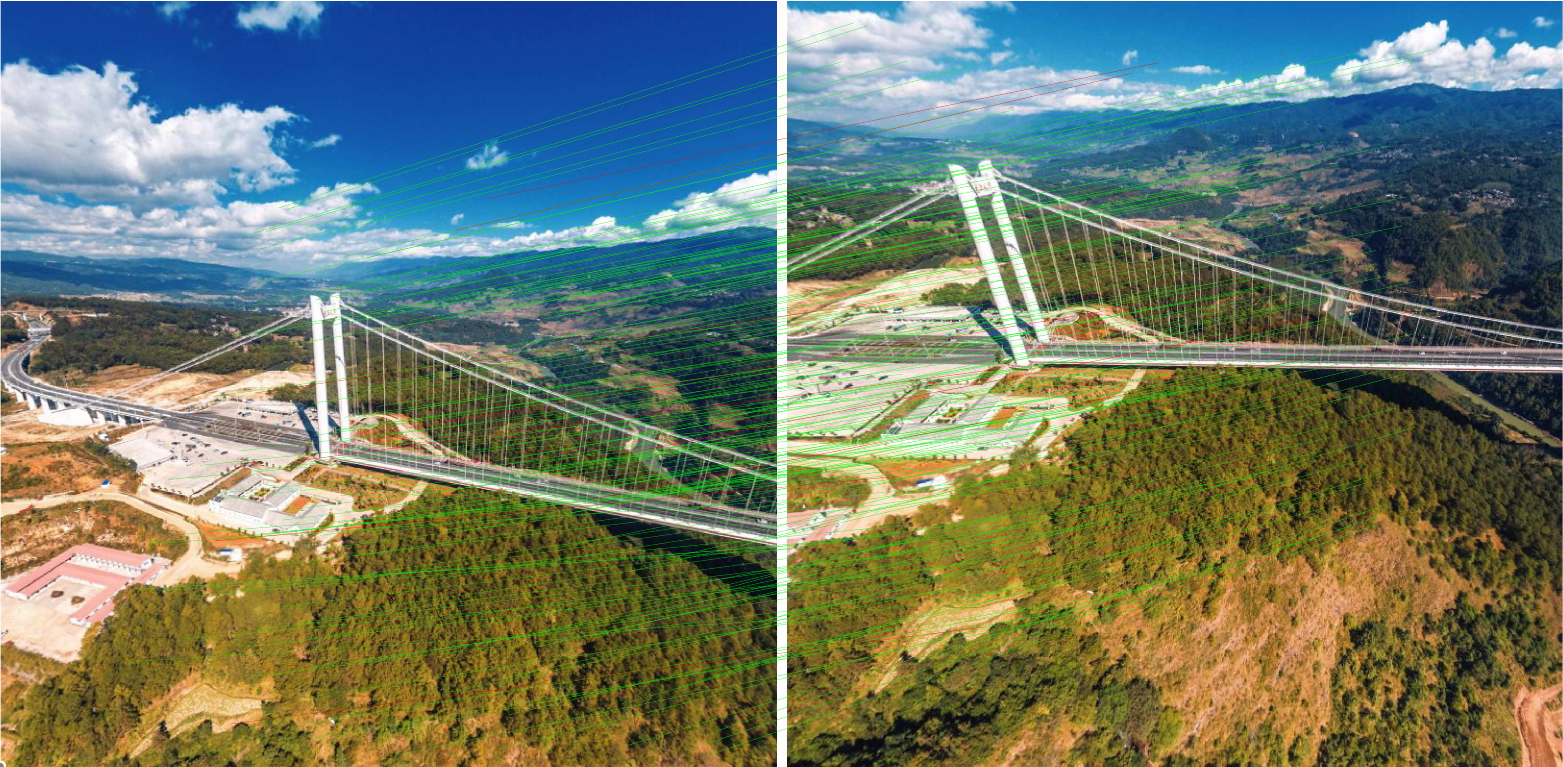}}
  \centerline{\includegraphics[width=1.0\textwidth,height=20mm]{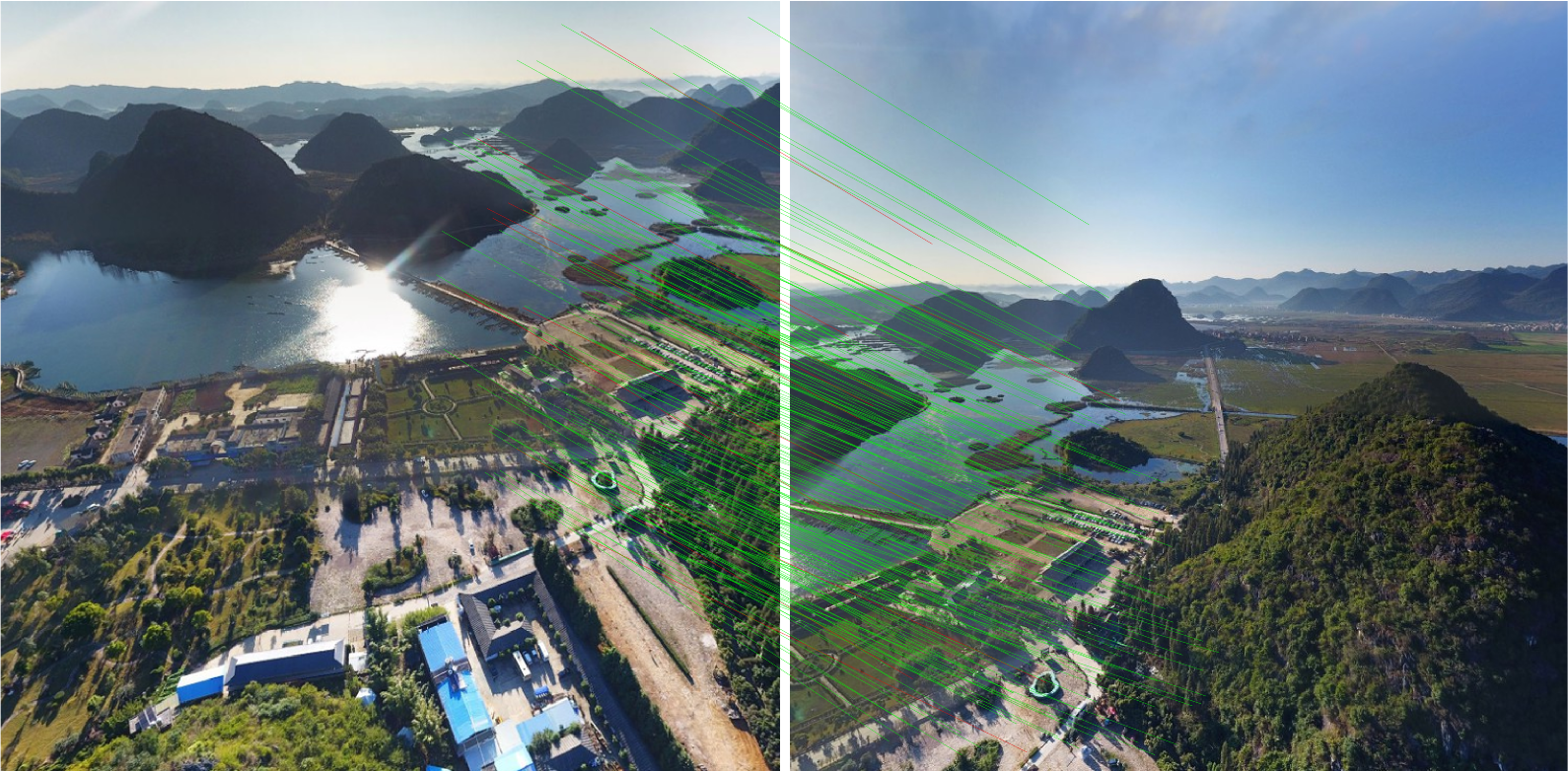}}
   \centerline{\includegraphics[width=1.0\textwidth,height=20mm]{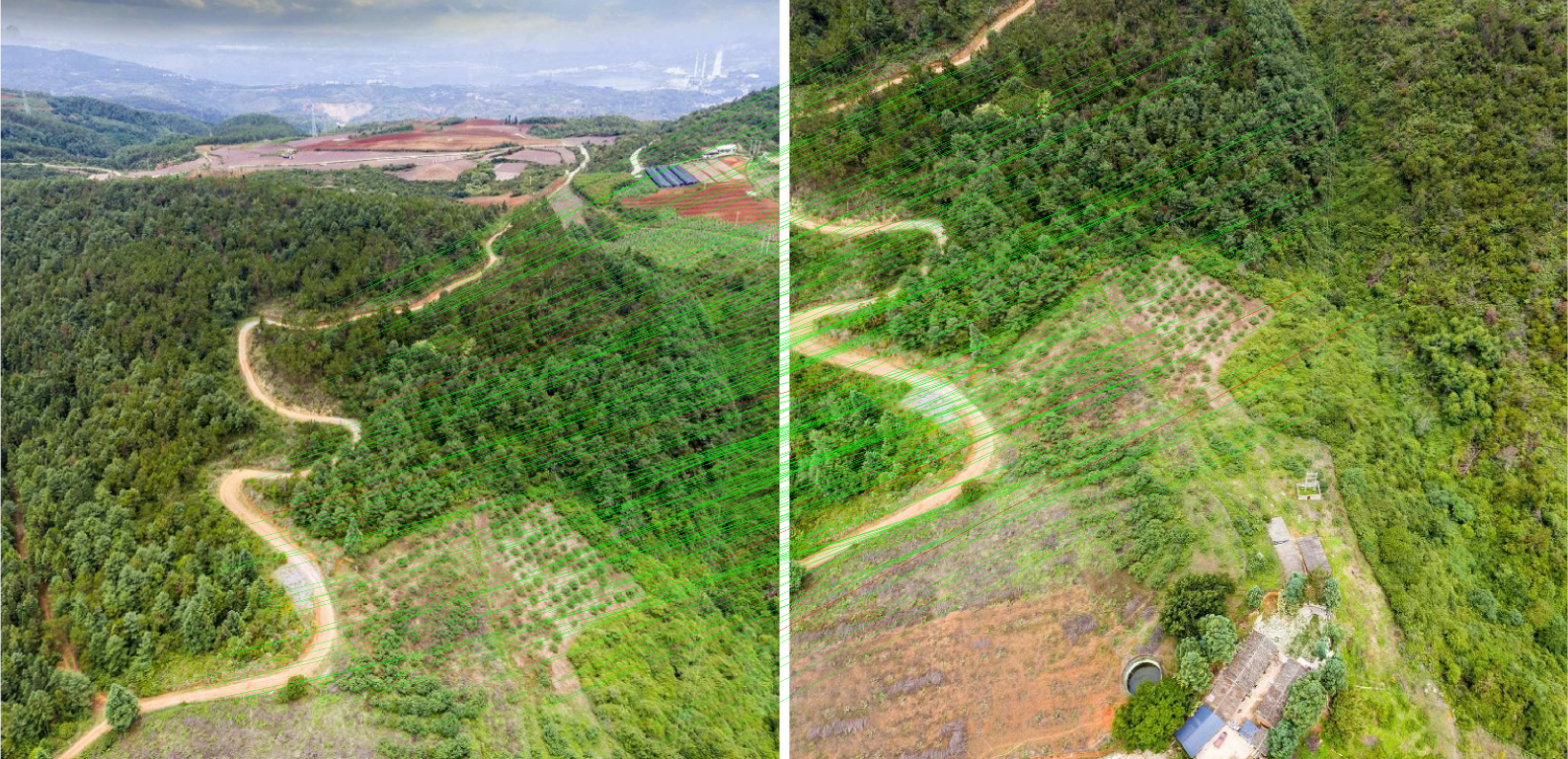}}
     \centerline{\includegraphics[width=1.0\textwidth,height=20mm]{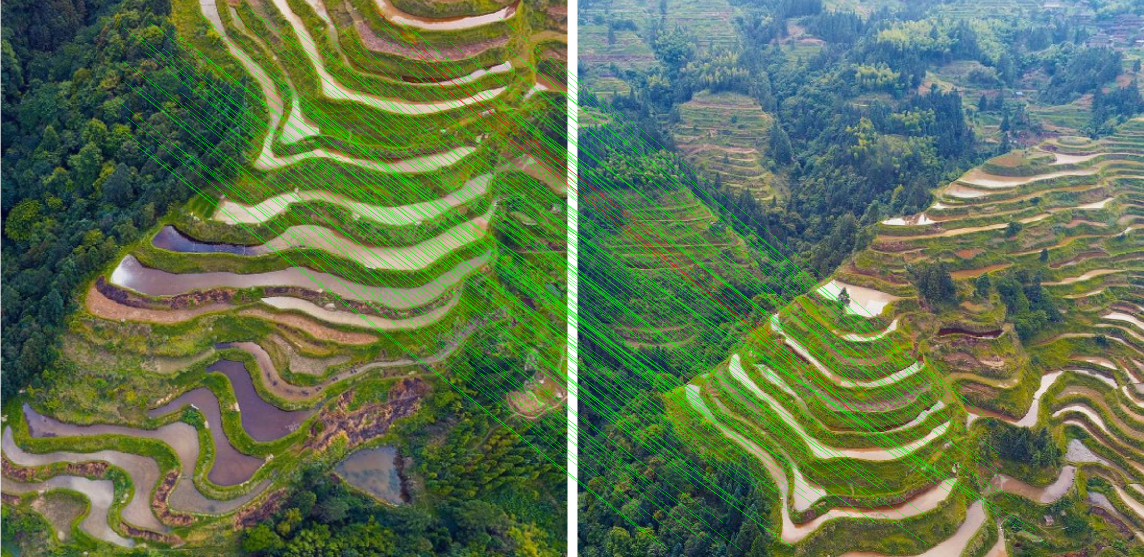}}
 \centerline{\footnotesize(a)}
\end{minipage}
\begin{minipage}[t]{.12\textwidth}
  \centering
 \centerline{\includegraphics[width=1.0\textwidth,height=20mm]{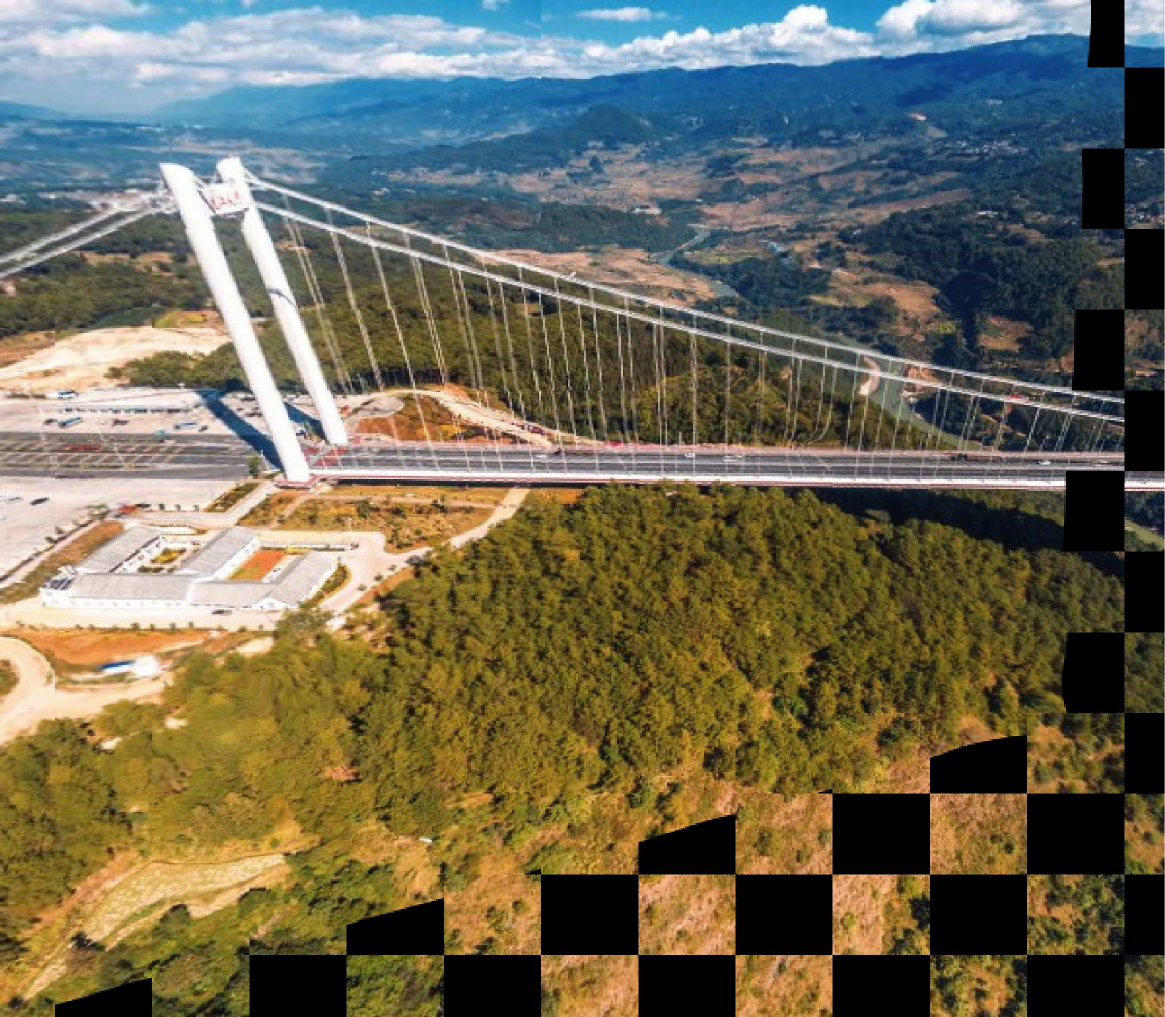}}
  \centerline{\includegraphics[width=1.0\textwidth,height=20mm]{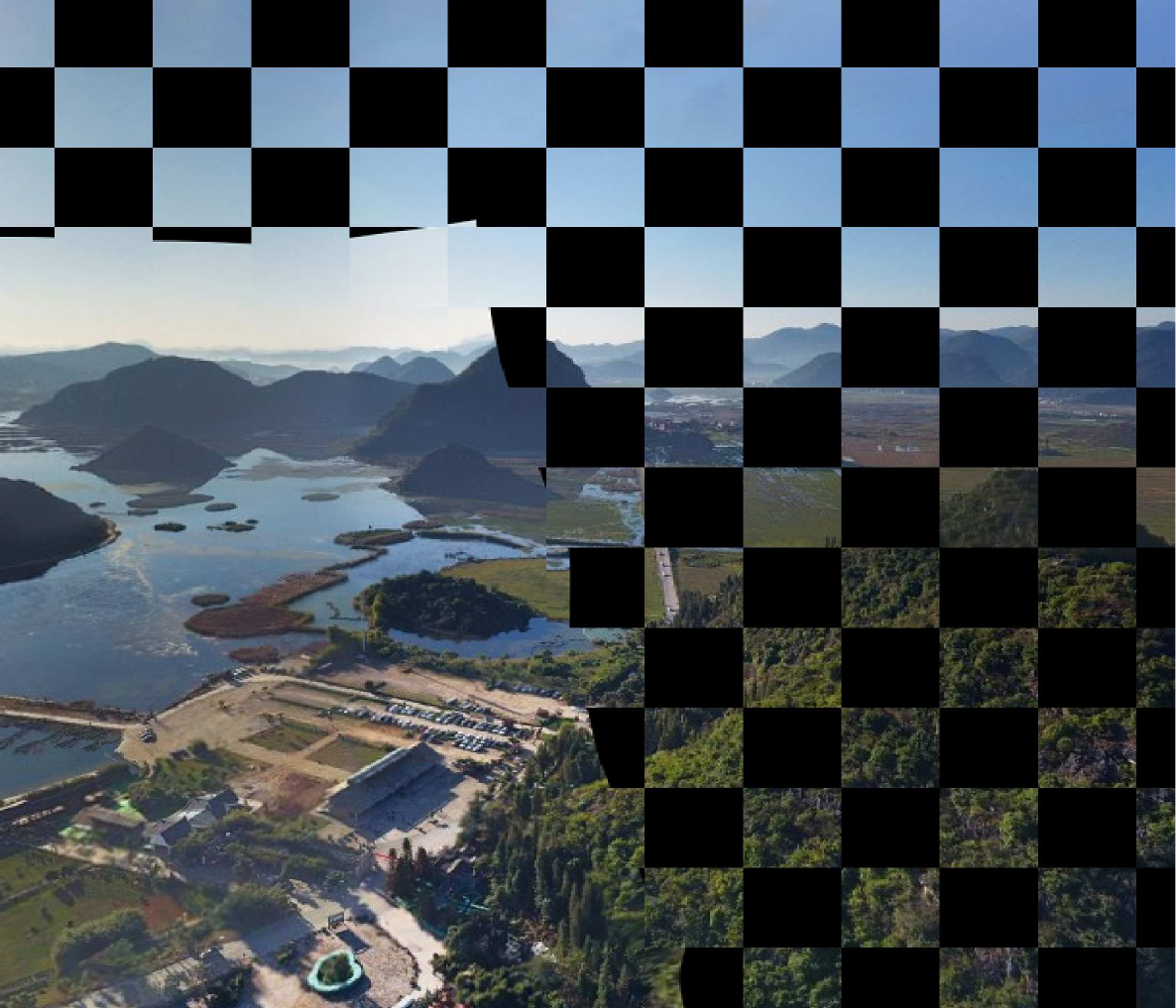}}
   \centerline{\includegraphics[width=1.0\textwidth,height=20mm]{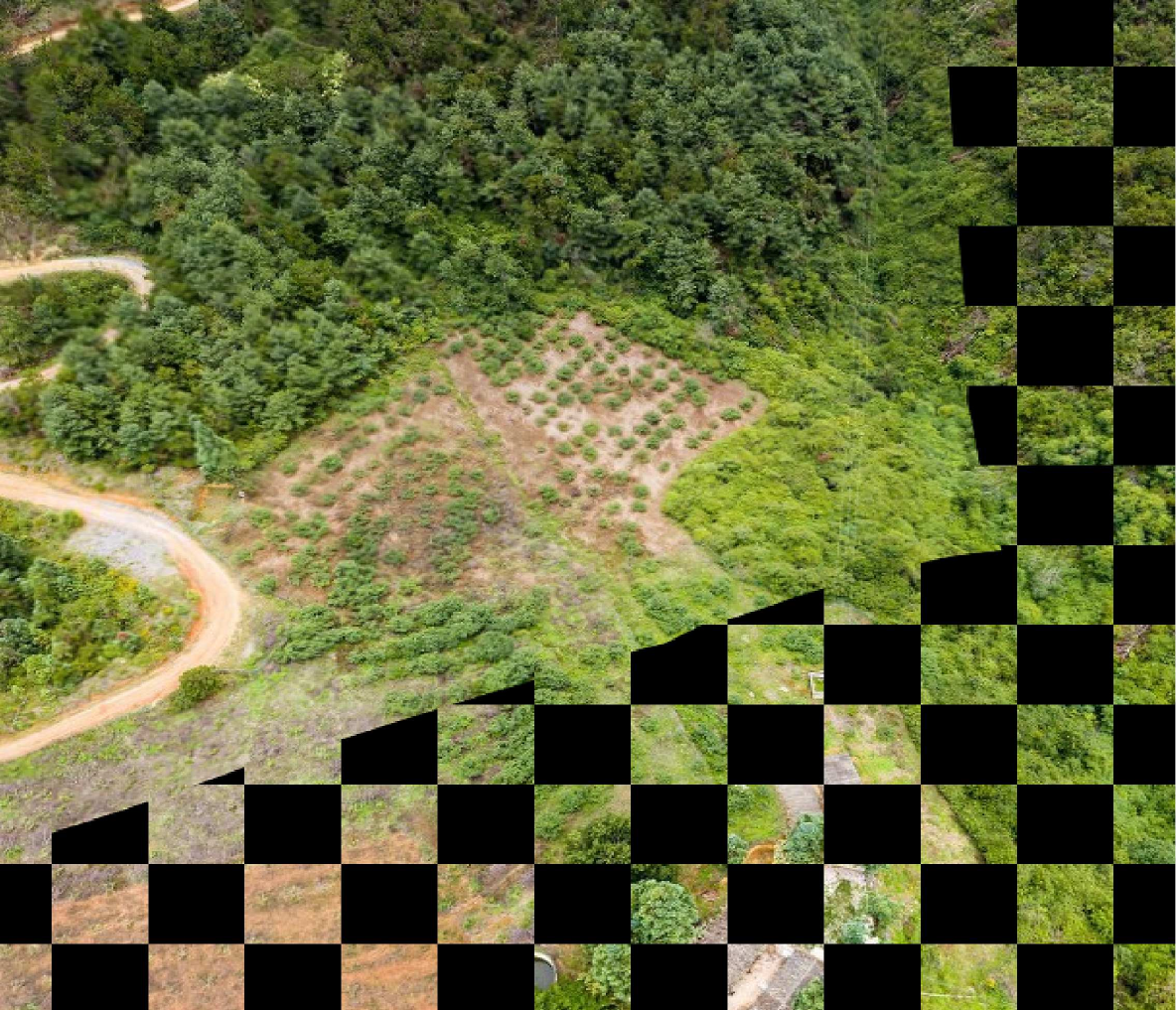}}
        \centerline{\includegraphics[width=1.0\textwidth,height=20mm]{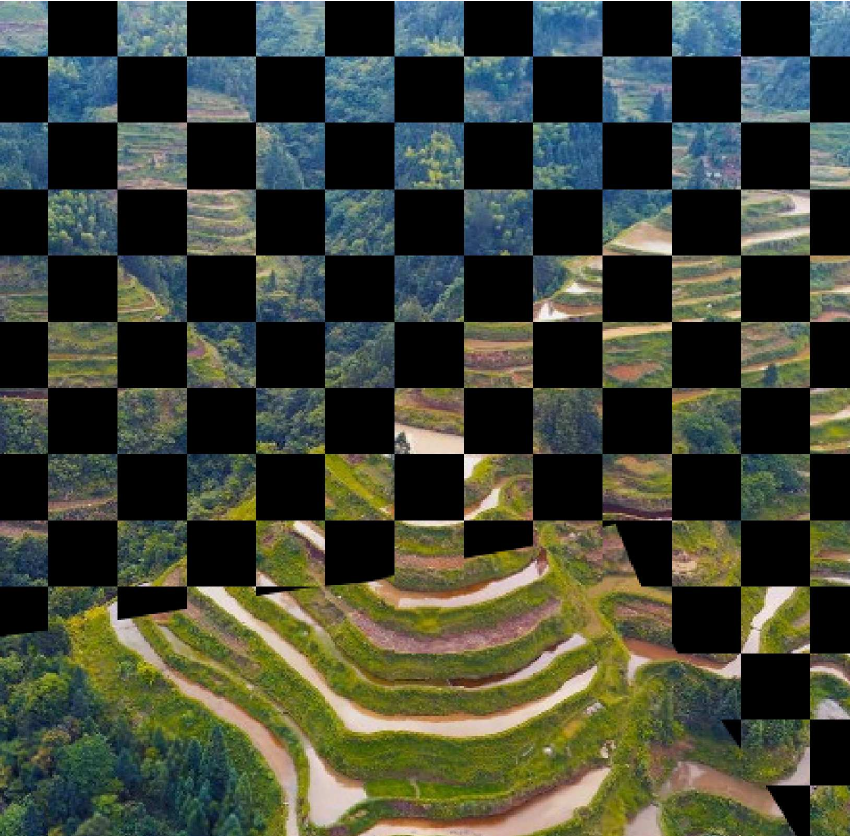}}
 \centerline{\footnotesize(b)}
\end{minipage}
\begin{minipage}[t]{.12\textwidth}
 \centering
 \centerline{\includegraphics[width=1.0\textwidth,height=20mm]{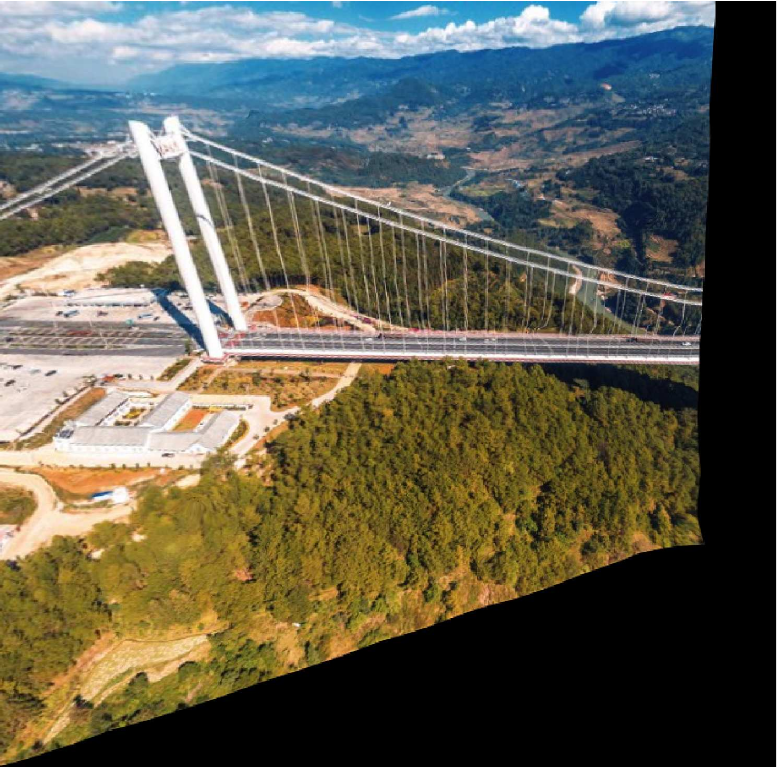}}
  \centerline{\includegraphics[width=1.0\textwidth,height=20mm]{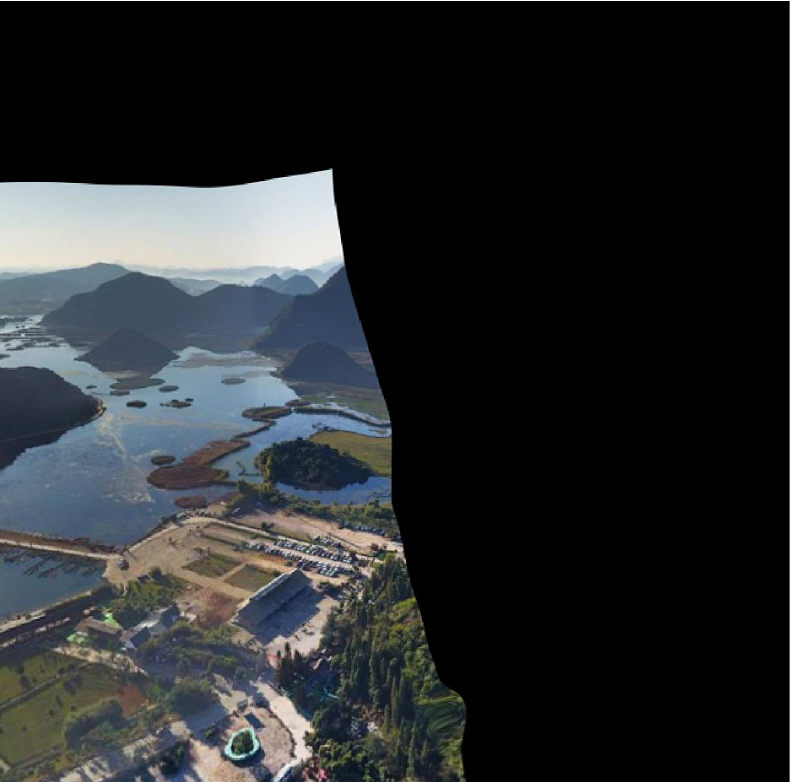}}
   \centerline{\includegraphics[width=1.0\textwidth,height=20mm]{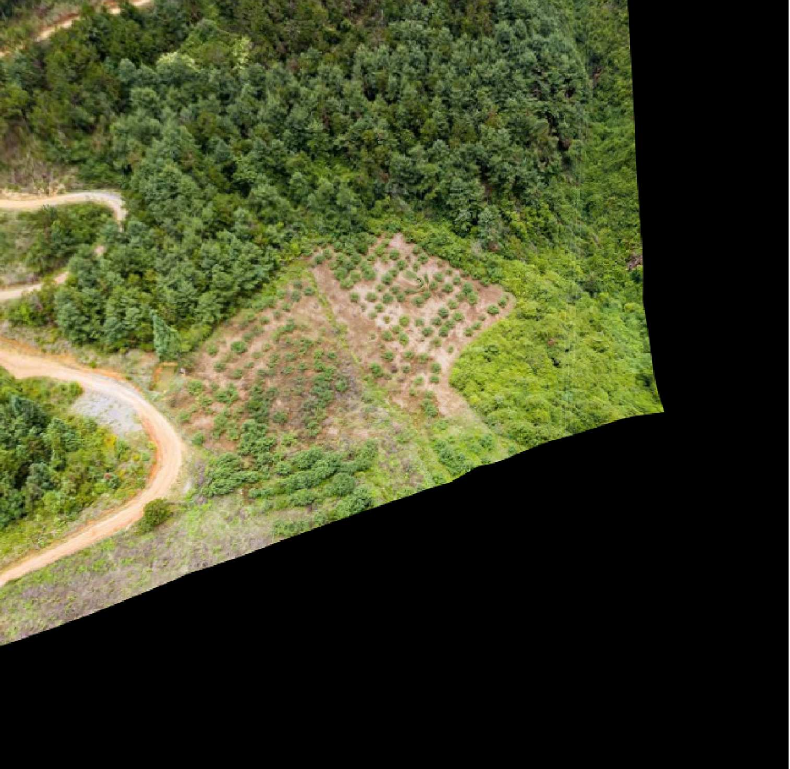}}
        \centerline{\includegraphics[width=1.0\textwidth,height=20mm]{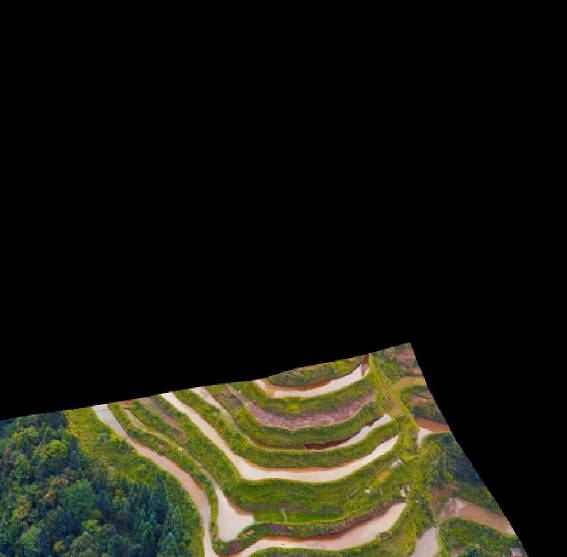}}
             \centerline{\footnotesize(c)}
\end{minipage}
\caption{Some remote sensing image registration results obtained by the proposed LSC method.
(a) Input images consist of the (left) source and (right) target image in each group. (b) and (c) are the checkerboard image and the warped sensed image, respectively.}
\label{fig:imageregistionexample}
\end{figure}
\begin{table}[t]
\centering
\caption{{Quantitative comparison results obtained by the three competing methods on remote sensing images for the task of image registration. $\downarrow$ means that the lower the value, the better.}}
\scalebox{0.9}{\begin{tabular}{cccccc}
\toprule
 Method& RMSE$\downarrow$ & MAE$\downarrow$ & MEE$\downarrow$ & Time (s)$\downarrow$ \\
\midrule
RANSAC & {22.43} & {81.27} & {24.67} & {1.76} \\
LGF & {16.00} & {72.25} & {1.22} & {13.78} \\
LSC & {5.53} & {25.95} & {0.00} & {1.02} \\
\bottomrule
\end{tabular}}
\label{tab:ImageRegistration}
\end{table}
In this subsection, we evaluate the performance of the proposed LSC on remote sensing dataset~\cite{jiang2020robust}, for the task of image registration that aims to estimate the transformation model to align source images with target images with the maximum extent. We compare LSC with the deterministic method LGF, and we also use RANSAC as baseline. We show some remote sensing image registration results obtained by the proposed LSC in Fig.~\ref{fig:imageregistionexample}. In Table~\ref{tab:ImageRegistration}, we report the mean value of the root mean square error (RMSE), maximum error (MAE), median error (MEE) and the CPU time obtained by the three competing methods~\cite{jiang2020robust}:
\begin{equation}
RMSE=\sqrt{\frac{1}{L} \sum_{i=1}^{L}\left\|t_{i}-{H}\left(s_{i}\right)\right\|_{2}^{2}},
\end{equation}
\begin{equation}
MAE={max} \left(\left\|t_{i}-{H}\left(s_{i}\right)\right\|_{2}\right)_{i=1}^{L},
\end{equation}
\begin{equation}
MEE={median}\left(\left\|r_{i}-{H}\left(s_{i}\right)\right\|_{2}\right)_{i=1}^s{L},
\end{equation}
where $s_{i}$ and $t_{i}$ are $i$-th pixel coordinates in source and target images, respectively. $L$ and $H$ are the number of coordinates and the transformation derived from the source and target image. $\|\cdot\|_{2}$ denotes the Euclidean norm of vectors.

From Fig.~\ref{fig:imageregistionexample} and Table~\ref{tab:ImageRegistration}, we can see that LSC is able to obtain good performance for image registration. For the all four evaluation metrics, LSC show obvious advantages over RANSAC and LGF. Note that, LGF is slower than RANSAC, since LGF includes a hash function that often tasks much time in complex scenes.

\section{{Ablation study}}
\label{sec:limitations}
\subsection{The impact of sampling algorithms}
\begin{figure}[t]
\centering
\begin{minipage}{.238\textwidth}
\centerline{\includegraphics[width=1.0\textwidth]{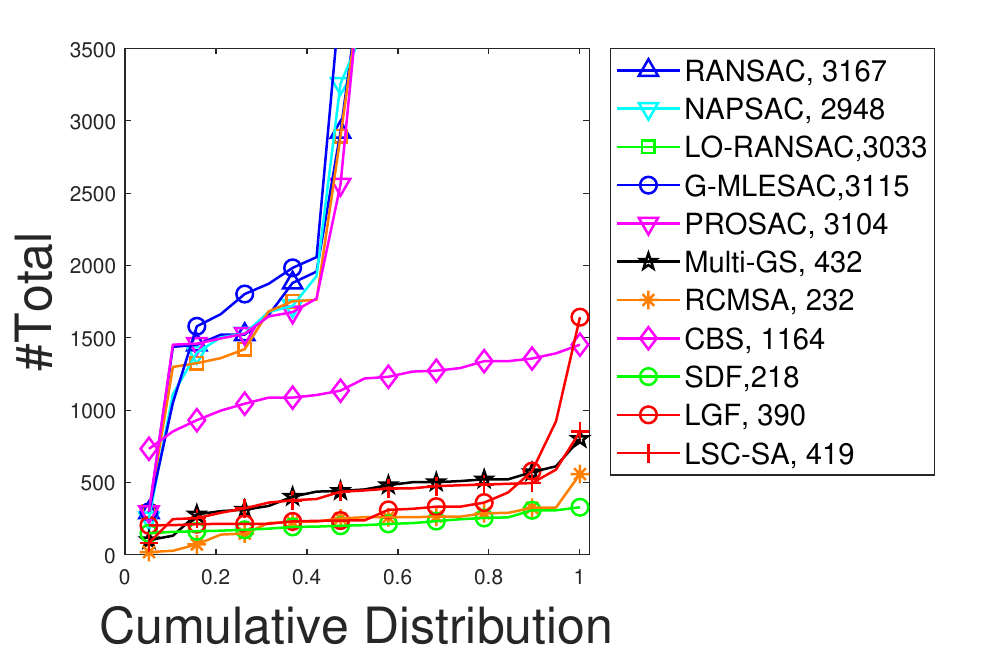}}
\centerline{\includegraphics[width=1.0\textwidth]{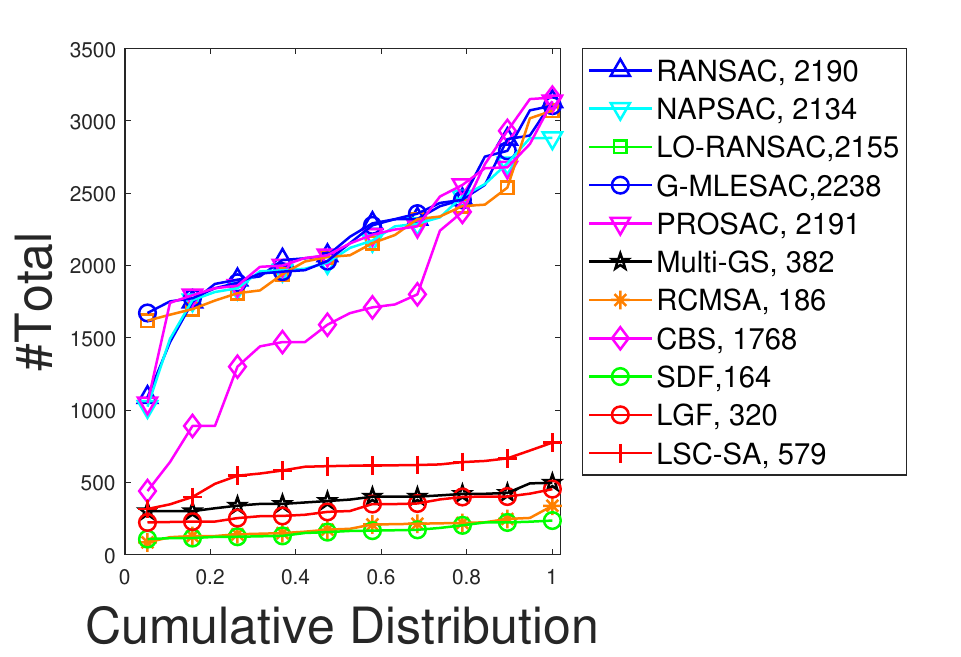}}
 \centerline{\footnotesize(a) Total}
 \end{minipage}
 \begin{minipage}{.238\textwidth}
\centerline{\includegraphics[width=1.0\textwidth]{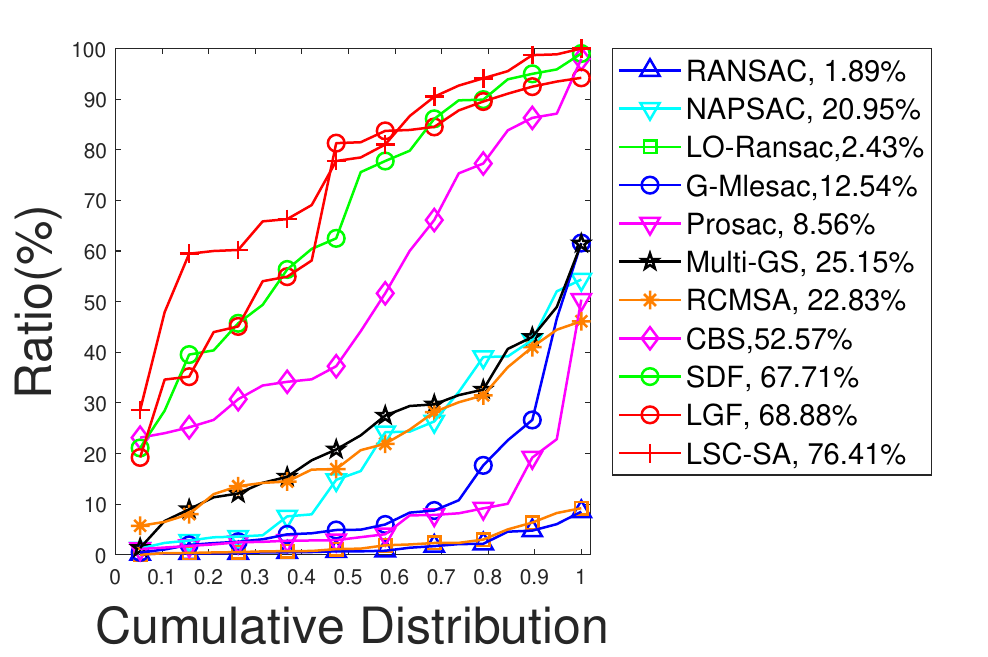}}
\centerline{\includegraphics[width=1.0\textwidth]{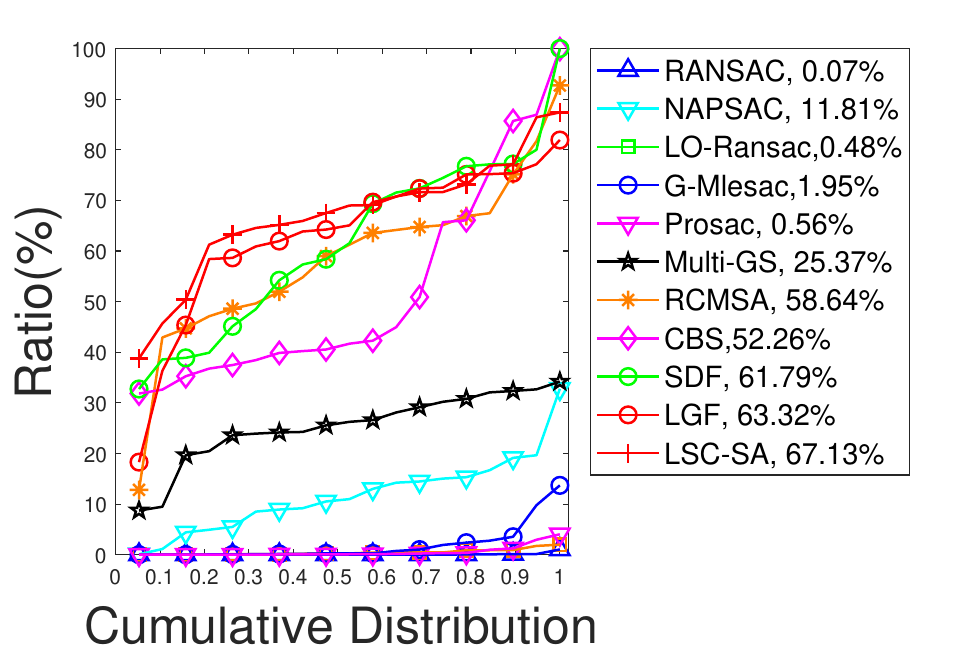}}
 \centerline{\footnotesize(b) Ratio}
\end{minipage}
\caption{Quantitative comparisons of ten sampling methods on all image pairs from the AdelaideRMF dataset for (Top) homography estimation and (Bottom) fundamental matrix estimation. (a) The total number of sampled minimal subsets and (b) the ratio of all-inlier minimal subsets on the sampled minimal subsets obtained by the competing methods within one second with respect to the cumulative distribution.}
\label{fig:samplingresults}
\end{figure}
A sampling algorithm is very important for the final fitting performance, thus, we further discuss the sampling algorithm in this section. Specifically, we provide a quantitative comparison on all image pairs from the AdelaideRMF dataset with the ten state-of-the-art sampling methods, including RANSAC, NAPSAC~\cite{nasuto2002napsac}, {LO-RANSAC}~\cite{chum2003locally}, G-MLESAC~\cite{tordoff2005guided}, {PROSAC}~\cite{chum2005matching}, Multi-GS~\cite{chin2012accelerated}, RCMSA~\cite{pham2014random}, CBS~\cite{tennakoon2018effective}, SDF~\cite{IJCVXiao2019} and LGF~\cite{xiao2020deterministic}.

We report the total number of sampled minimal subsets and the ratio of all-inlier minimal subsets on the sampled minimal subsets (called the inlier-subset ratio) obtained by the competing methods within one second in Fig.~\ref{fig:samplingresults}. We can see that, most of random sampling methods achieve a large number of minimal subsets within one second, while three deterministic fitting methods (i.e., SDF, LGF and LSC-SA) only sample a few hundred subsets. This is because most of random sampling methods try to cover all model instances in data by increasing the number of subsets, in contrast, deterministic sampling methods often adopt a heuristic strategy to improve the quality of subsets while reducing the subsets including outliers. Thus,  the inlier-subset ratios obtained by deterministic sampling methods are significantly higher than the ones obtained by random sampling methods.

For deterministic sampling methods, LSC-SA shows obvious advantages over SDF and LGF, that is, LSC-SA not only samples the most number of subsets among the three deterministic sampling methods, but also achieves the largest inlier-subset ratio.
\subsection{The impact of initialization sampling density}
\begin{figure}[t]
\centering
\begin{minipage}{.24\textwidth}
\centerline{\includegraphics[width=1.0\textwidth]{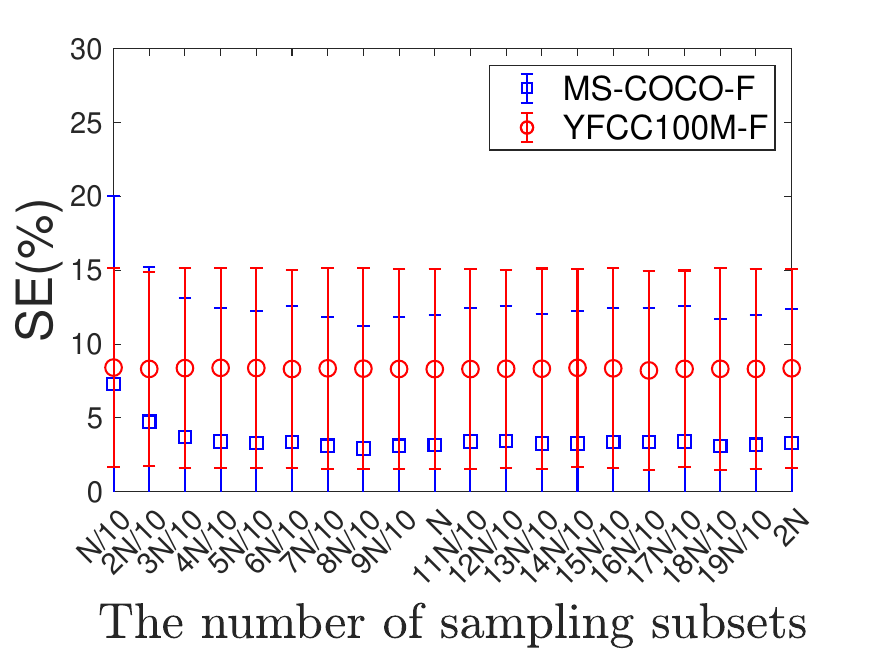}}
\end{minipage}
\begin{minipage}{.24\textwidth}
\centerline{\includegraphics[width=1.0\textwidth]{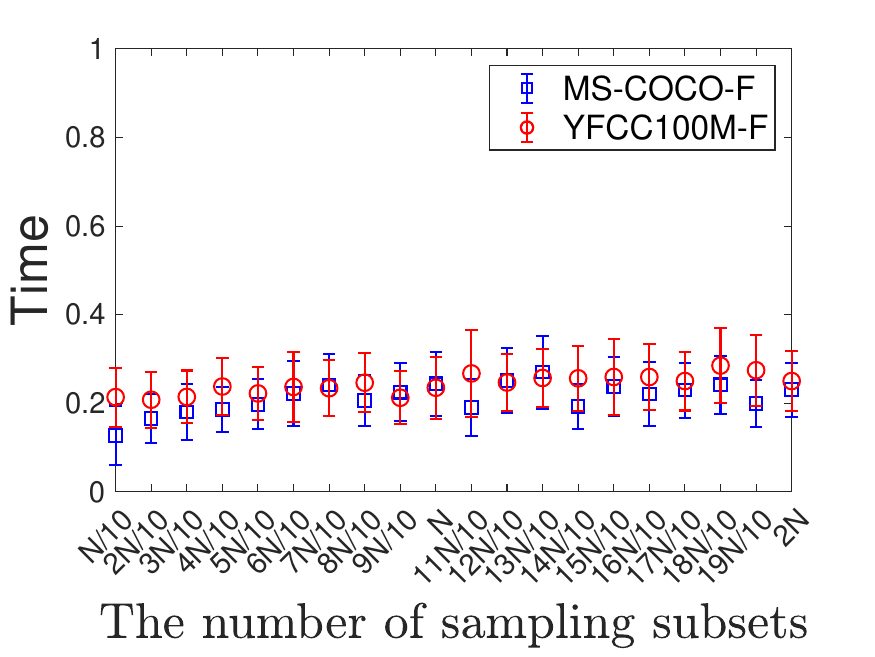}}
\end{minipage}
\caption{{Quantitative results obtained by the proposed LSC fitting scheme with different numbers of initialization sampling subsets for the task of homography estimation and fundamental matrix estimation on the MS-COCO-F and YFCC100M-F dataset, respectively.}}
\label{fig:samplingnumber}
\end{figure}

To evaluate the impact of initialization sampling density on the proposed method, we test different numbers of initialization sampling subsets for the task of homography estimation and fundamental matrix estimation on the MS-COCO-F and YFCC100M-F dataset, respectively. We set the sampling number from $N/10$ to $2N$, increasing by $N/10$ each time, where $N$ represents the number of input correspondences. We report the quantitative results in Fig.~\ref{fig:samplingnumber}.

We can see that, on the MS-COCO-F dataset, LSC achieves a relatively stable mean value of SE when the sampling number exceeds $N/5$. On the YFCC100M-F dataset, there is no significant fluctuation in SE at different sampling numbers. In terms of the CPU time, LSC also exhibits no significant variation. Thus, LSC is not sensitive to sampling initialization across different densities.
\subsection{The impact of model selection strategy}
\begin{figure}[ht]
\centering
\begin{minipage}{.24\textwidth}
\centerline{\includegraphics[width=1.0\textwidth]{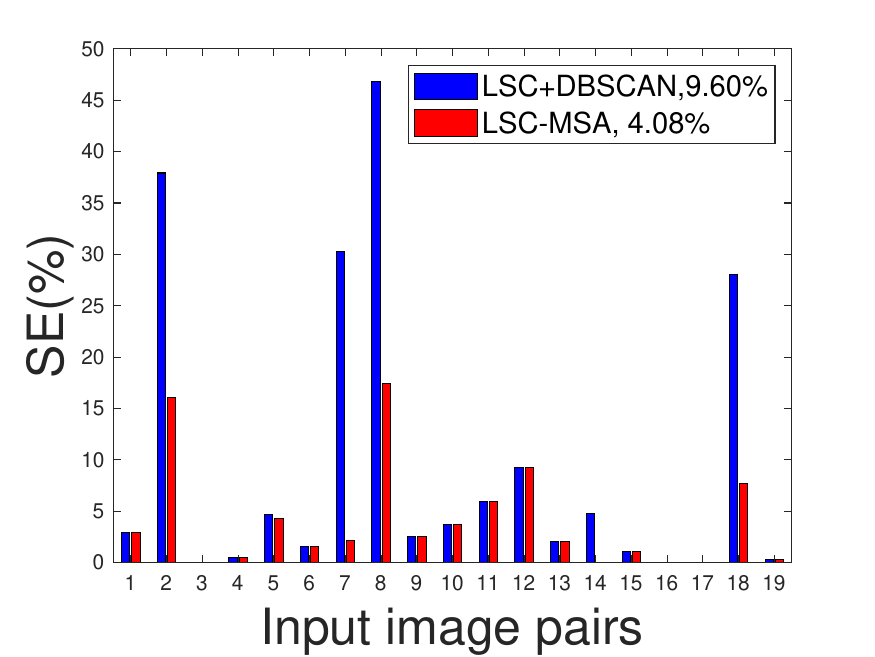}}
\end{minipage}
\begin{minipage}{.24\textwidth}
\centerline{\includegraphics[width=1.0\textwidth]{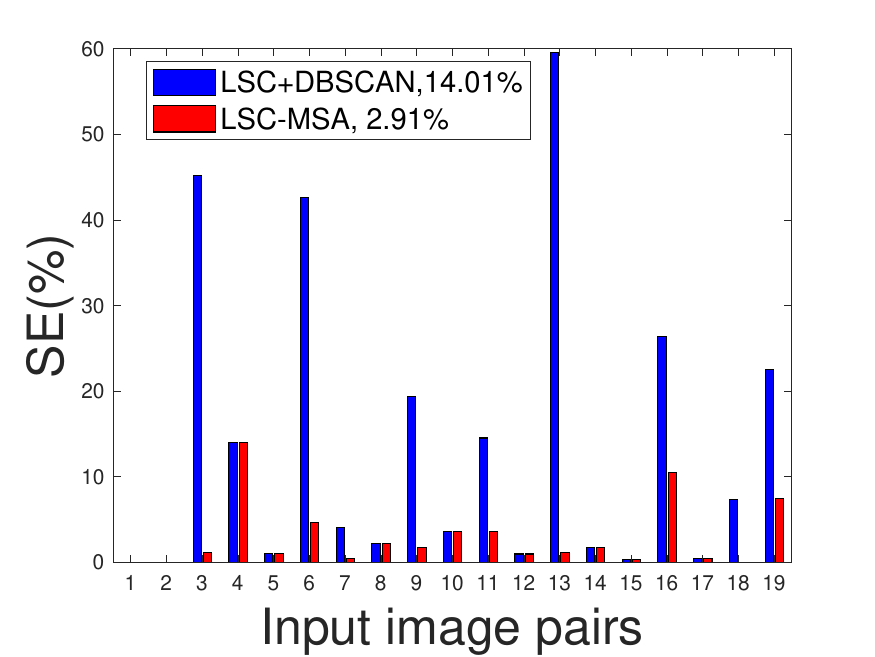}}
\end{minipage}
\caption{{Quantitative results obtained by the proposed LSC fitting scheme with different model selection algorithms for the task of (Left) homography estimation and (Right) fundamental matrix estimation on the AdelaideRMF dataset.}}
\label{fig:dbscan}
\end{figure}

The proposed LSC includes an important components, i.e. LSC-MSA for model selection. To evaluate the impact of LSC-MSA, we test a new version LSC-DBSCAN that uses DBSCAN~\cite{mar1996} to cluster the remaining points in MH-LSS for model selection. We perform LSC-MSA and LSC-DBSCAN on all image pairs of the AdelaideRMF dataset for the task of homography estimation and fundamental matrix estimation, and report SE in Fig.~\ref{fig:dbscan}.

We can see that, for the homography estimation, LSC-MSA and LSC-DBSCAN are able to obtain good performance on $15$ out of the $19$ image pairs. LSC-MS still achieves low SE on the remaining four image pairs, but LSC-DBSCAN fails. For the fundamental matrix estimation, LSC-MSA and LSC-DBSCAN achieves the same values of SE on $10$ out of the $19$ image pairs, and LSC-MSA achieves lower values of SE than LSC-DBSCAN on all the remaining nine image pairs. This can show the effectiveness of LSC-MSA on the model selection for model fitting.
\section{Conclusions}
\label{sec:conclusion}
In this paper, we propose LSC, a deterministic fitting method designed for general multi-structural model fitting problems. LSC provides consistent and reliable solutions compared to many existing methods. Unlike current deterministic approaches, LSC handles both multi-structural data and general model fitting problems, such as line and circle fitting. Specifically, LSC leverages two latent semantic spaces of data points and model hypotheses, preserving consensus to remove outliers, generate high-quality model hypotheses, and effectively estimate model instances. Experimental results on both synthetic data and real images demonstrate that LSC outperforms several state-of-the-art model fitting methods in terms of accuracy and speed.

While LSC demonstrates commendable fitting performance, its model selection algorithm may encounter challenges when the number of models significantly increases, especially in complex fitting tasks. This challenge primarily arises from the uneven distribution of model hypotheses across various models in the input data. As the number of models escalates, this existing imbalance becomes more pronounced, posing difficulties for most fitting methods to identify models with fewer model assumptions. Moreover, optimal results often require adjusting parameter settings for various fitting tasks, driven by distinctions in data characteristics, noise levels and task complexities. These issues will be the focus of our ongoing research efforts.

 \section{Acknowledgment}
{\small This work was supported by the National Natural Science Foundation of China under Grants 62125201, 62072223 and 62020106007.}
{\small
\bibliographystyle{IEEEtran}
\bibliography{Pami4}
}
\begin{IEEEbiography}[{\includegraphics[width=1in,height=1.05in,clip,keepaspectratio]{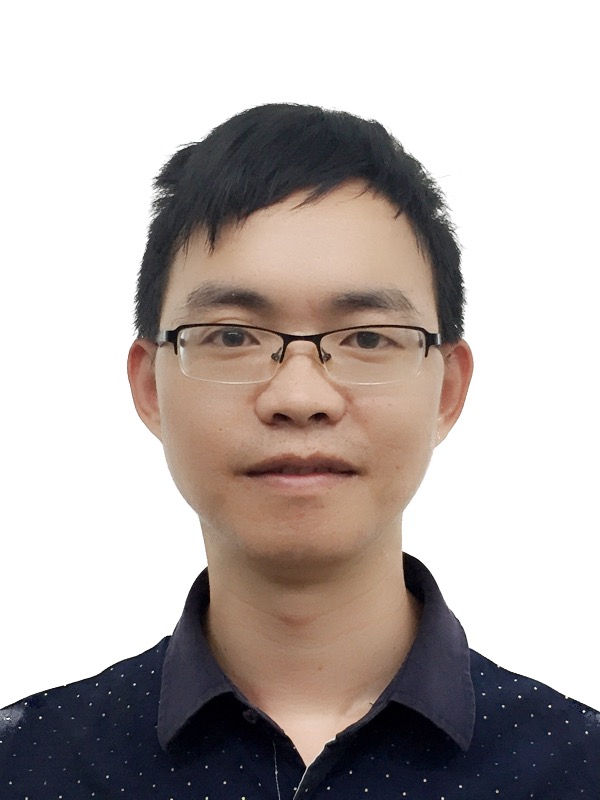}}]{Guobao Xiao}\scriptsize  (Senior Member, IEEE) received the Ph.D. degree from Xiamen University, China. He is currently a Tenured Professor at Tongji University, China. He has published over $50$ papers in journals and conferences including IEEE TPAMI/TIP, IJCV, ICCV, ECCV, etc. His research interests include machine learning, computer vision and pattern recognition. He has been awarded the best PhD thesis award in China Society of Image and Graphics (a total of ten winners in China). He also served on the program committee (PC) of CVPR, ICCV, ECCV, etc.
\end{IEEEbiography}
\begin{IEEEbiography}[{\includegraphics[width=1in,height=1.15in,clip,keepaspectratio]{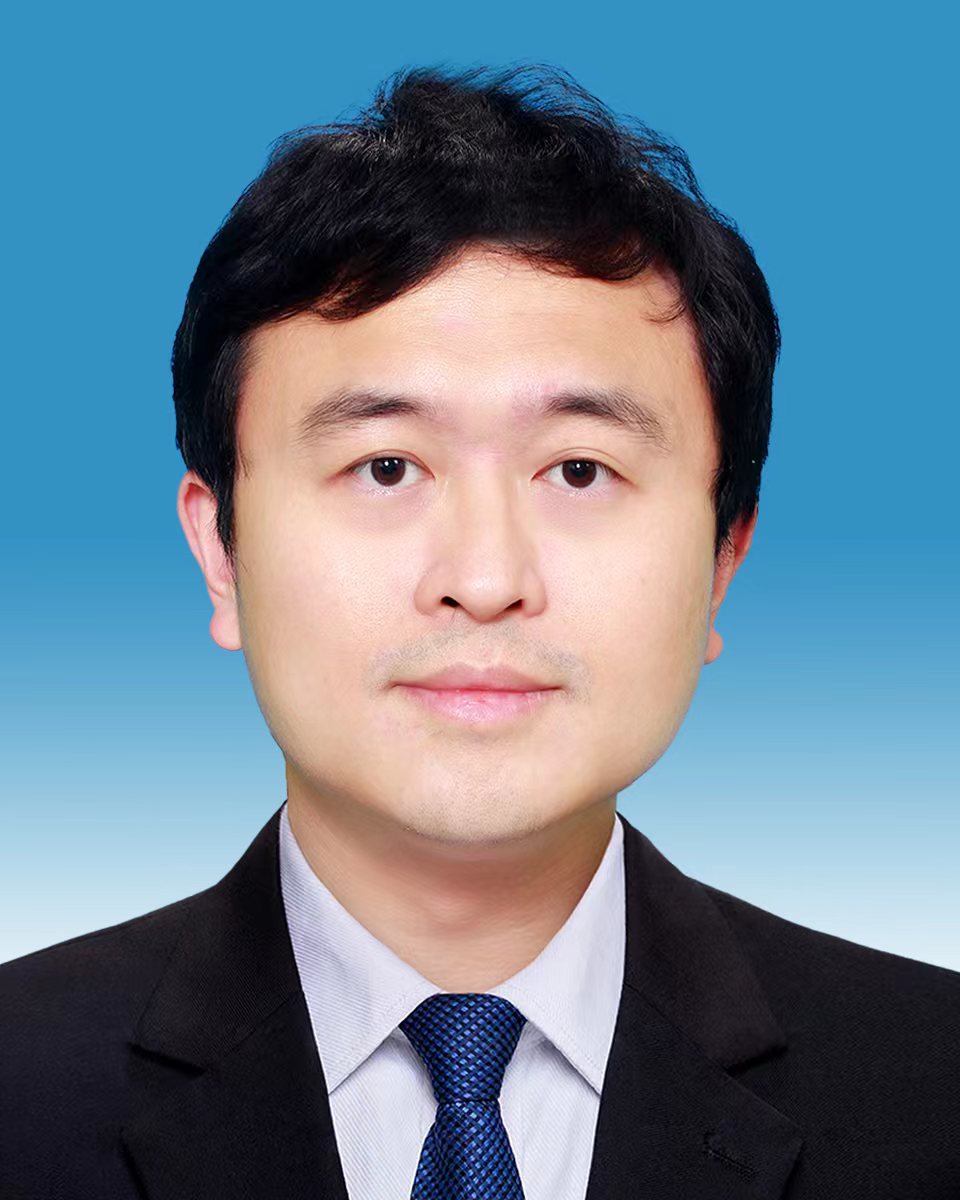}}]{Jun Yu}\scriptsize (Senior Member, IEEE) received the B.Eng. and Ph.D. degrees from Zhejiang University, China. He is currently a Professor at Harbin Institute of Technology (Shenzhen), China. He has authored or co-authored more than 100 scientific articles. His research interests have included multimedia analysis, machine learning, and image processing. He has served as a program committee member for top conferences including CVPR, ACM MM, AAAI, IJCAI, and has served as associate editors for prestigious journals including IEEE Trans. CSVT and Pattern Recognition.
\end{IEEEbiography}
\begin{IEEEbiography}[{\includegraphics[width=1in,height=1.05in,clip,keepaspectratio]{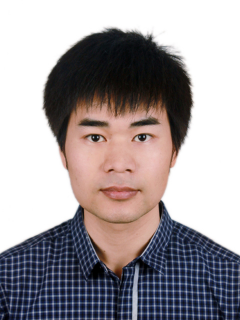}}]{Jiayi Ma}\scriptsize
 (Senior Member, IEEE) received the B.S. and Ph.D. degree from the Huazhong University of Science and Technology, Wuhan, China. He is currently a Professor at Wuhan University. He has authored or co-authored more than 200 refereed journal and conference papers, including IEEE TPAMI/TIP, IJCV, CVPR, ICCV, ECCV, \emph{etc}. He has been identified in the 2019-2022 Highly Cited Researcher lists from the Web of Science Group. He is an Area Editor of \emph{Information Fusion}.
\end{IEEEbiography}
\begin{IEEEbiography}[{\includegraphics[width=1in,height=1.05in,clip,keepaspectratio]{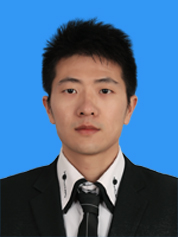}}]{Deng-Ping Fan}\scriptsize
 (Senior Member, IEEE)  received his PhD degree from the Nankai University in 2019. He joined Inception Institute of AI in 2019. He has published about $25$ top journal and conference papers such as TPAMI, CVPR, ICCV, ECCV, etc. His research interests include computer vision and visual attention, especially on Camouflaged Object Detection. He won the Best Paper Finalist Award at IEEE CVPR 2019, the Best Paper Award Nominee at IEEE CVPR 2020.
\end{IEEEbiography}

\begin{IEEEbiography}[{\includegraphics[width=1in,height=1.05in,clip,keepaspectratio]{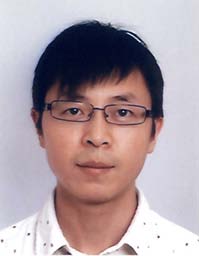}}]{Ling Shao}\scriptsize
 (Fellow, IEEE)is a Distinguished Professor with the UCAS-Terminus AI Lab, University of Chinese Academy of Sciences, Beijing, China. He was the founding CEO and Chief Scientist of the Inception Institute of Artificial Intelligence, Abu Dhabi, UAE. He was also the Initiator, founding Provost and EVP of MBZUAI, UAE. His research interests include generative AI, vision and language, and AI for healthcare. He is a fellow of the IEEE, the IAPR, the BCS and the IET.
\end{IEEEbiography}
\end{document}